\documentclass{article} 
\usepackage{iclr2027_conference,times}

\usepackage{hyperref}
\usepackage{url}
\usepackage{tabularx}

\usepackage{booktabs}       
\usepackage{amsfonts}       
\usepackage{nicefrac}       
\usepackage{microtype}      
\usepackage{xcolor}         
\usepackage{threeparttable}
\usepackage{multirow} 
\usepackage{makecell}
\usepackage{graphicx}
\usepackage{nicematrix}
\usepackage{subcaption}
\usepackage{algorithm}
\usepackage{algpseudocode}
\usepackage{amsmath}
\usepackage{amsthm}
\usepackage[most]{tcolorbox}
\usepackage{fvextra}
\usepackage{fancyvrb}

\usepackage{amsmath,amsfonts,bm}
\usepackage{enumitem}

\newtheorem{theorem}{Theorem}
\newtheorem{lem}{Lemma}
\newtheorem{definition}{Definition}

\newcommand{\Real}{\mathbb{R}}
\newcommand{\Natural}{\mathbb{N}}
\newcommand{\Z}{\mathbb{Z}}
\newcommand{\E}{\mathbb{E}}
\newcommand{\R}{\mathbb{R}}
\newcommand{\reject}{\tau^{\times}}

\newcommand{\given}{\,|\,}

\newcommand{\indic}[1]{\mathbf{1}\!\left\{#1\right\}}

\definecolor{baselinered}{RGB}{192,0,0}
\definecolor{baselinedarkred}{RGB}{139,0,0}

\definecolor{myred}{RGB}{214, 39, 40}
\definecolor{myblue}{RGB}{31, 119, 180}

\definecolor{mygreen}{RGB}{106, 168, 79}
\definecolor{mydarkblue}{RGB}{28, 69, 135}

\newcommand{\va}[2]{#1$\pm$#2}
\newcommand{\best}[1]{\textbf{#1}}
\newcommand{\good}[1]{\textbf{\underline{#1}}}  
\newcommand{\norm}[1]{\textbf{\textit{#1}}}

\title{A statistical measure of trust in generative AI}

\author{Antiquus S.~Hippocampus, Natalia Cerebro \& Amelie P. Amygdale \thanks{ Use footnote for providing further information
about author (webpage, alternative address)---\emph{not} for acknowledging
funding agencies.  Funding acknowledgements go at the end of the paper.} \\
Department of Computer Science\\
Cranberry-Lemon University\\
Pittsburgh, PA 15213, USA \\
\texttt{\{hippo,brain,jen\}@cs.cranberry-lemon.edu} \\
\And
Ji Q. Ren \& Yevgeny LeNet \\
Department of Computational Neuroscience \\
University of the Witwatersrand \\
Joburg, South Africa \\
\texttt{\{robot,net\}@wits.ac.za} \\
\AND
Coauthor \\
Affiliation \\
Address \\
\texttt{email}
}

\author{
  \textbf{Alina Ermilova}$^{1,}$\thanks{Corresponding author. Email: \texttt{a.ermilova@applied-ai.ru}}, \textbf{Valeria Yakupova}$^1$, \textbf{Nikita Sushko}$^2$, \textbf{Andrey Lovyagin}$^1$, \textbf{George Ovchinnikov}$^2$, \\
  \textbf{Maksim G. Sharaev}$^{3,1,4}$, \textbf{Evgeny M. Mirkes}$^1$, \textbf{Denis Turdakov}$^5$, \textbf{Ivan Y. Tyukin}$^{6,7}$ \\
  $^1$Applied AI Institute \quad $^2$Independent researcher \quad $^3$Hamad Bin Khalifa University \\
  $^4$BIMAI-Lab \quad $^5$Lomonosov Moscow State University \quad $^6$King's College London \quad $^7$University of Leicester 
}

\iclrfinalcopy 
\begin{document}

\maketitle

\begin{abstract}
Trustworthiness is a crucial issue for generative AI, as modern models can produce convincing and confident responses even when their outputs are incomplete, unsupported, or wrong. The frequent absence of ground-truth knowledge or labels complicates verification of generative AI outputs. 
To address this issue, we propose a trustworthiness framework grounded in model-internal confidence calibrated against an external expert reference. We estimate \textbf{per-token trustworthiness} of a student generative model as the statistical typicality of each generated token relative to an expert model's distribution, where the expert is used as a distributional reference rather than as a source of factual verification labels. To this end, we introduce an \textbf{AI ontology calibrator} that leverages the student model's internal latent features and token-level output statistics to compute expert-reference acceptance scores. 
Combined with traditional token-level confidence, entropy, and perplexity statistics, calibrator scores yield a compact \textbf{set of $15$ response-level indicators} for correctness classification. Evaluated across BF program generation, mathematical problem solving, and medical question answering, combined calibrator and traditional indicators provide complementary reliability signal which could be used to build  detectors of untrustworthy outputs and hallucinations. Remarkably, our experiments show that these detectors are transferable across domains, although transferability is not always symmetric.

\end{abstract}

\section{Introduction}

Generative AI has become a common component of modern computational systems, supporting coding, text processing, search, learning, and decision making. As these systems are increasingly used in consequential domains, the problem of trustworthiness becomes central. This is particularly visible for Large Language Models (LLMs), which can produce fluent and confident responses even when the underlying content is incomplete, unsupported, or wrong. 

Current gold standard approaches to  trustworthiness in LLMs 
emphasize factual incorrectness~\citep{bazarova2025hallucinationdetectionllmstopological, li2024dawn, ye2023cognitive}, unfaithfulness to input context~\citep{zhang2023siren, cossio2025taxonomy}, or internal inconsistency~\citep{mundler2023selfcontradictory}.
They presuppose access to some notion of ground truth, such as an external knowledge base, a reference document, an objective fact, or a set of internally consistent statements~\citep{huang2023survey}. Dependence on ground truth introduces a practical limitation. In open-domain settings, long-tail knowledge may be unavailable or poorly documented. In domains such as medicine, correctness can be context-dependent, and accepted standards may evolve over time. 
These constraints motivate methods that estimate trustworthiness without requiring verified factual labels at inference time.


\textbf{Our goal} is circumvent this issue by showing learnability of alternative confidence-style reliability signals in a student-expert framework enabling to ensure statistical trustworthiness. There a smaller generative model constitutes a \textit{student}. A larger \textit{expert} model can be used offline during calibration. The expert is not treated as an oracle and does not define absolute truth. Instead, it provides a distributional reference against which the student's token-level behavior can be calibrated. At deployment time, the resulting trustworthiness indicators are computed from the student model's own internal states and token-level statistics.

To this end, we propose an \textbf{AI ontology calibrator}. By ontology we denote a task domain, an operational representation of its  outputs, and evaluation protocol. The calibrator compares student hidden representations and token-level output with expert-reference acceptance scores. The scores estimate if a generated token agrees with the expert reference under chosen alignment criteria. 


We evaluate the framework on three ontologies of increasing complexity. In {\it BF program generation}, correctness can be verified by executing a generated program and comparing its output to the requested behavior. In {\it mathematical reasoning}, correctness is evaluated using reference-based and ProcessBench-style labeling protocols, reflecting the fact that multiple solution paths may be valid. In {\it medical multiple-choice question answering}, we evaluate several student LLMs using response-level labels obtained through an automated annotation pipeline, and interpret detector outputs as reliability or routing signals rather than clinical validation. Our theoretical analysis gives a finite-sample justification for the calibration step. 
 Empirical results show that calibrator-derived indicators provide information complementary to standard confidence features. 

\textbf{Our specific contributions are as follows:}
\begin{enumerate}
 \item We frame trustworthiness estimation through the lens of confidence, distinguishing it from uncertainty and consistency. Complementary to approaches looking for factual errors, we treat unreliable generations as outputs whose token-level internal signals are atypical relative to an expert-reference calibration criterion.

\item We propose \textit{an AI ontology calibrator} -- a tool for estimating token-level trustworthiness of a student generative model using an expert model as an offline distributional reference. 

\item As part of the AI ontology calibrator, we introduce expert-reference acceptance scores: confidence-style token-level statistics grounded in the typicality of generated tokens relative to the expert reference. These scores are supported by bounds on threshold-defined acceptance and rejection probabilities.

\item We construct a small set of $15$ response-level indicators by aggregating calibrator scores with traditional token-level confidence, entropy, and perplexity statistics. These indicators can be used to train supervised classifiers for discriminating correct and incorrect responses under ontology-specific labeling protocols.

\item We evaluate the proposed indicators across BF program generation, mathematical problem solving, and medical question answering. The combined calibrator and traditional indicators provide complementary reliability signal, with the clearest gains in Medicine and BF and more mixed behavior in Math.

\item We analyze transfer across models within the same ontology and across different ontologies. The results show selected improvements from calibrator-derived features, but transfer remains heterogeneous rather than universal.

\end{enumerate}

\section{Related Work}\label{sec:related}

An overview of classical approaches such as consistency-based, retrieval-based, and model-based are discussed in Appendix~\ref{appendix:related2}. 
Since our method studies trustworthiness through the lens of confidence, 
 we concentrate here on confidence-based and internal-signal methods.  

\textit{Low-confidence validation} methods monitor minimum token probabilities, token entropy, or related scalar statistics to flag candidates for external verification~\citep{varshney2023stitch}. Length-Normalized Token Probability (LNTP)~\citep{bouchard2025uncertainty} is a representative example: it aggregates token probabilities into a length-normalized sequence-level score, providing a strong and efficient baseline that avoids multiple generations. However, such approaches usually rely on raw intrinsic confidence signals and remain vulnerable to miscalibration, especially after instruction tuning, where aleatoric ambiguity and epistemic ignorance can become conflated~\citep{tao2025revisiting}.

To better isolate \textit{epistemic uncertainty}, ensemble-based estimation has been proposed. Input clarification ensembling generates $K$ semantic rephrasings, while Monte Carlo Dropout approximates Bayesian posterior variance via $T$ stochastic forward passes~\citep{hou2023decomposing, gal2016dropout, zeevi2024monte}. These methods can provide richer uncertainty estimates, but they incur linear overhead in the number of samples or stochastic passes and typically operate at the sequence or concept level rather than as token-level reliability scores available from a single generation.

Beyond scalar probability and entropy measures, hidden-state and logit-based methods aim to use richer internal model information. \textit{INSIDE}~\citep{chen2024inside} and related probing approaches leverage hidden states across transformer layers via eigenvalue analysis or learned classifiers, while energy-based methods such as Semantic Energy use pre-softmax logits to mitigate entropy saturation~\citep{ma2025semantic}. These methods show that internal representations contain reliability-relevant information, but they are often framed as out-of-distribution or uncertainty detectors and do not directly calibrate token-level student behavior against an external expert reference.

Our work occupies a complementary position. Like confidence-based methods, it uses model-internal signals and can produce lightweight indicators from a single student generation. Unlike raw confidence proxies, however, the proposed AI ontology calibrator uses an expert model offline as a distributional reference to calibrate student hidden states and token-level statistics. The resulting expert-reference acceptance scores are not factuality guarantees, but calibrated reliability features that can be aggregated with traditional confidence, entropy, and perplexity-style statistics for supervised response-level correctness prediction.


\section{Overall Approach}
\label{sec:background}

In what follows, a generative model whose trustworthiness is being assessed will be referred to as a \textbf{student}.  We assume that the student's generation process can be decomposed into individual steps and at each step the student generates a \textbf{token}. Since we focus on LLMs, tokens include whole words, subword units, or single characters as determined by the underlying LLM tokenizer.

At the core of our approach is the observation that a statistically grounded measure of a student model's token-level behavior, relative to what an expert would produce in the same context and ontology, correlates with human judgments of the quality and factuality of the student's overall response \citep{yuan2021bartscore, fu2024gptscore}. We build on this by treating such a measure as a signal, among others, for detecting whether a response can be trusted.


In order to test and assess our method, we consider three ontologies of increasing complexity, \textbf{BF}, \textbf{Math}, and \textbf{Medicine}, and their corresponding  trustworthiness/correctness definitions:

\begin{itemize}
    \item \textbf{BF program generation (BF)}:  a student-LLM generated program is considered trustworthy if it compiles, executes, and produces  correct intended output.
    \item \textbf{Math problem solving (Math)}:  a generated solution is considered trustworthy if it correlates with reference solution according to an agreed metric. Reference solution constitutes solution provided by an external \textbf{\textit{expert}}, a human or another AI system. 
    Technical details explaining how we compared student's and expert's responses are provided in Section~\ref{sec:methodology_our_ontologies}.
    \item \textbf{Medical question answering (Medicine)}: student LLMs answer multiple-choice questions with all reasoning traces, and  \textbf{\textit{expert}} responses are taken as a reference. In this paper, we treat all inaccuracies as equally important. A promising direction for future research is to introduce semantic-context awareness, where errors are weighted by their clinical significance.  A variant of labeling accounting for context criticality is presented in Appendix~\ref{appendix:medicine}.
\end{itemize}

Before diving into details of the method, we want to distinguish between three related but fundamentally different concepts for evaluating statistical  trustworthiness of a model: 
\begin{itemize}
    \item \textbf{Uncertainty} is a \textit{problem-level} property that characterizes the spread of the answer space for a given prompt, regardless of any particular output.
    \item \textbf{Consistency} measures \textit{behavioral stability} across multiple outputs, requiring at least two generations or inputs to compare.
    \item \textbf{Confidence}, by contrast, is an \textit{instance-level} property of a \textit{single prediction} -- it reflects how strongly internal or expressed signals point to correctness for that specific answer.
\end{itemize}   

Our framework is specifically designed to estimate trustworthiness through a \textit{single generation pass} and with no ground truth information required. This contrasts with consistency measures which require at least two outputs or inputs to be compared. 
Regarding remaining two measures, 
we prefer confidence over uncertainty for two complementary reasons. First, uncertainty is a \textit{prompt-level} property shaped by factors external to the prediction itself such as training data distribution, prompt ambiguity, and the overall spread of the answer space. 
Confidence, by contrast, is an \textit{instance-level} signal reflecting the model's internal conviction about a specific generated element, independently of these distributional factors and without any knowledge of the model's training data. Second, in many real-world ontologies, ground truth is neither static nor universally accessible: medical treatment protocols are revised as evidence accumulates, scientific consensus shifts, and accepted best practices change over time. 
By grounding our framework in the model's own internal confidence signals, we obtain a formulation that remains feasible in absence of external ground truth stability. 

\section{Methodology}
\label{sec:methodology}

\begin{figure}[tb]
\centering
\includegraphics[width=\linewidth]{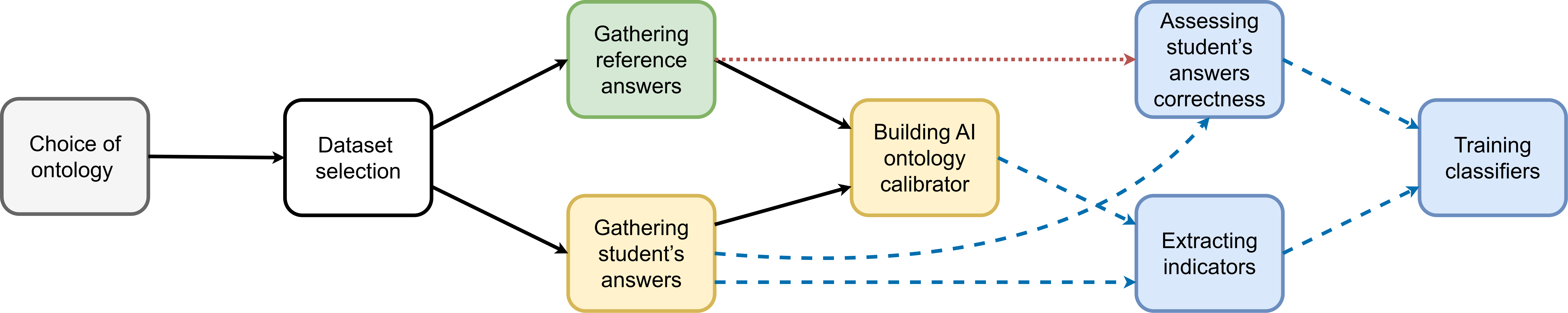}
\caption{Our workflow. \textbf{Solid lines} indicate the necessary steps for building an AI ontology calibrator; \textbf{\textcolor{myblue}{dashed lines}} correspond to the proposed procedure for obtaining indicators and evaluating how well they reflect response correctness; the \textbf{\textcolor{myred}{dotted line}} highlights the optional use of reference answers for labeling the student's correctness for classifier training.}
\label{fig:workflow}
\end{figure}

Our framework comprises two components: (1)  measuring \textbf{statistical confidence of each generated token}, relative to that of expert's  through a single generation pass without accessing expert's own confidence, uncertainty and state, and without ground truth information; and (2) determining whether the \textbf{generated response as a whole is trustworthy}, by aggregating per-token indicators that combine our proposed statistical confidence scores with traditional measures. The resulting workflow is illustrated in Figure~\ref{fig:workflow}, whose components are described in detail in the following sections.

\subsection{Framework component 1: AI ontology calibrator}
\label{sec:methodology_ai_calibrator}
The AI ontology calibrator's  aim is to assess  each student token's typicality with respect to the conditional distribution induced by the reference answer. The process of its construction is further illustrated with \textbf{solid lines} in Figure~\ref{fig:workflow}.  Reference answers are responses provided by \textbf{\textit{expert}} models -- SOTA models of much larger size than the student models. For each token position we learn token-conditional generation probability with  expert's response taken as a reference. 

More formally, let $G$ be a generative AI model mapping inputs $u \in \mathcal{U}$ to finite output sequences $\left(T_1, T_2, \dots, T_m\right)$, $m \in \mathbb{N}$, where $m$ is the length of generated sequence and $T_i$ represents a token. In this paper, we consider the map $G$ represents an LLM, while our method can be implemented to other generative models. Let us denote the set of possible tokens by $\mathcal{T} = \{\tau_1, \dots, \tau_q\}$. Since output sequences are generated sequentially, generation decomposes into individual steps. For each step of the generation, $G$ generates many objects, including \textbf{the vector of probabilities for all possible tokens} and \textbf{the vector of features} $\phi = \Phi(\cdot)$. Here we use outputs of the last fully connected layer before the heads as features, albeit other model variables could be used too. 

The response generation process of an AI model proceeds as follows. Let us denote input $u$ corresponding to generation of the first token  as $u_1=u$. Output of the first step of generation is a $q$ dimensional vector of probabilities $s_1$, and the vector of features is $\Phi(u)\in\Real^d$, where $d$ is the dimension of the last fully connected layer before the heads. Based on the values of $s_1$, the system selected one token $T_1$.  At the next step, the model's  input becomes $u_2=u\cup \{T_1\}$, and the process repeats. As a result,  generation of a single token in the output sequence can be modelled as a map from the set of inputs $\mathcal{U}$ (inputs are different for different steps) to a  finite set of outputs $\mathcal{T}$:
\begin{equation}\label{eq:generative_model}    G:\ \mathcal{U} \to \mathcal{T}.\end{equation}
To distinguish different steps of generation, input at the $k$-th step is denoted as $u_k=u\cup_{j=1}^{k-1}\{T_j\}$.

In what follows student model will be denoted as $G^S$, and expert (a model judge or a human) will be referred to as $G^E$. In response to the same initial input these models produce two sequences of tokens: $\left(T^E_1,\ldots, T^E_{m_E}\right)$ (expert) and $\left(T^S_1,\ldots, T^S_{m_S}\right)$  (student). For the student, we assume access to its vector of features $\Phi(u_j)$. For the expert model, this is not required.

\textbf{Our goal is} to estimate the probability of correctness of token $T^S_i$. 
Token $T^S_i$ is deemed correct if $T^S_i = T^E_i$, and it is considered incorrect otherwise. If $m_E<m_S$, then all tokens $T^S_i$ with $i>m_E$ are deemed wrong. 
This enables posing the task of identifying "correct" generations as a classical classification problem at the level of each individual token $\tau_i$: we need to find a map accepting or rejecting tokens produced by $G^S$ with some inferrable confidence. This map could formally be defined as:
$$  \mathcal{A}^i: \mathcal{T}\times\Real^d \to  \mathcal{T}\cup \{\tau^{\times}\},$$
where $\tau^{\times}$ is the label corresponding to rejected tokens.

As a measure of model $G^S$ confidence relative to expert's response we consider conditional probabilities (or suitable proxies of):
$$P(\mathcal{A}^i(G^S(u_k),\Phi(u_k))=\tau_j|G^S(u_k)=\tau_j, G^E(u_k)=\tau_j),$$
where $\mathcal{A}^i$ is map discussed above, and $k$ is the token's position.
Low values of these probabilities indicate disagreement between $G^S$'s outputs and the target distribution, and therefore insufficient trust. 
We also can estimate the probability of the opposite event that token $\tau_j$ produced by $G^S$ is in position $k$ does not match that of $G^E$:
$$P(\mathcal{A}^i(G^S(u_k),\Phi(u_k))=\tau^\times|G^S(u_k)=\tau_j, G^E(u_k)\neq\tau_j).$$
If the distribution underpinning the above data generation process would have been known then these probabilities could be computed. However, in most practically-relevant instances, this is not the case, and hence we need to learn or estimate the corresponding probabilities from empirical data. 

In this work, we introduce a tool to estimate these probabilities which we are referring to in the text as an \textbf{AI ontology calibrator}.  
To form/build the calibrator, we assume access to a dataset, which we will refer to as the \textbf{forming dataset}:
\begin{equation}\label{eq:training_dataset}
\mathcal{S} = \{(u_i,\tau_i) \; : \; u_i \in \mathcal{U},\; \tau_i \in \mathcal{T},\; i = 1,\dots,M\}. 
\end{equation}
The forming dataset $\mathcal{S}$ is further partitioned into
$\mathcal{S}_j=\{(u,\tau) \in\mathcal{S}\; : \; \tau=\tau_j\}$,
where each such set is non-empty.
For each  $j=1,\dots,q$, partition $\mathcal{S}_j$ into:
\[
\mathcal{S}_{+,j} = \{(u,\tau)\in \mathcal{S}_j \mid \tau=\tau_j\}, \ \mathcal{S}_{-,j} = \{(u,\tau)\in\mathcal{S}_j \mid \tau\neq \tau_j\},
\]
that are assumed not empty, with cardinalities $M_{+,j}$ and $M_{-,j}$.
Let $\mathcal{H}=\{h_1,\dots,h_q\}$ be a class of measurable projection functions $h : \mathbb{R}^d \to \mathbb{R}$.
Each function $h_j$ defines empirical cumulative distribution functions:
\begin{align}\label{eq:F+-}
F_{+,j}(s) & = \frac{1}{M_{+,j}} \sum_{(u,\tau)\in\mathcal{S}_{+,j}} \mathbf{1}_{h_j(\Phi(u)) \le s}, \ \  
F_{-,j}(s) = \frac{1}{M_{-,j}} \sum_{(u,\tau)\in\mathcal{S}_{-,j}} \mathbf{1}_{h_j(\Phi(u)) \le s},
\end{align}
and $\mathbf{1}_D$ is the indicator function of the set $D$. Detailed theoretical background and justification of the calibrator's capability to learn desired probability estimates from empirical data are presented in Appendix~\ref{appendix:theorem}.


In general, we have several attributes for each token: the output token itself  $G^S(u_j)=T_j$, the vector of output probabilities (softmaxes in standard models) $(s^j_1,\ldots,s^j_q)$, and the vector of features $\Phi(u_j)=(\phi^j_1,\ldots,\phi^j_n), j$ is the position in the generated sequence. 
The vectors $\phi^j$ are used as feature vectors $\Phi(u)$ for the calibrator. These vectors could be used `as is' without any pre-processing. For each token $i$ with rich enough forming set (in this study we consider only tokens that occur at least $1,000$ times in the dataset) we computed three arrays: (1) projection functions $h_j\in\mathcal{H}$, which for simplicity have been chosen as functionals projecting data onto normalized direction of the corresponding Fisher discriminants;
(2) thresholds of acceptance $th_i^j$ (score representation of $th_i^j=F_{-,j}^\dagger(\Delta_j^{i})$), where $\Delta_j^{i} \in \left(0, 1\right)$;
(3) the vector of the adjusted lower bounds on the probabilities of acceptance $p_i^j$ for the corresponding threshold $th_i^j$. 
   
Calculation (estimation) of the probability of correct acceptance of the token $\tau_j$ with feature vector $\phi$ can be computed as follows. Calculate dot product of feature vector with corresponding projection function $r=(h_\tau,\phi)$. Define the maximal $k$, such that $th_\tau^k<r$. Take  $p_\tau=p_\tau^k$ as answer. For each generated token $T_j$,  compute the following quantities:
\begin{enumerate}
    \item the probability of correct acceptance $p_j$ (or its lower bound), hereafter referred to as \textbf{calibrator probabilities};
    \item the maximum softmax probability of the generated token $s^j = s_1^j$ (where $s_1^j, ..., s_q^j$ are sorted in the descending order), i.e., the probability assigned to the model's top-predicted token at position $j$, hereafter referred to as \textbf{softmaxes};
    \item the Shannon entropy of the predictive distribution $e^j = -\sum_{i=1}^{q} s_i^j \ln(s_i^j)$, hereafter referred to as \textbf{entropies};
    \item the log-probability of the next generated token under the model's predictive distribution at position $j$, $l^j = \ln P(T_{j+1} \mid T_1, \ldots, T_j) = \ln s^j_{\mathrm{rank}(T_{j+1})}$, where $\mathrm{rank}(T_{j+1})$ indicates the position of the actual token $T_{j+1}$ in the sorted softmaxes list; hereafter referred to as \textbf{token log-probabilities}.
\end{enumerate}
Quantities (1)-(3) are collectively used to construct a set of \textbf{indicators} for  training trustworthiness classifiers, while quantities (4) are utilized for \textbf{baseline indicators (baselines)} calculation. Details on indicator construction and classifier training are provided in the following section.

\subsection{Framework component 2: determining response correctness}
\label{sec:methodology_train_classifiers}
For each generated response of an LLM, we obtain four per-token quantities described in Section~\ref{sec:methodology_ai_calibrator}. To determine trustworthiness of responses, we aggregate quantities into indicators, and train a classifier mapping indicators of each generated sequence to a probability that the response/generation is correct (trustworthy). We thus now deal with a binary classification task where labels are available. During training, responses whose probability of being correct $\ge 0.5$ are considered correct (see Section~\ref{sec:methodology_our_ontologies} for details).

Our set of indicators for each generated sequence includes the following:
\begin{itemize}
    \item Mean ($\bar{\cdot}$), standard deviation ($std(\cdot)$), and geometric mean ($\bar{G}_\cdot$) aggregations of calibrator probabilities ($p$), softmaxes ($s$), and entropies ($e$).
    \item Entropy-like ($E_{\cdot}$) and perplexity-like ($PP_{\cdot}$) values of calibrator probabilities and softmaxes.
    \item Pearson ($PCC$) and Spearman ($SCC$) correlation coefficients between calibrator probabilities and entropies.
\end{itemize}

The complete set of indicators comprises $15$ quantities:
\begin{equation}\label{eq:15_indicators}    \{\bar{p}, \bar{s}, \bar{e},\ std(p), std(s), std(e),\ PCC, SCC,\ E_{p}, E_{s},\ PP_{p}, PP_{s},\ \bar{G}_p, \bar{G}_s, \bar{G}_e\}.\end{equation}

\begin{figure}[tb]
\centering
\includegraphics[width=0.7\linewidth]{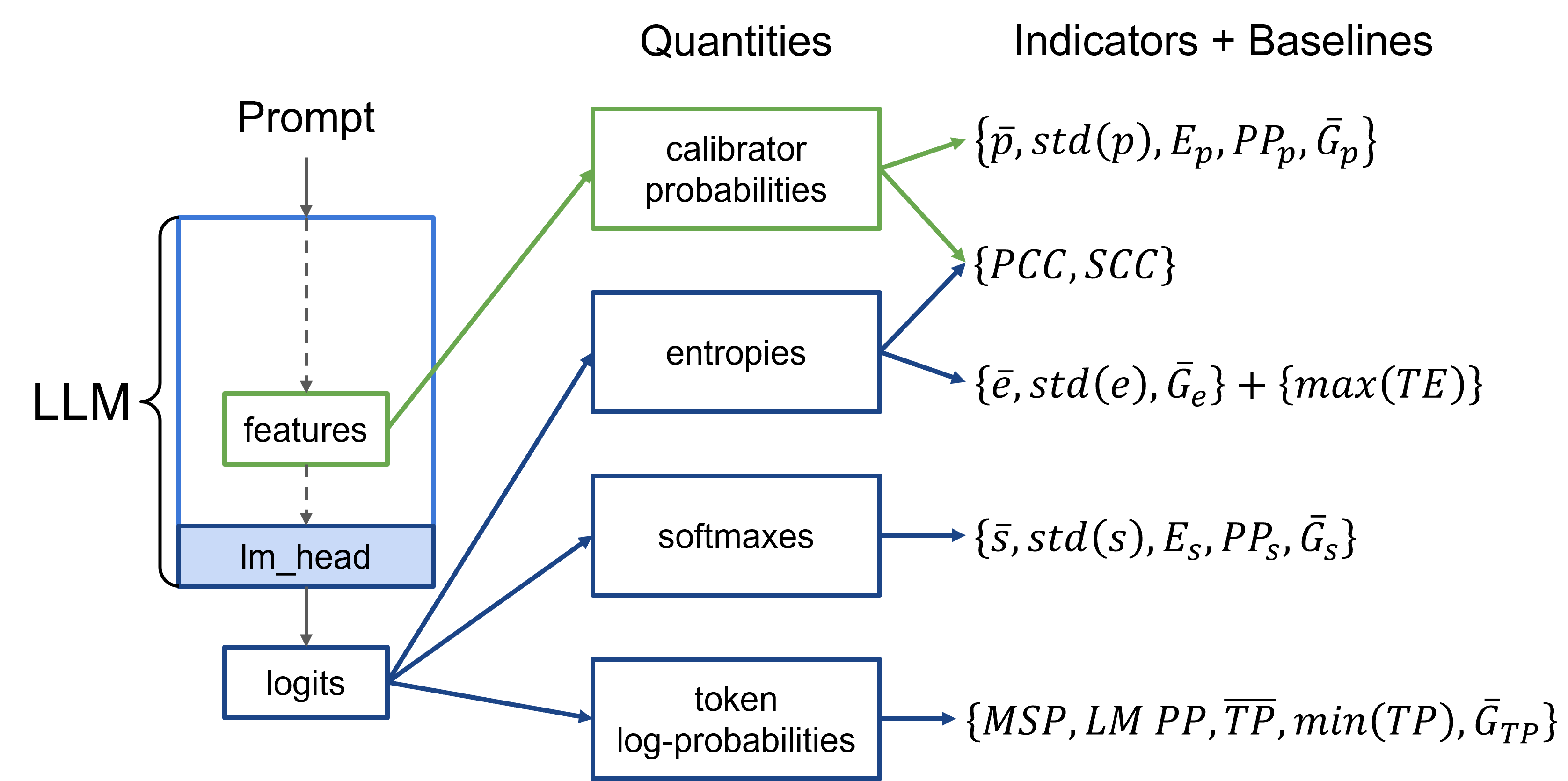}
\caption{Sources of derivation of quantities, indicators, and baselines. Entropies provide both indicators and baselines, token log-probabilities -- only baselines, calibrator probabilities and softmaxes -- only indicators. \textbf{\textcolor{mygreen}{Green arrows}} highlight indicators produced by our proposed calibrator probabilities, \textbf{\textcolor{mydarkblue}{blue arrows}} -- traditional indicators. $\{PCC, SCC\}$ -- indicators of mixed category: to calculate them, both calibrator probabilities and traditional indicators are needed.} 
\label{fig:indicators_sources}
\end{figure}

Among these indicators, we have two traditional confidence measures, namely, Mean Token Entropy ($\bar{e}$) and LNTP~\citep{bouchard2025uncertainty}($\bar{G}_s$), along with the newly proposed ones. Moreover, two of our indicators coincide with standard baselines under greedy decoding
: Mean Token Probability ($\overline{TP}$) $\approx \bar{s}$ and Geomean Token Probability ($\bar{G}_{TP}$) $\approx \bar{G}_s$. Note also that LM Perplexity ($LM\ PP$)
$= 1/\bar{G}_{TP}$. We also consider MSP, Max Token Entropy ($\max{(TE)}$), and Min Token Probability ($\min{(TP)}$) as baselines. A diagram illustrating computation of all provided indicators and baselines based on quantities from Section~\ref{sec:methodology_ai_calibrator} is shown in Figure~\ref{fig:indicators_sources}.

\subsection{Specifics of our framework and the considered ontologies}
\label{sec:methodology_our_ontologies}
For math, we use the first $5,000$ problems to build the calibrator and train classifiers in the following settings: (V1) concatenate the first $5,000$ problems with the second $5,000$ problems and train classifiers via $20$-fold cross-validation; and (V2) train classifiers on the first $5,000$ problems and test them on the second $5,000$ problems. As Figure~\ref{fig:workflow} shows, using reference answers when labeling students' responses is not always necessary. For math, we compare two labeling strategies: (1) \textbf{LLM-Judge} labeling with Qwen2.5-32B, which uses expert answers, and (2) \textbf{ProcessBench} labeling with DeepSeek-V3.2, which does not require expert answers. For the (V1) and (V2) settings described above, we use only those problems from the first $5,000$ and the second $5,000$ that have both LLM-Judge and ProcessBench labels. For BF and medicine, we build calibrators on the full datasets and train classifiers via $20$-fold cross-validation. For BF labeling, we select $10,000$ programs and label them by compiling each program and comparing its output with the desired output: programs whose outputs match are assigned label $1$ (correct), while those that do not are assigned label $0$ (incorrect). For medical question answers, we apply the ProcessBench technique with GLM-4.7.

\section{Experiments}
\label{sec:experiments}

Table~\ref{tab:exp_setup} details  experimental setups for all ontologies and the statistics of built calibrators. Full implementation is available in~\citep{anonymous2026code}.

\begin{table*}[t]
    \caption{Experimental setups, naming convention, and the statistics on the built calibrators and the classifier data for all ontologies. In the classifiers category, we indicate the total number of considered responses with labels of correctness (denoted as ``Total''), the number of correct (``1s'') and incorrect (``0s'') responses. ``-'' means the absence of an experiment for an ontology.}
    \label{tab:exp_setup}
    \centering
    \resizebox{\textwidth}{!}{%
    \begin{NiceTabular}{*{9}{c}}[hlines, vlines={2,3,4,5,6,7,8,9},
                                 cell-space-limits=2pt]
    \Block{1-2}{Ontology} & & BF & \Block{1-2}{Math} & & \Block{1-4}{Medicine} & & & \\
    \Block{1-2}{Data for building calibrators} & &
        \Block{1-1}{Custom 433,688 pairs\\of (program, output)} &
        \Block{1-2}{Open-source 5,000 pairs of\\(problem, solution) -- batch 1} & &
        \Block{1-4}{Open-source 12,369 pairs of\\(problem, answer)} & & & \\
    \Block{1-2}{Expert} & & BF compiler & \Block{1-2}{DeepSeek-R1} & & \Block{1-4}{GPT-OSS-120B} & & & \\
    \Block{1-2}{Student (N params)\\* if not mentioned in model name} & &
        \Block{1-1}{Custom Transformer\\(22M)} & \Block{1-2}{Qwen3-8B} & &
        \Block{1-1}{MedGemma\\(4B)} & Qwen3-8B & Qwen3-0.6B & Qwen3.5-0.8B \\
    \Block{1-2}{Unique tokens} & & 8 & \Block{1-2}{18,089} & & 38,926 & 29,481 & 27,783 & 29,826 \\
    \Block{1-2}{Unique calibrated tokens} & & 8 & \Block{1-2}{808} & & 851 & 1,061 & 1,058 & 1,010 \\
    \Block{7-1}{Classifiers} &
        \Block{2-1}{20-fold cross-validation\\(V1)} &
        \Block{2-1}{Total: 10,000\\1s: 3,818\\0s: 6,182} &
        \Block{1-2}{Two concatenated batches\\(batch 1 and batch 2) -- Total: 6,845} & &
        \Block{1-4}{All data from 12,369 question-answer pairs,\\successfully labeled with ProcessBench} & & & \\
    & & & \Block{1-1}{LLM-Judge (Qwen2.5-32B)\\1s: 6,453\\0s: 392}
        & \Block{1-1}{ProcessBench (DeepSeek-V3.2)\\1s: 3,452\\0s: 3,393}
        & \Block{1-1}{Total: 5,978\\1s: 3,919\\0s: 2,059}
        & \Block{1-1}{Total: 3,207\\1s: 2,326\\0s: 881}
        & \Block{1-1}{Total: 3,663\\1s: 564\\0s: 3,099}
        & \Block{1-1}{Total: 1,954\\1s: 595\\0s: 1,359} \\
    & ID for transfer experiment & A & B & C & D & E & F & G \\
    & \Block{4-1}{Train/test\\(V2)} & \Block{4-1}{-} &
        \Block{1-2}{Train -- batch 1 \quad Total: 3,493} & &
        \Block{4-1}{-} & \Block{4-1}{-} & \Block{4-1}{-} & \Block{4-1}{-} \\
    & & & \Block{1-1}{1s: 3,200\\0s: 293} & \Block{1-1}{1s: 1,029\\0s: 2,464} & & & & \\
    & & & \Block{1-2}{Test -- batch 2 \quad Total: 3,352} & & & & & \\
    & & & \Block{1-1}{1s: 3,253\\0s: 99} & \Block{1-1}{1s: 2,423\\0s: 929} & & & & \\
    \end{NiceTabular}}
\end{table*}

\textbf{Ontology 1: BF}. We consider the problem of developing a program in the BF language that produces a specified output. BF is a Turing-complete language consisting of only eight possible symbols. We chose BF due to its simplicity and the straightforward process of verifying the correctness of a generated program. The dataset comprises $433,688$ program-output pairs. Programs are constrained to a maximum of $18$ commands, while outputs are limited to $256$ character codes. A detailed description of BF is presented in Appendix~\ref{appendix:bf_programs_generation_details}. Details regarding the model architecture and training procedure are provided in Appendix~\ref{appendix:tr_train_details}. 

\textbf{Ontology 2: Math}. In this setup, we consider mathematical tasks from ~\citep{2025synthetic1}. Since the tasks are open-ended mathematical problems requiring full reasoning traces rather than just final answers, evaluation of their correctness cannot be reduced to simple answer matching. Manual annotation of such responses would be both time-consuming and costly -- particularly given that most tasks admit multiple valid solution paths. To address this, we employ an LLM to generate reference solutions (as described in Section~\ref{sec:methodology_ai_calibrator}), offering a practical balance between annotation quality and scalability. Our aim is not to establish absolute ground truth, but to validate a general method for assessing student-model quality in a regime where expert's answers can be treated as references. Specifically, as \textit{an expert}, we use DeepSeek R1~\citep{DeepSeek} (671B parameters), one of the strongest publicly available LLMs~\citep{guo2025deepseek}, generating a single reference solution per task, while as \textit{a student} we use Qwen3-8B~\citep{Qwen3_8b}, which is a relatively small LLM. To evaluate the correctness of the student's solutions, we employ two techniques. Details about labeling strategies and their comparison are provided in Appendix~\ref{appendix:math_labels_descr}.

\textbf{Ontology 3: Medicine}. 
We use \textit{Medical-Reasoning-SFT-GPT-OSS-120B} dataset~\citep{med_questions_hf}. From the original collection, we sample and curate $12,369$ questions, focusing on multiple-choice items that require reasoning about human biology and medicine. To improve topical consistency, we apply a series of filters and exclude: (i) animal-related questions, (ii) questions about medical legal procedures, (iii) ecology- and taxonomy-focused questions, (iv) non-English questions, (v) plant and marine biology questions, (vi) questions centered on organizations (e.g., WHO, UNICEF), and (vii) microbiology questions (e.g., bacteria-related items). We found that a subset of slightly more than $12,000$ questions provides sufficient coverage for our experiments while keeping evaluation and analysis tractable. Most questions focus on medical conditions and diagnosis, as well as the biological processes underlying diseases and broader aspects of human biology. As the dataset has already provided the results of GPT-OSS-120B generations, we consider them as the \textit{expert's answers}. As student LLMs, we consider four models:
\begin{itemize}
    \item \textbf{MedGemma-1.5-4B}~\citep{google2026medgemma} is a multimodal instruction-tuned variant of Gemma~3, specifically trained on medical text and images;
    \item Qwens~\citep{Qwen3_8b, qwenteam2026qwen35}: two representatives of the Qwen3 family -- \textbf{Qwen3-8B} and \textbf{Qwen3-0.6B} -- and one from the more recent Qwen3.5 small series, \textbf{Qwen3.5-0.8B}.
\end{itemize}

To annotate correct answers we use ProcessBench labeling~\citep{processbench} with GLM-4.7.  

\textbf{Results: prediction of generations correctness.}
As described in Section~\ref{sec:methodology_train_classifiers}, for each student from each ontology, we obtain a set of $15$ indicators and six baselines: MSP, $LM\ PP$, $\max{(TE)}$, $\overline{TP}$, $\min{(TP)}$, and $\bar{G}_{TP}$. Feature correlations across all considered ontologies are provided in Appendix~\ref{appendix:clf_features_correlations}, and classifier technical details in Appendix~\ref{appendix:clf_techincal_details}. Notably, several papers~\citep{kadavath2022language, lin2022teaching} report a correlation between generation correctness and the model's self-assessment labels; our results on medical ontology show no such correlation (see Appendix~\ref{appendix:clf_features_correlations_medicine}).

Table~\ref{tab:clf_res} presents the classifiers' results. Logistic Regression was trained only on the provided cases, while MLPs were trained on all possible combinations of the $15$ indicators, with only the best results reported here. The table shows that our proposed indicator set consistently outperforms traditional confidence measures. Even using features derived from calibrator probabilities alone provides reasonable quality across all ontologies. This also supports the validity of using the same data for classifier training and calibrator construction, since cross-validation relies on stratified $20$-fold splitting and a weighted loss function. MLPs' results show that the best results were obtained on the indicators subsets including at least one derived from calibrators. The poor performance on Math, V2 reflects a train/test distribution mismatch rather than weaker classifier quality -- particularly for ProcessBench labeling, where the class ratio is inverted between train and test.

\begin{table*}[t]
    \caption{\textbf{Mean $\pm$ standard deviation} of balanced accuracy of Logistic Regression and MLP across $20$ folds for all ontologies. For MLP, we provide only the best results with the subset of indicators used for training. For \textbf{Math, V2}, we provide \textbf{train/test} balanced accuracy. ``15 indicators'' denotes using all proposed indicators, whereas ``12 indicators'' refers to the first 12 indicators defined in Equation~\ref{eq:15_indicators}. For each ontology, top-1 scores are in \textbf{bold}, top-2 in \textbf{\underline{bold underlined}}, and top-3 in \textbf{\textit{bold italic}}. We also provide results for four naive baselines: constant predictions of ones (Const = 1) and zeros (Const = 0), and random predictions where the probability of predicting $1$ is either fixed at $0.5$ (Rand. 0.5) or set to the majority class proportion (Rand. major).}
    \label{tab:clf_res}
    \centering
    \resizebox{\textwidth}{!}{%
    \begin{NiceTabular}{l|l|cccc|c|cc|cc}[hlines={1-4,9,13,20,27,31-34}, vlines={2,3,7,8,10}]
\Block{2-2}{Indicators} & & \Block{1-4}{Medicine} & & & & \Block{1-1}{BF} & \Block{1-2}{Math, V1} & & \Block{1-2}{Math, V2} & \\
 & & MedGemma & Qwen3-8B & Qwen3-0.6B & Qwen3.5-0.8B & Custom Transformer & LLM-Judge & ProcessBench & LLM-Judge & ProcessBench \\
\Block{1-11}{\textbf{Logistic Regression}} & & & & & & & & & & \\
\Block{5-1}{Individual\\(ours)}
 & $\bar{p}$ & \va{0.5681}{0.0299} & \va{0.5131}{0.0433} & \va{0.5674}{0.0403} & \va{0.4922}{0.0562} & \norm{\va{0.7179}{0.1410}} & \va{0.5049}{0.0730} & \va{0.4798}{0.0221} & 0.5026/0.5220 & 0.5162/0.4777 \\
 & $std(p)$ & \va{0.5779}{0.0329} & \va{0.5281}{0.0416} & \va{0.5564}{0.0396} & \va{0.5263}{0.0718} & \va{0.5709}{0.1358} & \va{0.5397}{0.0832} & \va{0.5676}{0.0674} & 0.5101/0.5055 & 0.5217/0.4691 \\
 & $E_{p}$ & \va{0.6701}{0.0377} & \va{0.4857}{0.0392} & \va{0.7890}{0.0518} & \va{0.5311}{0.0486} & \va{0.5509}{0.2132} & \va{0.6511}{0.0725} & \va{0.5134}{0.0570} & \good{0.6004/0.7610} & \good{0.5379/0.5463} \\
 & $PP_{p}$ & \va{0.6693}{0.0361} & \va{0.5708}{0.0356} & \va{0.6375}{0.0447} & \va{0.5154}{0.0483} & \va{0.5028}{0.2195} & \good{\va{0.6528}{0.0831}} & \va{0.4985}{0.0540} & 0.6080/0.7349 & 0.5390/0.5244 \\
 & $\bar{G}_p$ & \va{0.5019}{0.0025} & \va{0.5098}{0.0180} & \va{0.6180}{0.0350} & \va{0.5107}{0.0077} & \va{0.6365}{0.1184} & \va{0.5531}{0.1048} & \va{0.5910}{0.1169} & 0.4927/0.5034 & 0.4953/0.4967 \\
\Block{4-1}{Mixed\\(ours and\\traditional)}
 & $PCC$ & \va{0.5231}{0.0309} & \va{0.5017}{0.0466} & \va{0.5869}{0.0369} & \va{0.5189}{0.0438} & \va{0.5314}{0.1194} & \va{0.4338}{0.0579} & \va{0.5703}{0.0784} & 0.5453/0.5163 & 0.5120/0.5096 \\
 & $SCC$ & \va{0.5161}{0.0239} & \va{0.5359}{0.0261} & \va{0.5406}{0.0542} & \va{0.4818}{0.0611} & \va{0.5235}{0.0934} & \va{0.5238}{0.0524} & \va{0.5118}{0.0500} & 0.5036/0.5267 & 0.4992/0.5108 \\
 & 12 indicators & \good{\va{0.7081}{0.0366}} & \good{\va{0.7245}{0.0414}} & \good{\va{0.8432}{0.0313}} & \good{\va{0.6626}{0.0514}} & \good{\va{0.7813}{0.1483}} & \best{\va{0.6529}{0.0892}} & \norm{\va{0.7099}{0.2264}} & 0.6115/0.5323 & 0.5739/0.4985 \\
 & 15 indicators & \best{\va{0.7197}{0.0320}} & \best{\va{0.7276}{0.0485}} & \good{\va{0.8426}{0.0305}} & \best{\va{0.6634}{0.0557}} & \best{\va{0.7815}{0.1403}} & \norm{\va{0.6523}{0.0901}} & \va{0.7095}{0.2275} & 0.6121/0.5233 & 0.5709/0.4954 \\
\Block{7-1}{Traditional\\(included\\to our\\indicators'\\set)}
 & $\bar{e}$ & \va{0.5892}{0.0284} & \va{0.5680}{0.0509} & \va{0.8087}{0.0361} & \va{0.5026}{0.0037} & \va{0.6921}{0.2067} & \va{0.6099}{0.1639} & \va{0.6723}{0.2059} & 0.5482/0.4997 & 0.5485/0.4991 \\
 & $std(e)$ & \va{0.5507}{0.0296} & \va{0.5940}{0.0456} & \va{0.8203}{0.0298} & \va{0.5367}{0.0510} & \va{0.6807}{0.1161} & \va{0.6179}{0.1918} & \va{0.6935}{0.2198} & 0.5304/0.4980 & 0.5259/0.4967 \\
 & $\bar{G}_e$ & \va{0.6009}{0.0343} & \va{0.5494}{0.0504} & \va{0.7394}{0.0441} & \va{0.5305}{0.0443} & \va{0.7049}{0.1976} & \va{0.5968}{0.1190} & \va{0.6324}{0.1675} & 0.5456/0.4997 & 0.5365/0.4997 \\
 & $\bar{s}$ & \va{0.5761}{0.0311} & \va{0.4839}{0.0659} & \va{0.7525}{0.0426} & \va{0.5026}{0.0037} & \va{0.6842}{0.2050} & \va{0.6016}{0.1594} & \va{0.6611}{0.1954} & 0.5448/0.5127 & 0.5520/0.4991 \\
 & $std(s)$ & \va{0.5815}{0.0312} & \va{0.4292}{0.0567} & \va{0.8108}{0.0327} & \va{0.5469}{0.0543} & \va{0.6957}{0.1446} & \va{0.6280}{0.1823} & \va{0.6837}{0.2147} & 0.5447/0.5044 & 0.5405/0.4981 \\
 & $E_{s}$ & \va{0.6816}{0.0349} & \va{0.4835}{0.0372} & \va{0.7873}{0.0508} & \va{0.5026}{0.0037} & \va{0.6861}{0.1634} & \va{0.6359}{0.0938} & \va{0.5409}{0.0408} & \norm{0.6023/0.7501} & \best{0.5367/0.5580} \\
 & $PP_{s}$ & \norm{\va{0.6825}{0.0333}} & \va{0.5669}{0.0355} & \va{0.6337}{0.0491} & \va{0.5026}{0.0037} & \va{0.6861}{0.1636} & \va{0.6506}{0.1003} & \va{0.5289}{0.0422} & \best{0.6105/0.7650} & \norm{0.5349/0.5352} \\
\Block{11-1}{\color{baselinered}Baselines}
 & \color{baselinered}$\bar{G}_s$ & \color{baselinered}\va{0.5757}{0.0293} & \color{baselinered}\va{0.5516}{0.0516} & \color{baselinered}\va{0.7884}{0.0350} & \color{baselinered}\va{0.5026}{0.0037} & \color{baselinered}\va{0.6867}{0.2047} & \color{baselinered}\va{0.6129}{0.1640} & \color{baselinered}\va{0.6651}{0.1989} & \color{baselinered}0.5491/0.5087 & \color{baselinered}0.5534/0.4990 \\
 & \color{baselinered}MSP & \color{baselinered}\va{0.6531}{0.0303} & \color{baselinered}\va{0.5685}{0.0380} & \color{baselinered}\va{0.6635}{0.0531} & \color{baselinered}\norm{\va{0.5976}{0.0613}} & \color{baselinered}\va{0.6853}{0.1494} & \color{baselinered}\va{0.6317}{0.0861} & \color{baselinered}\va{0.6136}{0.1575} & \color{baselinered}0.6011/0.5289 & \color{baselinered}0.5454/0.4978 \\
 & \color{baselinered}$LM\ PP$ & \color{baselinered}\va{0.5362}{0.0274} & \color{baselinered}\va{0.6036}{0.0412} & \color{baselinered}\norm{\va{0.8224}{0.0327}} & \color{baselinered}\va{0.5309}{0.0469} & \color{baselinered}\va{0.5026}{0.0037} & \color{baselinered}\va{0.6012}{0.1823} & \color{baselinered}\va{0.6148}{0.1866} & \color{baselinered}0.5188/0.5000 & \color{baselinered}0.5367/0.5000 \\
 & \color{baselinered}$\max{(TE)}$ & \color{baselinered}\va{0.5409}{0.0217} & \color{baselinered}\va{0.4892}{0.0548} & \color{baselinered}\va{0.7402}{0.0321} & \color{baselinered}\va{0.5270}{0.0292} & \color{baselinered}\va{0.6679}{0.1637} & \color{baselinered}\va{0.6286}{0.2203} & \color{baselinered}\good{\va{0.7147}{0.2461}} & \color{baselinered}0.5003/0.5000 & \color{baselinered}0.5004/0.5000 \\
 & \color{baselinered}$\overline{TP}$ & \color{baselinered}\va{0.5552}{0.0293} & \color{baselinered}\va{0.5740}{0.0498} & \color{baselinered}\va{0.8124}{0.0311} & \color{baselinered}\va{0.5026}{0.0037} & \color{baselinered}\va{0.7126}{0.2022} & \color{baselinered}\va{0.6148}{0.1866} & \color{baselinered}\va{0.6884}{0.2165} & \color{baselinered}0.5357/0.4998 & \color{baselinered}0.5437/0.5002 \\
 & \color{baselinered}$\min{(TP)}$ & \color{baselinered}\va{0.4971}{0.0223} & \color{baselinered}\norm{\va{0.6225}{0.0311}} & \color{baselinered}\va{0.5074}{0.0391} & \color{baselinered}\va{0.4954}{0.0567} & \color{baselinered}\va{0.4996}{0.0094} & \color{baselinered}\va{0.5302}{0.0644} & \color{baselinered}\va{0.5478}{0.0548} & \color{baselinered}0.5060/0.4998 & \color{baselinered}0.5059/0.5002 \\
 & \color{baselinered}$\bar{G}_{TP}$ & \color{baselinered}\va{0.5370}{0.0265} & \color{baselinered}\va{0.5862}{0.0348} & \color{baselinered}\va{0.8019}{0.0362} & \color{baselinered}\va{0.5026}{0.0037} & \color{baselinered}\va{0.4930}{0.0333} & \color{baselinered}\va{0.6277}{0.2206} & \color{baselinered}\best{\va{0.7148}{0.2438}} & \color{baselinered}0.5087/0.5000 & \color{baselinered}0.5370/0.5000 \\
 & \color{baselinedarkred}Const = 1 & \color{baselinedarkred}\va{0.5000}{0.0000} & \color{baselinedarkred}\va{0.5000}{0.0000} & \color{baselinedarkred}\va{0.5000}{0.0000} & \color{baselinedarkred}\va{0.5000}{0.0000} & \color{baselinedarkred}\va{0.5000}{0.0000} & \color{baselinedarkred}\va{0.5000}{0.0000} & \color{baselinedarkred}\va{0.5000}{0.5000} & \color{baselinedarkred}0.5000/0.5000 & \color{baselinedarkred}0.5000/0.5000 \\
 & \color{baselinedarkred}Const = 0 & \color{baselinedarkred}\va{0.5000}{0.0000} & \color{baselinedarkred}\va{0.5000}{0.0000} & \color{baselinedarkred}\va{0.5000}{0.0000} & \color{baselinedarkred}\va{0.5000}{0.0000} & \color{baselinedarkred}\va{0.5000}{0.0000} & \color{baselinedarkred}\va{0.5000}{0.5000} & \color{baselinedarkred}\va{0.5000}{0.5000} & \color{baselinedarkred}0.5000/0.5000 & \color{baselinedarkred}0.5000/0.5000 \\
 & \color{baselinedarkred}Rand.\ 0.5 & \color{baselinedarkred}\va{0.5018}{0.0304} & \color{baselinedarkred}\va{0.5009}{0.0430} & \color{baselinedarkred}\va{0.5046}{0.0314} & \color{baselinedarkred}\va{0.5112}{0.0367} & \color{baselinedarkred}\va{0.5223}{0.0015} & \color{baselinedarkred}\va{0.5387}{0.0304} & \color{baselinedarkred}\va{0.4821}{0.0046} & \color{baselinedarkred}0.5201/0.5375 & \color{baselinedarkred}0.5078/0.5185 \\
 & \color{baselinedarkred}Rand.\ major & \color{baselinedarkred}\va{0.4935}{0.0320} & \color{baselinedarkred}\va{0.5099}{0.0329} & \color{baselinedarkred}\va{0.5108}{0.0221} & \color{baselinedarkred}\va{0.5040}{0.0485} & \color{baselinedarkred}\va{0.4877}{0.0007} & \color{baselinedarkred}\va{0.4965}{0.0084} & \color{baselinedarkred}\va{0.5150}{0.0046} & \color{baselinedarkred}0.4898/0.5099 & \color{baselinedarkred}0.5016/0.4859 \\
\Block{1-11}{\textbf{MLP}} & & & & & & & & & & \\
\Block{1-2}{Best results MLP} & & \va{0.7141}{0.0286} & \va{0.7284}{0.0417} & \va{0.8604}{0.0334} & \va{0.7035}{0.0557} & \va{0.8299}{0.1130} & \va{0.6389}{0.2116} & \va{0.7153}{0.2354} & 0.6148/0.7823 & 0.5309/0.5703 \\
\Block{1-2}{Best indicators combination}
 & &
 \makecell[c]{$\bar{s}, \bar{e}, std(s),$ \\ $\mathbf{E_{p}}, \mathbf{PP_{p}},$ \\ $PP_{s}, \mathbf{\bar{G}_p},$ \\ $\bar{G}_s, \bar{G}_e$}
 &
 \makecell[c]{$\bar{s}, \bar{e}, std(s),$ \\ $\mathbf{E_{p}}, E_{s},$ \\ $PP_{s}, \bar{G}_s, \bar{G}_e$}
 &
 \makecell[c]{$\mathbf{\bar{p}}, \mathbf{std(p)},$ \\ $\bar{e}, std(e),$ \\ $\bar{s}, E_{s},$ \\ $\mathbf{\bar{G}_p}, \bar{G}_s$}
 &
 \makecell[c]{$\bar{s}, std(e), \mathbf{E_{p}},$ \\ $PP_{s}, \mathbf{\bar{G}_p}, \bar{G}_s$}
 &
 \makecell[c]{$\mathbf{\bar{p}}, \bar{s}, \mathbf{std(p)},$ \\ $std(s), std(e),$ \\ $\mathbf{PCC}, \mathbf{E_{p}}, E_{s},$ \\ $PP_{s}, \mathbf{\bar{G}_p},$ \\ $\bar{G}_s, \bar{G}_e$}
 &
 \makecell[c]{$\bar{s}, \bar{e},$ \\ $\mathbf{PCC}, \mathbf{SCC},$ \\ $\mathbf{E_{p}}, \mathbf{PP_{p}}, \bar{G}_e$}
 &
 \makecell[c]{$\mathbf{\bar{p}}, \mathbf{std(p)},$ \\ $std(s), \mathbf{E_{p}},$ \\ $E_{s}, \mathbf{PP_{p}}, PP_{s}, $ \\ $ \mathbf{\bar{G}_p}, \bar{G}_s, \bar{G}_e$}
 &
 \makecell[c]{$\bar{s}, E_{s},$ \\ $\mathbf{\bar{G}_p}, \bar{G}_e$}
 &
 \makecell[c]{$\mathbf{std(p)},$ \\ $\mathbf{PP_{p}}, PP_{s}$}
 \\
\end{NiceTabular}}
\end{table*}

Experiments demonstrating transferability of our trustworthiness detectors across different domains and models are provided in Appendix~\ref{subset:indicators_transfer}.

\section{Conclusion}

In this work we considered the problem of assessing the trustworthiness of outputs of generative AI models. The major challenge of this task is that the ground truth data or gold-standard methods for checking the correctness of responses are not always available. Sometimes, it may not even be possible to expect that such information could be provided. To mitigate this issue, we "calibrate" outputs of generative models with that of experts models, representing intended model outputs to given inputs. We also introduced a metric of trust characterising probabilities of the model's correct generations conditional on the generations of expert models or agents. These estimated probabilities (calibrator probabilities) are statically grounded and together with a set of traditional features, after aggregations form a set of $15$ indicators that may constitute a new basis for AI reliability evaluation.
Our results demonstrate that classifiers trained on these $15$ indicators can discriminate correct and incorrect answers. Our experiments illustrate this across three ontologies: the BF programming language, math problems, and medical question answering. 
In all settings, subsets including probability-based indicators yielded better discrimination than subsets comprising traditional indicators. Moreover, adding probability-based indicators to the traditional ones either improves transferability or leaves it nearly unchanged, both across models within the same ontological class and across ontologies entirely. Where the subsets of only traditional indicators transfer better, the margin is small.
\section{Limitations}
While the proposed framework is theoretically applicable to diverse generative AI modalities -- for instance, by treating image patches as discrete tokens -- the empirical validation in this study was limited to generative language models. 
The observed transferability of the trust measure within and across ontologies suggests that universal patterns underlying hallucinations may exist across language model outputs. However, since the indicators are predicated on probability distributions derived from an ontology-specific calibrator, adapting the classifier to a novel ontology (including alternative language model architectures) necessitates construction of a new calibrator for the target ontology.  Calibrator-based probabilities improve performance only in combination with the traditional indicators (see Appendix~\ref{subset:indicators_transfer}), and the most effective combination may differ across ontologies.
\subsection*{AI use statement}

In this work, we used generative AI tools for text checking on grammatical issues.
We have not used generative AI tools for research ideas generation, experiment setup and run, making conclusions.

\subsection*{Reproducibility statement}

The code for all experiments is provided in our repository~\citep{anonymous2026code}. Section~\ref{sec:methodology} provides a detailed methodological descriptions of our framework, Appendix~\ref{appendix:clf_techincal_details} -- technical details for experiments reproducing.



\bibliography{iclr2027_conference}
\bibliographystyle{iclr2027_conference}

\appendix
\section{Related work}\label{appendix:related2}

In this Section, we review existing work on hallucination detection and mitigation in LLMs, with a particular focus on approaches that motivate or relate to our proposed framework.

\subsection{Consistency-based detection}\label{appendix:consistency}

Consistency-based methods detect hallucinations by identifying contradictions or instability within a model's outputs or across multiple generations. These approaches rely on internal agreement rather than external knowledge, making them applicable in black-box settings. 

{\it Self-contradiction detection} generates paired sentences or internal claims and employs automated checking (via prompting or classification) to identify logical inconsistencies. Constrained prompting techniques elicit comparable sentence pairs that can be systematically evaluated for contradictions~\citep{mundler2023selfcontradictory}. Self-contradiction methods, however, only detect internal contradictions; many factual hallucinations produce no internal contradiction. Moreover, these methods require generating multiple outputs or auxiliary prompts, which increases computational cost and introduces sensitivity to the prompting strategy used to elicit contradictions.

{\it Self-consistency} and {\it Best-of-N sampling} approaches sample multiple outputs (n-best) and measure agreement through majority voting or semantic similarity. Outputs with low agreement across samples are flagged as potentially hallucinated~\citep{zhang2023siren, zhang2025siren}.
These methods are computationally expensive, requiring multiple forward passes; they may  miss consistently repeated hallucinations that appear across all samples.

\subsection{External verification and retrieval-based methods}\label{appendix:external}

External verification methods ground model outputs in retrieved sources or knowledge bases, enabling detection of unsupported claims through comparison with authoritative information.

{\it Retrieval-Augmented Generation (RAG) and grounding}
conditions generation on retrieved documents or knowledge snippets, allowing outputs to be traced to sources. Detection involves comparing generated claims to retrieved evidence or enforcing grounding constraints during decoding~\citep{liang2024thames}.
However,  retrieval noise, stale sources, or biased corpora can propagate errors.  Incorrect alignment between retrieved passages and claims can produce grounded-but-incorrect outputs.

{\it Fact-checking with reward models and rejection sampling}. Fine-grained reward or classifier models are trained on human annotations marking hallucinations. These models score generations, enabling rejection sampling or fine-tuning to prefer non-hallucinatory outputs~\citep{gunjal2023detecting}. The approach requires substantial annotated data, which is costly to obtain.  It inherits annotator biases and coverage limitations. Furthermore, generalization to out-of-distribution ontologies can be limited.

{\it Knowledge base verification and benchmarks} verify model statements by querying structured knowledge bases or curated benchmarks to identify unsupported facts~\citep{li2024dawn}. Their drawbacks include limited recall when knowledge bases lack coverage for the ontology. Also, mapping free-text claims to structured knowledge base queries is non-trivial.

\subsection{Model-based and prompt-based detection}\label{appendix:model}

This category encompasses training separate classifiers to detect hallucinations and using prompt engineering to elicit self-verification from the model itself.

{\it Separate detector models and classifiers}. A classifier or detector model is fine-tuned on annotated examples to predict whether generated content is hallucinated. Detectors can operate at token, span, or full-answer levels~\citep{agarwal2024codemirage, gunjal2023detecting}.
The methods are heavily dataset-dependent and require costly annotation efforts.

{\it Prompted self-verification} and {\it Chain-of-Thought (CoT) checks}. The model is prompted to list supporting evidence, explain its reasoning chain, or directly judge whether a generated claim is supported (self-critic approach)~\citep{mundler2023selfcontradictory, liang2024thames}. Models, however, can produce plausible but fabricated supporting evidence. Self-verification may fail when models lack knowledge to judge their own outputs.

\subsection{Counterfactual perturbations and intervention methods}\label{appendix:counterfactual}

These methods intentionally alter claims or model components to measure sensitivity, using behavioral responses to distinguish hallucinations from robust knowledge.

{\it Counterfactual probing and perturbation sensitivity} generate counterfactual variants of a claim  (introducing subtle factual changes), and the model's confidence or output consistency is measured. Hallucinated content is expected to show inconsistent sensitivity compared to grounded knowledge~\citep{feng2025counterfactual}. The challenge, however, is that  generating high-quality counterfactuals is non-trivial and may require ontology knowledge. False conclusions can arise if counterfactuals are unnatural or easily distinguished. Computational cost increases with the volume of counterfactuals tested.

{\it Component interventions during generation} target specific model components (attention heads, MLP layers, residual streams) to identify causal factors in hallucination generation. Steering pre-hallucination signals can prevent hallucinations, informing detection strategies~\citep{simhi2024constructing}.  Intervention success varies by component and may harm language modeling quality. It also requires extra computational resources. Importantly, findings may not generalize across model architectures.

\subsection{Summary}\label{appendix:summary}

Despite the breadth and depth of the repertoire of methods, approaches and tools assisting the identification of trusted AI generations, several issues remain open. Common limitations across detection approaches include:

\textbf{Dataset and ontology sensitivity:} Detectors and reward models rely on annotated distributions and can fail under ontology shift~\citep{agarwal2024codemirage, simhi2024constructing}.
    
\textbf{Overconfidence problem:} Model confidence is an imperfect proxy for factuality, producing both false negatives (overconfident hallucinations) and false positives (uncertain correct outputs)~\citep{varshney2023stitch}.
    
\textbf{Computational cost:} Many effective methods (ensemble sampling, retrieval, counterfactual generation) impose significant computational overhead~\citep{zhang2023siren, feng2025counterfactual}.

\textbf{Sentence-level detection:} Most methods only support sentence- or response-level hallucination detection, which can jeopardize the perceived reliability of the entire answer even when it is largely correct.

\section{Theoretical justification}
\label{appendix:theorem}

In what follows we provide a theoretical justification of the possibility to learn ontology-calibrated confidence from expert's data. We do so by showing that the probability bounds on threshold-defined acceptance or rejection events, specified by $\Delta_{j}^i$ in Algorithm \ref{alg:confidence}, can indeed be inferred from empirical calibration. In fact these bounds converge to true probabilities as the size of the calibration sets grow. For simplicity, the results are presented in two steps. In Section \ref{appendix:theorem:simple} we show how these bounds can be derived within standard i.i.d. setting. This enables a clear and focused explanation of main theoretical machinery supporting our approach.  In Section \ref{appendix:theorem:general} we demonstrate how the results can be extended to more practical setups in which i.i.d. assumptions could be violated. 

\subsection{Simplified settings}\label{appendix:theorem:simple}

Let elements of the set $\mathcal{S}$ be independent samples  drawn from the same, albeit unknown, distribution. In this case conditional probabilities of confidence judges' $\mathcal{A}^{i}$ acceptance or rejection of AI model's output at thresholds $\Delta_j^i$ could be estimated from empirical data. Moreover, when subsets $\mathcal{S}_{+,j}$ and   $\mathcal{S}_{-,j}$ are sufficiently large, these estimates converge to true conditional probabilities. The results are formally stated in Theorem  \ref{thm:gen_ai_judge}.

\begin{theorem}\label{thm:gen_ai_judge}
Let  measurable maps $G^S, G^E$, feature map $\Phi$, and confidence judges' training set $\mathcal{S}$ be defined as in (\ref{eq:generative_model}) -- (\ref{eq:training_dataset}). Suppose that the elements of $\mathcal{S}$ are independently sampled from the (unknown) data distribution $P_D$, that $\mathcal{S}_{+,j}$ and $\mathcal{S}_{-,j}$ are not empty, and that $G^S, G^E$ and $\Phi$ are chosen independently of $\mathcal{S}$. Pick $\lambda\in(0,\sqrt{16/e})$.

For each class index $j$, select $h_j$ from $\mathcal{H}$ independently of $\mathcal{S}$, and let $\mathcal{A}^i$ be the outputs of Algorithm~\ref{alg:confidence}  with inputs $G^S,\Phi,\mathcal{S}, \{h_j\in\mathcal{H}| j=1,\dots,q\}$ and $\Delta_j^i\in(0,1)$.

\begin{algorithm}[t]
\caption{Confidence Judge}\label{alg:confidence}
\begin{algorithmic}[1]
\Require Generative model $G$, feature map $\Phi$, forming dataset $\mathcal{S}$, projectors $h_j\in\mathcal{H}: \Real^d\rightarrow\Real$, and parameters $\Delta_j^{i}\in(0,1)$, $i=1,\dots,k$, $k\in\mathbb{N}$ defining the judge's confidence grid. 
\For{$j=1$ to $q$}
    \State Compute $\mathcal{S}_{-,j}$, $\mathcal{S}_{+,j}$ and functions $F_{-,j}, F_{+,j}$ (as defined in (\ref{eq:F+-})).
    \State Define the class-wise calibrators
    \[
    \mathcal{A}_j^i(z) = 
    \begin{cases}
       \tau^{\times}, & \text{if } h_j(z) \le F_{-,j}^\dagger(\Delta_j^{i}),  \\
       \tau_j, & \text{otherwise}.
    \end{cases}
    \]
    The map $F_{-,j}^\dagger$ is defined as
    \[
    F_{-,j}^\dagger(z)=\inf\{x\in \mathbb{R} \ : \ F_{-,j}(x)\geq z\}.
    \]
\EndFor
\State Define global calibrators 
\[
\mathcal{A}^{i}(\tau_j,z)=\mathcal{A}_j^{i}(z).
\]
\Ensure calibrator maps $\mathcal{A}^{i}$.
\end{algorithmic}
\end{algorithm}

Then, for a new independent sample $(u, \ell)\sim P_D$ and each $j=1,\dots,q$, conditional on the realized $M_{+,j},M_{-,j}$ the following hold: 

 With probability at  least
\[ \begin{array}{l}
\max\{1-\lambda r(M_{+,j},\lambda),0\}
\end{array}
\]
over the draw of $\mathcal{S}_{+,j}$
\[
\begin{array}{ll}
1 -  F_{+,j}(F_{-,j}^{\dagger}(\Delta_j^i)) +  r(M_{+,j},\lambda)& \geq
 P(\mathcal{A}^i(G^S(u),\Phi(u))=\tau_j \given G^S(u)=\tau_j, \ G^E(u)=\tau_j ) \\ & \geq   1 -  F_{+,j}(F_{-,j}^{\dagger}(\Delta_j^i)) -  r(M_{+,j},\lambda).
 \end{array}\nonumber
\]

Similarly,  with probability  at least 
\[ \begin{array}{l}
\max\{1-\lambda r(M_{-,j},\lambda),0\}
\end{array}
\]
over the draw of $\mathcal{S}_{-,j}$
\begin{equation}
\begin{array}{ll}
F_{-,j}(F^{\dagger}_{-,j}\Delta_j^i) + r(M_{-,j},\lambda)  & \geq 
 P(\mathcal{A}^i(G^S(u),\Phi(u))=\reject  \given G^S(u)= \tau_j, \ G^E(u)\neq \tau_j )\\ &\geq \Delta_j^i - r(M_{-,j},\lambda),  \nonumber
\end{array}
\end{equation}
where
\[
r(x,\lambda)= \left(\frac{1}{4 x} \left[\log(16 x \lambda^{-2}) - \frac{1}{2} \log \log(16 x \lambda^{-2}) \right] \right)^{1/2}.  
\]
\end{theorem}
{\it Proof of Theorem \ref{thm:gen_ai_judge}.} Given that $G^S$, $G^E$ are measurable and elements in $\mathcal{S}$ are independently sampled from the distribution $P_D$, we can conclude that $\mathcal{S}_j$ and $\mathcal{S}_{+,j}$, $\mathcal{S}_{-,j}$ are independent samples from their corresponding conditional distributions $P_D((u,\tau) \given G^S(u)=\tau_j)$, $P_D((u,\tau) \given G^E(u)=\tau_j, \ G^S(u)=\tau_j)$, and $P_D((u,\tau) \given G^E(u)\neq \tau_j, \ G^S(u)=\tau_j)$. 
Moreover, since $h_j\in\mathcal{H}$ is chosen independently of $\mathcal{S}$ and that $h_j$ and $\Phi$ are measurable, we may conclude that elements of 
the multi-sets
\[
\begin{split}
\mathcal{X}_{+,j}=&\{h_j(\Phi(u)), \text{ for each } (u,\tau) \in \mathcal{S}_{+,j}\},\\
\mathcal{X}_{-,j}=&\{h_j(\Phi(u)), \text{ for each } (u,\tau) \in \mathcal{S}_{-,j}\}{,}\\
\end{split}
\]
are independent samples from the respective conditional probability distributions induced on $\Real$ by $h_j$.

To proceed with the proof we will need the following lemma.

\begin{lem}\label{lem:empirical_probability} Let $\theta\in\Real$, $\lambda\in(0,\infty)$, and $\mathcal{X}\subset\mathcal{\Real}$ be a finite multi-set whose elements are sampled independently from some unknown probability distribution. Let $|\mathcal{X}|=n$, and $F_n$ be defined as follows:
\[
F_n(s)=\frac{1}{n}\sum_{x\in\mathcal{X}}\boldsymbol{1}_{x\leq s}.
\]
Let $z$ be a new independent sample from the same distribution and consider the events:
\[
E_1: \ z\leq \theta, \ E_2: \ z> \theta.
\]

Then, conditional on the realized $n$,  with probability  at least 
\[
\begin{array}{l}
\max\{1-\lambda \left(\frac{1}{4n} W_0\left(\frac{16n}{\lambda^2}\right)\right)^{1/2},0\}
\end{array}
\]
over the draw of $\mathcal{X}$
\begin{align}
\begin{array}{l}
   F_n(\theta) +  \left(\frac{1}{4n} W_0\left(\frac{16n}{\lambda^2}\right)\right)^{1/2}\geq  P(E_1)  \geq F_n(\theta) -  \left(\frac{1}{4n} W_0\left(\frac{16n}{\lambda^2}\right)\right)^{1/2}
\end{array}
\label{eq:probability_bounds:E1}
\end{align}
and  
\begin{align}
\begin{array}{l}
 1 - F_n(\theta) +  \left(\frac{1}{4n} W_0\left(\frac{16n}{\lambda^2}\right)\right)^{1/2}\geq P(E_2)
   \geq  1 - F_n(\theta) -  \left(\frac{1}{4n} W_0\left(\frac{16n}{\lambda^2}\right)\right)^{1/2}, \label{eq:probability_bounds:E2}
\end{array}
\end{align}
where $W_0$ is the principal branch of the Lambert $W$ function.
\end{lem}
\vspace{-5pt}
{\it Proof of Lemma \ref{lem:empirical_probability}.}
Let $P_c$ be the cumulative distribution function associated with the unknown probability distribution from which both the multi-set $\mathcal{X}$ and $z$ were independently drawn.
It is clear that 
\[
P(E_1)=P_c(\theta), \qquad P(E_2)=1-P_c(\theta).
\]
Introduce the auxiliary event
\[
E(\varepsilon): \ \sup_{s\in\Real} |F_n(s)-P_c(s)|>\varepsilon,
\]
and its complement 
$\bar{E}(\varepsilon):\operatorname{not} E(\varepsilon)$. The Dvoretzky-Kiefer-Wolfowitz inequality~\citep{DKW,Massart} states that, for any $\varepsilon > 0$,
\begin{equation}\label{eq:KDW}
P\left(\sup_{s\in\Real}|F_n(s)-P_c(s)|>\varepsilon \right) \leq 2 \exp(-2 n \varepsilon^2),
\end{equation}
and therefore,  event $E(\varepsilon)$ occurs with probability at most $2 \exp(-2 n \varepsilon^2)$ over the draw of $\mathcal{X}$. Similarly, event $\bar{E}(\varepsilon)$ occurs with probability at least $1 - 2 \exp(-2 n \varepsilon^2)$ over the draw of $\mathcal{X}$.

Now, consider the following equation
\[
\lambda \varepsilon = 2\exp(-2 \varepsilon^2 n).
\]
Its solution $\varepsilon^\ast$ is 
\[
\varepsilon^\ast = \left(\frac{1}{4n} W_0\left(\frac{16 n}{\lambda^2}\right)\right) > 0.
\]
Given that the Dvoretzky-Kiefer-Wolfowitz inequality holds true for any $\varepsilon>0$, it is certainly true for $\varepsilon^\ast$. Substituting $\varepsilon^\ast$ into the bounds and recalling that $\lambda \varepsilon^\ast = 2\exp(-2 {\varepsilon^\ast}^2 n)$ completes the proof.  \hfill $\square$

\textit{Proof of Theorem~\ref{thm:gen_ai_judge} continued.}
By the construction of $\mathcal{A}^i$ in Algorithm~\ref{alg:confidence}, the probability that the confidence judge correctly accepts the label assigned by the underlying AI system is:
\begin{align}
&P(\mathcal{A}^i(G^S(u),\Phi(u))=\tau_j  \given G^S(u)=\tau_j, \ G^E(u)=\tau_j)
 \label{eq:proxy_probability_true_positive}
\\&
=
P(h_j(\Phi(u))>F^{\dagger}_{-,j}(\Delta_j^i) \given G^S(u)=\tau_j, \ G^E(u)=\tau_j). \notag
\end{align}
 Since $F_{+,j}$, as defined in~\eqref{eq:F+-}, is  empirical cumulative distribution function of the conditional distribution $P(h_j(\Phi(u)) \given G^S(u)=\tau_j, \  G^E(u)=\tau_j)$, bounds  ~\eqref{eq:probability_bounds:E2} for event $E_2$ from Lemma \ref{lem:empirical_probability} apply (with $\theta = F^{\dagger}_{-,j}(\Delta_j^i)$, $F_n=F_{+,j}$, $n=M_{+,j}$).


Turning to the probability of the confidence judge $\mathcal{A}^i$ correctly rejecting an incorrect label assigned by the underlying AI system, we observe that
\begin{align*}
&P(\mathcal{A}^i(G^S(u),\Phi(u))=\reject  \given G^S(u)=\tau_j, \ G^E(u)\neq\tau_j)
 \\&\quad
=
P(h_j(\Phi(u))\leq F^{\dagger}_{-,j}(\Delta_j^i)  \given G^S(u)=\tau_j, \ G^E(u)\neq\tau_j).
\end{align*}
Similarly to the previous case, we note that the conditional distribution  $P(h(\Phi(u)) \given G^S(u)=\tau_j, \ G^E(u)\neq\tau_j)$ admits a cumulative distribution function with empirical approximation $F_{-,j}$ which is used by Algorithm \ref{alg:confidence}.
Therefore bounds ~\eqref{eq:probability_bounds:E1},  for the event $E_1$ from Lemma \ref{lem:empirical_probability} apply (with $\theta = F^{\dagger}_{-,j}(\Delta_j^i)$, $F_n=F_{-,j}$, $n=M_{-,j}$, and observing that $F_{-,j}(F_{-,j}^{\dagger}(\Delta_j^i))\geq \Delta_j^i$).

Finally, we note that for any $x\geq e$, the principal branch $W_0$ of the Lambert $W$ function satisfies \citep{hoorfar2008inequalities}
\[
\log(x)- \log\log (x) \leq W_0(x) \leq \log(x)-\frac{1}{2}\log \log(x).
\]
Combining this with bound stemming from Lemma \ref{lem:empirical_probability} completes the proof.\hfill $\square$

\subsection{Generalization beyond i.i.d. assumptions}\label{appendix:theorem:general}

Theorem \ref{thm:gen_ai_judge} explicitly assumes that 1) elements of the set $\mathcal{S}$ are all drawn independently from the same distribution, and 2) a new measurement is sampled independently from the same distribution. Here we show how these results could be  generalised to dependent samples.

In order to do so, we need to introduce notion enabling quantification of various degrees of dependencies. For LLMs the most relevant  framework concerns dependencies in sequences and sequential data. Therefore,  consider $(x_i)_{i \in \Z}$,  a strictly stationary sequence of real-valued random
variables defined on a common probability space $(\Omega, \mathcal{B}, P)$.
For integers $k \in \Z$, let
\[
\mathcal{F}_{-\infty}^{k} \;=\; \sigma(x_j : j \le k),
\qquad
\mathcal{F}_{k}^{\infty} \;=\; \sigma(x_j : j \ge k)
\]
denote the $\sigma$-algebras of the past up to time $k$ and the future from time
$k$ onward, respectively.
 
\begin{definition}[Strong-mixing coefficient]\label{def:alpha-mixing}
The strong-mixing (or $\alpha$-mixing) coefficient at lag $n \ge 1$ is
\[
\alpha(n) \;=\; \sup_{k \in \Z}
\sup_{\substack{A \in \mathcal{F}_{-\infty}^{k} \\ B \in \mathcal{F}_{k+n}^{\infty}}}
\bigl| P (A \cap B) - P(A)\,P(B) \bigr|.
\]
The sequence $(x_i)$ is said to be \emph{$\alpha$-mixing} (or
\emph{strongly mixing}, in the sense of Rosenblatt \citep{rosenblatt1956central}) if
$\alpha(n) \to 0$ as $n \to \infty$. 

We say that the mixing is
\emph{geometric} (or that the sequence is geometrically $\alpha$-mixing) if
there exists $\lambda > 0$ such that $\alpha(n) = O(e^{-\lambda n})$ as
$n \to \infty$.

The mixing is polynomial of order $p$ if there exist constants
$K_0 > 0$ and $p > 1$ such that
\begin{equation}\label{eq:poly-mixing}
\alpha(n) \;\le\; K_0\, n^{-p} \qquad \text{for all } n \ge 1.
\end{equation}
\end{definition}
 
The $\alpha$-mixing coefficient measures essentially how far the probability $P (A \cap B)$  of the joint event  $A \cap B$ deviates from the product $P(A)\,P(B)$. 

We note that for stationary sequences the supremum over $k$ in Definition \ref{def:alpha-mixing} is redundant and may be replaced by any single value of $k$. We also note that the motion of mixing presented in Definition \ref{def:alpha-mixing} is the least constrained amongst other classical mixing
notions. Indeed it is implied by $\beta$-mixing (absolute regularity) and $\phi$-mixing (uniform mixing), with $2\alpha(n) \le \beta(n) \le \phi(n)$. See
\citep{bradley2005basic} for details, Eqns. (1.11)--(1.18)).

At the heart of the proof of Theorem \ref{thm:gen_ai_judge}, and the sole place where independence of samples is required,  is Lemma \ref{lem:empirical_probability}. There the independence is needed to appropriately invoke the Dvoretzky–Kiefer–Wolfowitz inequality~\citep{DKW,Massart}. Hence if a suitable replacement to the classical Dvoretzky–Kiefer–Wolfowitz inequality is used that does not require independence, then Lemma \ref{lem:empirical_probability} and Theorem \ref{thm:gen_ai_judge} could be generalized to dependent settings too.

\subsubsection{Geometric mixing}

The replacement follows immediately from \citep{naaman2021tight}.

\begin{lem}[Naaman 2021, Lemma 4.3, $k = 1$]\label{thm:naaman-dkw}
Let $(x_i)$ be a stationary $\alpha$-mixing sequence of real-valued random
variables with marginal CDF $F$, and suppose
$\alpha(n) = O(e^{-\lambda n})$ for some $\lambda > 0$. Let $(\delta_n)$ be a
non-decreasing positive integer sequence satisfying
\begin{equation}\label{eq:delta-conditions}
\delta_n \to \infty
\qquad \text{and} \qquad
\frac{\sqrt{n}}{\delta_n} \to \infty
\qquad \text{as } n \to \infty.
\end{equation}
Then for every $t \ge 0$ and all $n$ sufficiently large,
\begin{equation}\label{eq:naaman-dkw}
P\!\left(
\sup_{\theta \in \Real}
\biggl|
\sum_{i = \delta_n}^{n + \delta_n - 1}
\Bigl( \indic{x_i \le \theta} - F(\theta) \Bigr)
\biggr|
\;\ge\; n t
\right)
\;\le\;
(n + \delta_n)^{2}
\exp\!\left(
- \frac{n\, t^2}{32\, \delta_n^{2}\, (1 + \ln n)}
\right).
\end{equation}
\end{lem}
 
The sequence $\delta_n$ trades off the polynomial pre-exponential factor
against the rate inside the exponent in (\ref{eq:naaman-dkw}). With the choice
$\delta_n = \lfloor n^{1/p} \rfloor$ for any fixed $p > 2$, the conditions
in~\eqref{eq:delta-conditions} are satisfied, and the exponent
in~\eqref{eq:naaman-dkw} becomes
\[
\frac{n\, t^2}{32\, \delta_n^{2}(1 + \ln n)}
\;=\;
\frac{n^{1 - 2/p}\, t^2}{32 (1 + \ln n)},
\]
while the prefactor $(n + \delta_n)^{2}$ remains $O(n^{2})$. As $p \to \infty$,
the exponential rate approaches $n / \log n$, almost matching (up to constants and
the polynomial prefactor) the exponential rate of the classical Dvoretzky–Kiefer–Wolfowitz inequality.

 \subsubsection{Polynomial mixing}

When the mixing is polynomial, the dependencies have longer tails and convergence of empirical estimates of the cumulative distribution function to the true one may no longer follow exponential law. Yet, for large enough data samples the convergence still occurs, as is shown in the theorem below. 
  
\begin{theorem}\label{thm:poly-DKW}
Let $(x_i)$ be strictly stationary and $\alpha$-mixing with polynomial rate  
\eqref{eq:poly-mixing} for some $p>1$ and $K_0>0$. For every
$q\in(2,\,p+1)$ there exists a constant $C=C(q,p,K_0)<\infty$ such that
for every $n\ge 2$ and every $t\in(0,1]$,
\begin{equation}\label{eq:poly-DKW}
P\!\left(\sup_{\theta\in\Real}\frac{1}{n}\bigl|\sum_{i = 1}^{n}
\Bigl( \indic{x_i \le \theta})-F(\theta)\bigr|>t
\right)
\;\le\;
\frac{C}{n^{\,q/2}\,t^{q+1}}.
\end{equation}
\end{theorem}

{\it Proof of Theorem \ref{thm:poly-DKW}}. The proof proceeds in three steps: a moment bound for partial sums via
Yokoyama's inequality (step 1), a pointwise tail via Markov inequality (step 2), followed by a standard deterministic
grid reduction (step 3).

Recall that
 \[
 F_n(\theta)=\frac{1}{n} \sum_{i=1}^n\indic{x_i\le\theta}.
\]
In what follows we will use the standard notation for the \emph{left limit} of  a non-decreasing function $g:\R\to\R$ at  $x\in\R$: 
\[
g(x^-)\;:=\;\lim_{\varepsilon\downarrow 0} g(x-\varepsilon)
\;=\;\lim_{y\uparrow x} g(y).
\]
It is clear that
\[
F(x^-) \;=\; P(x_1<x), \qquad
 F_n(x^-) \;=\; \frac{1}{n}\,\bigl|\{i:x_i<x\}\bigr|,
\]
with the gap $F(x)-F(x^-)=P(x_1=x)$ equal to the atom mass of $F$
at $x$ (which vanishes when $F$ is continuous at $x$). Both $F$ and
$F_n$ are right-continuous and non-decreasing, so all left
limits exist and lie below the corresponding values:
$F(x^-)\le F(x)$ and $F_n(x^-)\le F_n(x)$.

{\it Step 1 (Accounting for dependencies via the Yokoyama's moments inequality)}. Fix $\theta\in\Real$ and define two centred indicator processes
\[
w_i(\theta) \;=\; \indic{x_i\le\theta}-F(\theta), \quad w_i^{-}(\theta) \;=\; \indic{x_i<\theta}-F(\theta^{-}), \qquad i\in\Z.
\]
Observe that $\E (w_i(\theta))=0$, $\E(w_i^{-}(\theta))=0$ and $|w_i(\theta)|\le 1$, $|w_i^{-}(\theta)|\leq 1$. That is, both sequences are bounded by $1$ in absolute value and centred.  Moreover, they are measurable functions of $x_i$ and as such inherit the
$\alpha$-mixing of $(x_i)$ with coefficients no larger than
$\alpha(\cdot)$. To see this, note first that the maps
$\phi(y)=\indic{y\le\theta}-F(\theta)$ and
$\phi^-(y)=\indic{y<\theta}-F(\theta^-)$ are Borel-measurable functions
$\R\to\R$: they are indicators of the Borel sets $(-\infty,\theta]$
and $(-\infty,\theta)$ respectively. By the
measurability of compositions
\cite[ Theorem~13.1, p. 182]{Billingsley1995}, $\phi(x_i)=w_i(\theta)$
and $\phi^-(x_i)=w_i^-(\theta)$ are themselves measurable
random variables on the underlying probability spaces
$(\Omega,\mathcal{F}^k_{-
\infty},P)$, $(\Omega,\mathcal{F}^\infty_{k+n},P)$ (see Definition \ref{def:alpha-mixing}).

Let
$\mathcal{B}_\R$ denote the Borel $\sigma$-field on $\R$. Recall that the
$\sigma$-field generated by a random variable $X:\Omega\to\R$ on
$(\Omega,\mathcal{B},P)$ is the $\sigma$-field of preimages
$\sigma(X)=X^{-1}(\mathcal{B}_\R)=\{X^{-1}(B):B\in\mathcal{B}_\R\}
\subseteq\mathcal{B}$. For $\psi\in\{\phi,\phi^-\}$, and $\mathcal{B}\in\{\mathcal{F}^k_{-
\infty},\mathcal{F}^k_{-
\infty}\}$
\[
\sigma(\psi(x_i))
\;=\;(\psi\circ x_i)^{-1}(\mathcal{B}_\R)
\;=\;x_i^{-1}\bigl(\psi^{-1}(\mathcal{B}_\R)\bigr)
\;\subseteq\;x_i^{-1}(\mathcal{B}_\R)\;=\;\sigma(x_i),
\]
where $\sigma$ stands for the corresponding sigma-fields, and the inclusion uses
$\psi^{-1}(\mathcal{B}_\R)\subseteq\mathcal{B}_\R$, i.e.\ Borel-measurability
of $\psi$. Therefore, all three $\sigma$-fields $\sigma(x_i)$,
$\sigma(w_i(\theta))$ and $\sigma(w_i^-(\theta))$ are sub-$\sigma$-fields of
the same ambient $\mathcal{B}$. Hence the $\sigma$-field monotonicity of the
mixing coefficient in Definition \ref{def:alpha-mixing} implies that the same mixing coefficient is inherited by $\omega_i(\theta), \omega_i^{-}(\theta)$.

 
Observe that both $\omega_i(\theta)$ and $\omega_i(\theta^{-})$ are bounded. This implies that their $q$-th moments, $q\in\Natural$ are all defined and boun dned. By Yokoyama's moment inequality
\cite[Theorem~1]{yokoyama1980moment},  for every $q>2$ such that
\begin{equation}\label{eq:yokoyama-condition}
\sum_{k=1}^{\infty}(k+1)^{q-2}\,\alpha(k) \;<\; \infty, 
\end{equation}
there exists a constant $A_q<\infty$, depending only on $q$ and on the
sum in \eqref{eq:yokoyama-condition}, such that
\begin{equation}\label{eq:yokoyama}
\E\!\left|\sum_{i=1}^{n} w_i(\theta)\right|^{\!q}
\;\le\;
A_q\, n^{q/2}, \qquad  \E\!\left|\sum_{i=1}^{n} w_i^{-}(\theta)\right|^{\!q}
\;\le\;
A_q\, n^{q/2}, \quad n\ge 1.
\end{equation}
Under polynomial mixing \eqref{eq:poly-mixing}, condition
\eqref{eq:yokoyama-condition} reduces to
$\sum_k (k+1)^{q-2-p}<\infty$, i.e.\ $q-2-p<-1$, equivalently $q<p+1$.

{\it Step 2 (point-wise tail bounds)}. Set $S_n(\theta)=\sum_{i=1}^{n} w_i(\theta)$. According to Markov's inequality and
\eqref{eq:yokoyama}, we obtain that for any $t>0$,
\begin{equation}\label{eq:pointwise-tail}
\begin{split}P \bigl(|F_n(\theta)-F(\theta)|\ge t\bigr)
&=
P\bigl(|S_n(\theta)|\ge nt\bigr)
\;\le\;
\frac{\E|S_n(\theta)|^{q}}{(nt)^{q}}
\;\le\;
\frac{A_q}{n^{q/2}\,t^{q}}, \\
P \bigl(|F_n(\theta^{-})-F(\theta^{-})|\ge t\bigr)
&\leq \frac{A_q}{n^{q/2}\,t^{q}}.
\end{split}
\end{equation}

{\it Step 3 (Bounding the probability of supremum over $\R$).} 
For an integer $N\ge 1$, define the grid via the standard quantile
function:
\begin{equation}
\theta_0=-\infty,\qquad
\theta_N=+\infty,\qquad
\theta_i=\inf\{x\in\R:F(x)\ge i/N\},\quad i=1,\ldots,N-1. \nonumber
\end{equation}
Note, that the grid points $\theta_i$, $i=1,\dots,N-1$ satisfy $\theta_{i-1}\leq\theta_i$, but they are not necessarily  all distinct (consider e.g. distributions comprising a single atom  containing all probability mass). 

Right-continuity of $F$, combined with the definition of the infimum,
yields 
\begin{equation}\label{eq:right-cont}
F(\theta_i^-)\;\le\;\frac{i}{N}\;\le\;F(\theta_i),
\qquad i=1,\ldots,N-1.
\end{equation}
For any $\theta\in[\theta_{i-1},\theta_i)$, monotonicity of
$F_n$ and $F$ ensures that
\[ F_n(\theta_{i-1})\;\le\;F_n(\theta)\;\le\;
F_n(\theta_i^-),
\qquad
F(\theta_{i-1})\;\le\;F(\theta)\;\le\;F(\theta_i^-),
\]
and from \eqref{eq:right-cont} the oscillation of $F$ on this interval
is bounded by
\[
F(\theta_i^-)-F(\theta_{i-1})\;\le\;\frac{i}{N}-\frac{i-1}{N}
\;=\;\frac{1}{N}.
\]
Combining,
\begin{equation}
\bigl|F_n(\theta)-F(\theta)\bigr|
\;\le\;
\max\!\Bigl(\bigl|F_n(\theta_{i-1})-F(\theta_{i-1})\bigr|,
\,\bigl| F_n(\theta_i^-)-F(\theta_i^-)\bigr|\Bigr)
\,+\,\frac{1}{N}.\nonumber
\end{equation}

Since $F$ is the cumulative density function of a probability measure, its atom set
$\{x:F(x)-F(x^-)>0\}$ is at most countable: for each $k\ge 1$, atoms of
mass $\ge 1/k$ are at most $k$ in number, and the atom set is the
countable union over $k$ of these finite sets. The function
$|F_n-F|$ is c\`adl\`ag, and its discontinuities are contained
in the (at most countable) union of the atom set of $F$ and the data
points $\{x_1,\ldots,x_n\}$. Consequently
$\sup_{\theta\in\R}|F_n(\theta)-F(\theta)|$ equals its
supremum over a countable set and is a well-defined measurable random
variable.

Taking the supremum over $\theta\in\R$,
\begin{equation}\label{eq:sup-grid}
\sup_{\theta}\bigl|F_n(\theta)-F(\theta)\bigr|
\;\le\;
\max_{1\le i\le N}\!\max\!\Bigl(
\bigl| F_n(\theta_{i-1})-F(\theta_{i-1})\bigr|,
\,\bigl| F_n(\theta_i^-)-F(\theta_i^-)\bigr|\Bigr)
\,+\,\frac{1}{N},
\end{equation}
since at $\theta=\theta_0=-\infty$ and $\theta=\theta_N=+\infty$ both
$ F_n$ and $F$ vanish or equal $1$ identically.

Fix $t\in(0,1]$ and choose $N=\lceil(q+1)/t\rceil$. Since $q>2$ and
$t\le 1$, $N\ge q+1\ge 3$. Then $1/N\le t/(q+1)$, so
$t-1/N\ge qt/(q+1)$, and $N-1\le (q+1)/t$. 

By \eqref{eq:sup-grid}, the
event $\bigl\{\sup_\theta|F_n-F|>t\bigr\}$ is contained in
the union, over $i\in\{1,\ldots,N\}$, of the events
$\bigl\{|F_n(\theta_{i-1})-F(\theta_{i-1})|>qt/(q+1)\bigr\}$
and
$\bigl\{|F_n(\theta_i^-)-F(\theta_i^-)|>qt/(q+1)\bigr\}$.

Two of these events occur with probability $0$ (at $\theta_0=-\infty$ and at $\theta_N=\infty$). Boole's
inequality combined with \eqref{eq:pointwise-tail} (each applied with
$qt/(q+1)$ in place of $t$) results in
\begin{align*}
P\!\left(\sup_\theta| F_n-F|>t\right)
&\le\;
2(N-1)\cdot\frac{A_q}{n^{q/2}\,(qt/(q+1))^{q}}
\;=\;
\frac{2(N-1)\,(q+1)^{q}\,A_q}{q^{q}\,n^{q/2}\,t^{q}} \\
&\le\;
\frac{2\,\bigl((q+1)/t\bigr)\,(q+1)^{q}\,A_q}{q^{q}\,n^{q/2}\,t^{q}}
\qquad\bigl(\text{using } N-1\le(q+1)/t\bigr)\\
&=\;
\frac{2\,(q+1)^{q+1}\,A_q}{q^{q}\,n^{q/2}\,t^{q+1}}.
\end{align*}

Noticing that $(1+q)^{q+1}/q^q\leq (q+1) e$,  setting $C(q,p,K_0):=2e (q+1) A_q$ completes the proof, since $A_q$
depends only on $q$ and on the mixing-tail sum, which under
\eqref{eq:poly-mixing} is bounded by a quantity depending only on $p$
and $K_0$.\qed

According to Theorem \ref{thm:poly-DKW}, generalization of Theorem \ref{thm:gen_ai_judge} to dependent data with dependencies conforming with polynomial mixing, could be achieved by replacing exponential tails with the ones in the rhs of (\ref{eq:poly-DKW}). Doing so will loosen then confidence bounds. Yet, the overall principle of learnable inference of relevant probabilities would remain.

\section{Details of building AI ontology calibrators}

In the following sections, we provide details on the ontologies for which AI calibrators were built.

\subsection{Ontology 1: BF}

\subsubsection{BF programs-outputs generation}
\label{appendix:bf_programs_generation_details}

Each BF program consists of commands represented by one of eight symbols, detailed in Table~\ref{tab:bf_commands}.

\begin{table}[b]
  \caption{BF commands' specifications. Our model uses all commands except input (\texttt{,}) and predicts an end-of-sequence token for program termination.}
  \label{tab:bf_commands}
  \centering
  \begin{tabular}{c|l}
    \toprule
    Symbol(s) & Command description \\
    \midrule
    \texttt{>/<} & Increment\texttt{/}decrement the data pointer by one. \\
    \texttt{+/-} & Increment\texttt{/}decrement the byte at the data pointer by one. \\
    \texttt{.} & Output the byte at the data pointer.\\
    \texttt{,} & Accept one byte of input, storing its value in the byte at the data pointer. \\
    \texttt{[/]} & If the byte at the data pointer is zero\texttt{/}non-zero, then instead of moving \\
         & the instruction pointer forward to the next command, jump it forward to \\ 
         & the command after the matching \texttt{]/[} command.\\
    \bottomrule
  \end{tabular}
\end{table}

The initial training data was generated using a genetic algorithm that evolves programs based on their output. This approach enables the generation of diverse program candidates.

We also implement a custom BF interpreter with the following specifications:
\begin{enumerate}
    \item Program outputs are represented as ASCII codes (values $0 - 255$). The interpreter truncates output sequences exceeding $256$ codes.
    \item All output values including zeros produced during execution are preserved in the output sequence.
    \item Programs entering infinite loops during execution are terminated, and a special token \texttt{<InfLoop>} is outputted.
\end{enumerate}

\subsubsection{BF programs prediction}
\paragraph{Model's training and the calibrator application details}
\label{appendix:tr_train_details}

For model training, we tokenize BF programs by mapping each symbol to a digit and introducing special tokens: beginning-of-sequence ($\texttt{BOS}=1$), end-of-sequence ($\texttt{EOS}=2$), and padding ($\texttt{PAD}=0$). Our programs do not contain the symbol ","; however, our model is also expected to generate \texttt{EOS} for each sequence. All programs are padded to length $18$ using the $\texttt{PAD}$ token.

Our Transformer architecture comprises a $12$-layer encoder followed by two linear projection layers with interleaved dropout and layer normalization (applied twice), and a $12$-layer decoder followed by two linear layers with three dropout-normalization stages. Sinusoidal positional encodings are applied to embeddings of both program tokens and output symbols. The training objective combines token-level cross-entropy loss with a set-level term that aggregates probability mass across all valid programs producing identical outputs. We train our Transformer model autoregressively.

\begin{figure}[tb]
\centering
\includegraphics[width=0.7\linewidth]{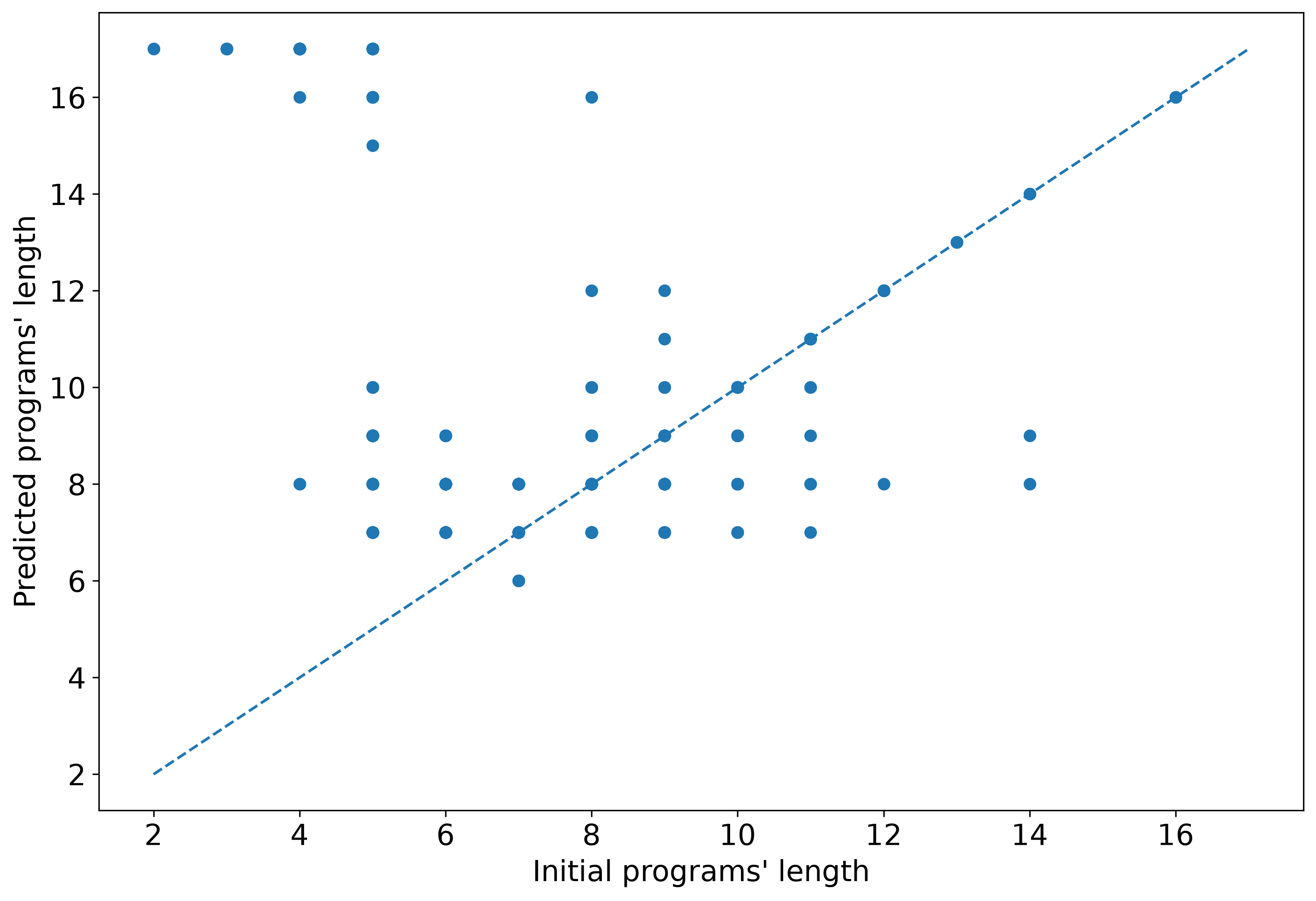}
\caption{The dependence of initial and predicted programs lengths of new BF programs (that differ from the training data). Among $1,490$ new programs, $198$ programs are shorter and $1,061$ are longer than in the initial data, and $231$ programs of the same length as in the initial data.}
\label{fig:init_pred_programs_lens}
\end{figure}

We apply the calibrator to the entire dataset and analyse results for the first $10,000$ program-output pairs. In $2,328$ cases, the model generates programs \textit{identical} to those in the training data. Surprisingly, in $1,490$ cases, it produces functionally equivalent \textit{new} programs that yield identical outputs to the original training programs. We count these as correct predictions since they satisfy the input-output specification. A length comparison of these novel programs is presented in Figure~\ref{fig:init_pred_programs_lens}.

As a result, we obtain $3,818$ correctly generated programs and $6,182$ incorrect ones. 

\paragraph{Predicted programs analysis}
\label{appendix:tr_programs_analysis}

As noted above, our trained Transformer model generates programs that differ from training examples while producing identical outputs. Figure~\ref{fig:init_pred_programs_lens} shows the length relationship between initial and predicted programs for these $1,490$ cases. Among them, $1,061$ programs ($71.2\%$) exceed their training counterparts in length (by up to $15$ symbols). Notably, we also observe $231$ structurally distinct programs of identical length ($15.5\%$), and $198$ shorter programs ($13.3\%$) reduced by a maximum of $6$ symbols.

\begin{figure*}[tb]
\centering
\begin{subfigure}[b]{0.7\textwidth}
    \includegraphics[width=\textwidth]{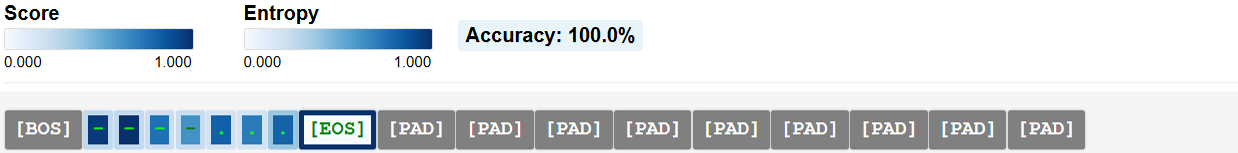}
    \caption{The example of correctly predicted program. Confidence scores and entropy correlate.}
    \label{fig:prob_entr_viz_1}
\end{subfigure}%
\vfill\begin{subfigure}[b]{0.7\textwidth}
    \includegraphics[width=\textwidth]{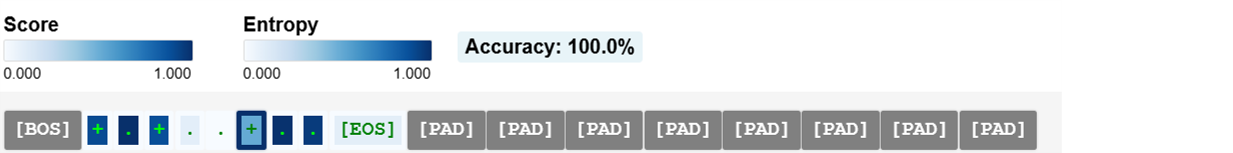}
    \caption{The example of correctly predicted program. Confidence scores and entropy do not correlate.}
    \label{fig:prob_entr_viz_1_2}
\end{subfigure}%
\vfill
\begin{subfigure}[b]{0.7\textwidth}
    \includegraphics[width=\textwidth]{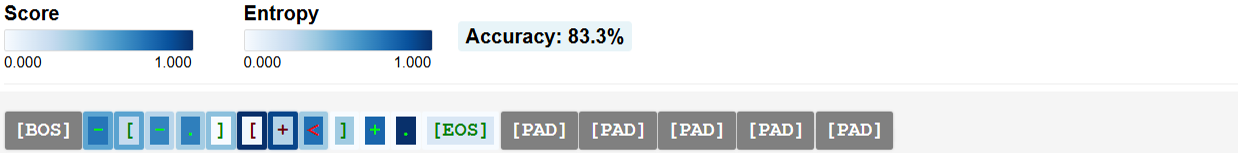}
    \caption{The example of incorrectly predicted program: the initial program is \texttt{-[-.][+<]+.} while the predicted program is \texttt{-[-.]+]]]+.}. Confidence scores and entropy do not correlate.}
    \label{fig:prob_entr_viz_2}
\end{subfigure}
\vfill
\begin{subfigure}[b]{0.7\textwidth}
    \includegraphics[width=\textwidth]{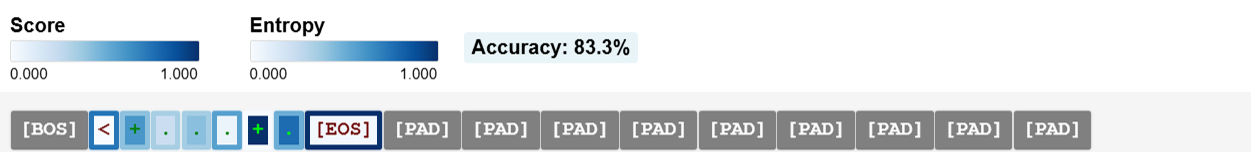}
    \caption{The example of incorrectly predicted program: the initial program is \texttt{<+...+.} while the predicted program is \texttt{++...+.-}. Confidence scores and entropy correlate.}
    \label{fig:prob_entr_viz_3}
\end{subfigure}
\caption{A symbol-level analysis of model confidence and uncertainty (entropy) for program predictions. Each token is visualized with triple encoding: (1) symbols’ foreground colour indicates prediction correctness (green for correct, red for incorrect symbols), (2) background intensity represents confidence scores, and (3) the border colour encodes predictive entropy.}
\label{fig:prob_entr_viz_all}
\end{figure*}

We further analyse the relationship between entropy and the model's internal statistics. First, we visualise entropy and the calibrator's probability distributions for presented 
programs. Figure~\ref{fig:prob_entr_viz_all} shows examples for both correctly and incorrectly predicted programs, demonstrating that confidence and entropy may exhibit an inverse linear relationship (e.g., Figures~\ref{fig:prob_entr_viz_1} and~\ref{fig:prob_entr_viz_3}) or no consistent correlation (e.g., Figures~\ref{fig:prob_entr_viz_1_2} and~\ref{fig:prob_entr_viz_2}).

\begin{figure}[tb]
\centering
\includegraphics[width=0.7\linewidth]{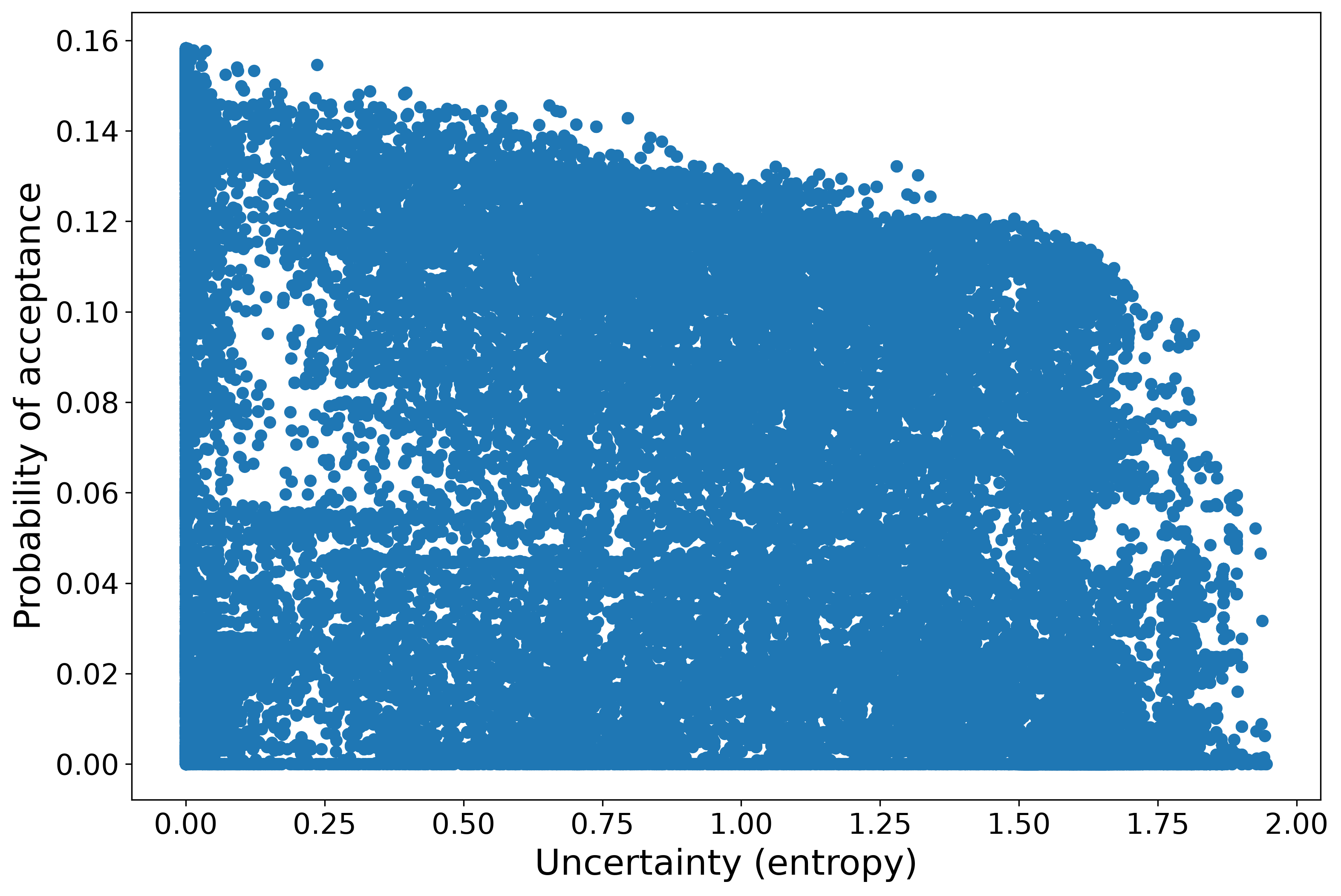}
\caption{The dependence of the probability acceptance on the entropy. The Pearson correlation equals to $-0.37$ with p-value $<0.0001$.}
\label{fig:entr_prob_dependence}
\end{figure}

Second, we compute the Pearson correlation coefficient between entropy and the calibrator's probability across $10,000$ program samples. Figure~\ref{fig:entr_prob_dependence} reveals a weak negative linear correlation ($r = -0.37$, $p < 0.0001$).

\begin{figure}[tb]
\centering
\includegraphics[width=\linewidth]{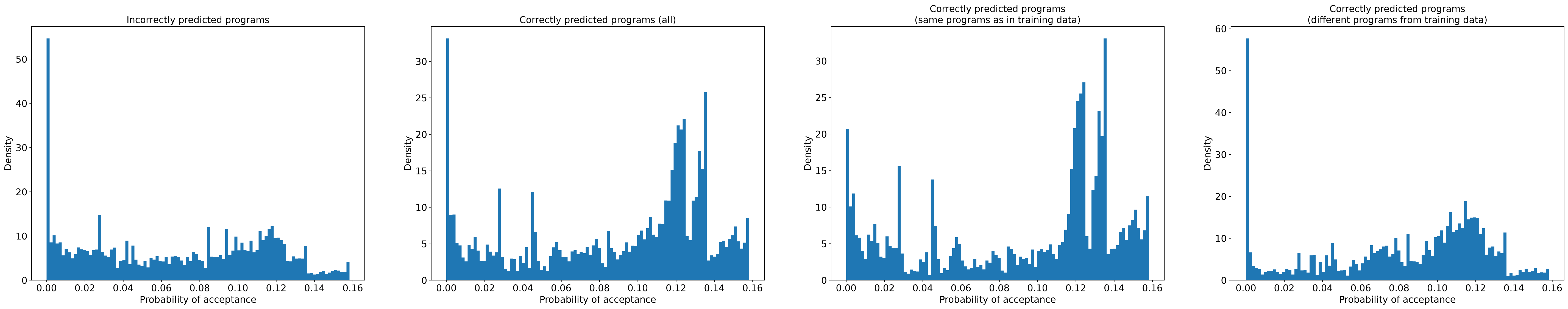}
\caption{The histograms of calibrator's probabilities.}
\label{fig:probs_hists}
\end{figure}

We also analyse the calibrator's probability in more detail. 
Figure~\ref{fig:probs_hists} characterizes the calibrator's probability distributions. The histogram for incorrectly predicted programs (left) exhibits a sharp unimodal peak at near-zero probability ($p \approx 0$, density $\approx 55$), indicating the model's low confidence in erroneous outputs, with only weak, diffuse elevation (density $< 15$) elsewhere in the range. 
Correctly predicted programs (all, second panel) instead show a clear bimodal distribution: a peak at $p \approx 0$ (density $\approx 33$) and a second mode around $p \approx 0.12-0.135$ reaching a comparable density of $\approx 25$, indicating two distinct confidence regimes among correct predictions.
Decomposing this by memorization status reveals a clear asymmetry. 
For programs identical to those seen during training (third panel), the high-confidence mode is in fact the dominant feature, peaking at $p \approx 0.135$ with density $\approx 33$ and exceeding the near-zero peak (density $\approx 20$) by roughly $1.6\times$ -- the calibrator is markedly more confident when the correct program matches training data. 
For correct predictions on programs structurally different from training (right panel), the pattern reverses: the near-zero peak becomes the single largest peak in the entire figure (density $\approx 58$), roughly $2.9\times$ its own high-confidence region (density $\approx 20$ at $p \approx 0.115$) and the corresponding near-zero peak in the training-identical subset (third panel). 

This divergence implies the model develops specialized confidence heuristics for generating functionally equivalent but structurally novel programs. The consistent concentration of all distributions in the low-probability regime ($p < 0.16$) further indicates that high-confidence predictions are rare, aligning with the weak entropy-probability correlation observed in Figure~\ref{fig:entr_prob_dependence}.

\subsection{Ontology 2: math problems}
\label{appendix:math_labels_descr}
To assess the correctness of math problem solutions, we employ two techniques:
\begin{enumerate}
    \item \textit{the LLM Judge} -- motivated by the impracticality of manual annotation at scale~\citep{teo2025enhance,huber2022ai}. We use Qwen2.5-32B~\citep{Qwen2.5-32b} as the judge. Each answer (from both the expert and the student) is split into double-newline-delimited fragments ("sentences"), which are assembled into a structured prompt asking the checker to align each sentence from the expert's trace to one or more sentences in the student's trace, assigning a confidence of "high", "medium", or "low" to each correspondence. The full prompt is provided in Appendix~\ref{appendix:llm_checker_math_problems_prompt}). Task correctness is then computed as the average confidence score ($1$, $0.5$, or $0$, respectively) across all associations and the final verdict is "correct" (label $1$) if the result $\ge 0.5$ otherwise "incorrect" (label $0$), following a similar method to~\citep{kahng2024llm}.
    
    \item \textit{the ProcessBench annotation} -- which reuses the benchmark's original evaluation protocol~\citep{processbench}. Answers are split into sentences using the same scheme as above and fed into the ProcessBench evaluator, which identifies the index of the first erroneous sentence in the student's trace, or signals a fully correct solution. We use DeepSeek-V3.2 as the annotator given its strong performance on math tasks~\citep{deepseekv32}. Inference was done locally, on NVidia H100 80GB GPUs, full annotation took six hours per dataset.
\end{enumerate}

\subsubsection{Prompt for LLM-based checker of the correctness of math problems'  solving}
\label{appendix:llm_checker_math_problems_prompt}

The full prompt for comparing the student’s model math problem solving with the expert’s via Qwen2.5-32B is the following: 

\begin{tcolorbox}[breakable, colback=gray!5, colframe=black]
You are analyzing two chain-of-thought (CoT) reasoning traces for the same math problem. One is from a large model and one is from a small model. \\

Your task is to map which sentences in the large CoT correspond to which sentences in the small CoT. Some sentences may correspond to multiple sentences, or some may have no clear correspondence. \\

\textbf{Question:}\\
\texttt{\{text\_of\_task\_prompt\}}\\

\textbf{Large Model CoT (numbered sentences):}\\
\texttt{\{expert\_list\_of\_sentences\}}\\

\textbf{Small Model CoT (numbered sentences):}\\
\texttt{\{student\_list\_of\_sentences\}}\\

Please output a JSON array where each element maps a large\_cot sentence index to matching small\_cot sentence indices. Use this format:

\begin{Verbatim}[breaklines=true, breakanywhere=true, fontsize=\small, breaksymbolleft={}, breaksymbolright={}]
[
  {{
    "large_idx": 0,
    "small_idx": [0, 1],
    "confidence": "high",
    "rationale": "Both introduce the problem and set up variables"
  }},
  {{
    "large_idx": 1,
    "small_idx": [2],
    "confidence": "medium",
    "rationale": "Similar equation setup but with different clarity"
  }}
]
\end{Verbatim}

Confidence levels: "high" (clear correspondence), "medium" (likely but not certain), "low" (weak correspondence) \\ 

Output only the JSON array, no additional text.
\end{tcolorbox}

\subsubsection{Results of labeling strategies for math tasks}
\label{appendix:math_labels_comparison}
\begin{figure}[tb]
\centering
\includegraphics[width=0.95\linewidth]{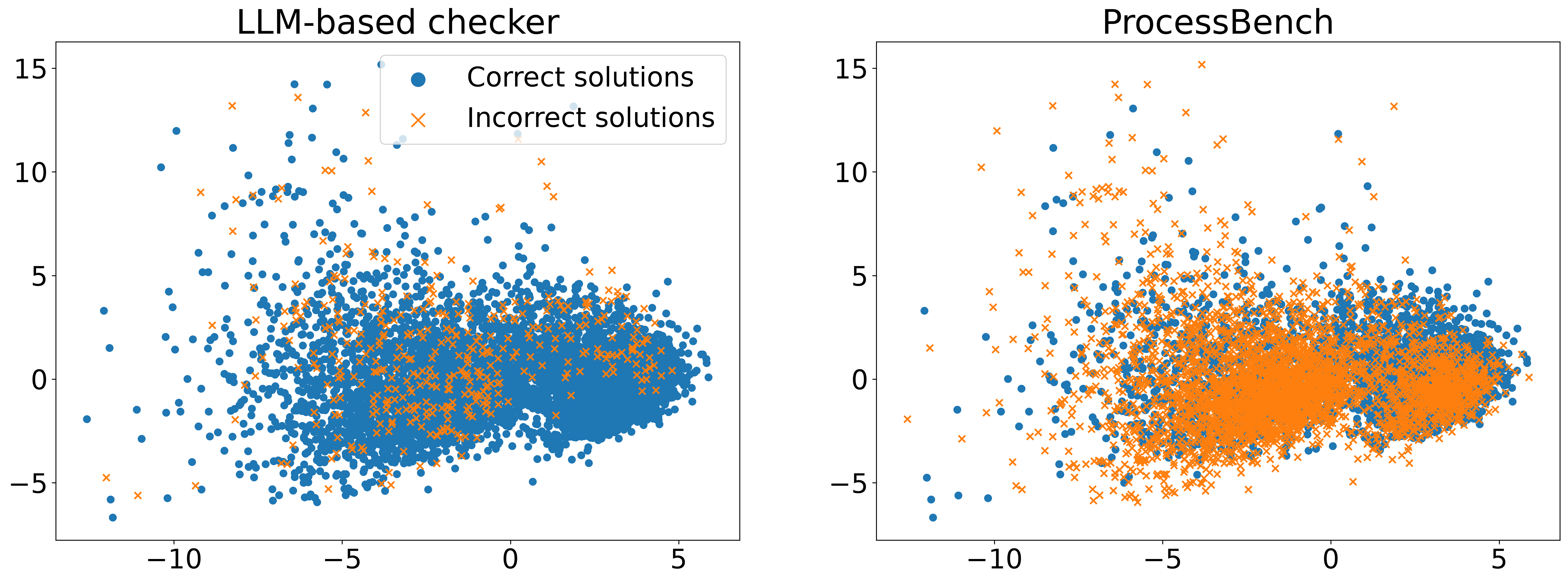}
\caption{PCA visualization of both labeling strategies.}
\label{fig:2_labeling}
\end{figure}

In this work, we consider only $6,845$ math problems for which we have both labelings: one from an LLM-based checker and one from ProcessBench. The LLM-based checker labeled $6,453$ problems as correctly solved, while ProcessBench labeled only $3,452$. 
Figure~\ref{fig:2_labeling} shows that neither labeling yields separable classes under PCA. Nevertheless, we consider the ProcessBench labeling as more reliable, since the LLM-based checker marked almost all problems as correct. Therefore, for medical question answering we use the ProcessBench labeling.

\subsection{Ontology 3: medical-related question answering}
\label{appendix:medicine}

For medical question answering, we provide the possible addition highlight of six parts of the models' outputs:
\begin{enumerate}
    \item \textbf{Decision} -- a clinical, surgical, or scientific action to take or avoid.
    \item \textbf{Reasoning} -- a pivotal logical step that drives the conclusion.
    \item \textbf{Conclusion} -- the final answer or diagnosis, expressed as a concept rather than a letter.
    \item \textbf{Caution} -- a risk, contraindication, warning, or negation that prevents a wrong action.
    \item \textbf{Mechanism} -- a biological, pharmacological, or molecular mechanism on which the answer hinges.
    \item \textbf{Elimination} -- reasoning that rules out a wrong option (critical for NOT/incorrect/except questions).
\end{enumerate}

To produce these highlights, we use MedGemma (\texttt{google/medgemma-1.5-4b-it}) with the following prompt: 
\begin{tcolorbox}[breakable, colback=gray!5, colframe=black]
You are a precise annotation judge for science and medicine questions.\\

Your task is to read a question and a model-generated answer, then identify spans
of text that are SEMANTICALLY PIVOTAL -- meaning that changing or removing them
would either reverse the meaning, lead to the wrong conclusion, or cause a
clinically or scientifically dangerous error.\\

The questions come from a broad range of ontologies:\\
  - Molecular and cell biology (mechanisms, pathways, organelles, proteins)\\
  - Genetics and genomics (mutations, inheritance, sequencing, gene regulation)\\
  - Physiology and anatomy (organ systems, hormones, homeostasis)\\
  - Neuroscience (synapses, neurotransmitters, brain function)\\
  - Ecology and evolution (selection, population dynamics, symbiosis)\\
  - Clinical medicine (diagnosis, treatment, surgical indications, patient management)\\
  - Pharmacology (drug targets, mechanisms of action, resistance)\\
  - Research methods (PCR, sequencing, microscopy, assays)\\
  - Healthcare systems and policy (EHR, public health, technology)\\

Semantically pivotal spans include:\\
  - Direction or magnitude words whose reversal flips the answer:\\
    "increase" vs "decrease", "inhibits" vs "activates", "proximal" vs "distal"\\
  - A negation that inverts the meaning of a claim:\\
    "not", "except", "unlikely", "absent"\\
  - A specific named entity the answer critically depends on:\\
    a drug name, enzyme, gene, pathway, or anatomical structure\\
  - A key quantity, threshold, or relationship:
    "above pH 7", "half-life of 8 days", "binds competitively"\\
  - The clause that eliminates a distractor in NOT/incorrect/except questions:\\
    the reasoning that rules out a wrong option, not just the option letter\\

Do NOT highlight:\\
  - Bare answer labels in isolation ("answer D", "option C") -- highlight the
    reasoning or concept that justifies the choice instead\\
  - Connector words alone ("because", "therefore", "however") -- highlight the
    clause they introduce if that clause is pivotal, not the connector itself\\
  - Background or contextual sentences that do not affect the correctness of
    the answer if removed\\

If the answer contains only a bare label with no reasoning, return an empty
highlights list and note this in the summary.\\

Return ONLY a valid JSON object (no markdown fences, no extra text):\\
\begin{Verbatim}[breaklines=true, breakanywhere=true, fontsize=\small, breaksymbolleft={}, breaksymbolright={}]
{
  "highlights": [
    {
      "span":   "<exact substring copied character-for-character from the answer>",
      "type":   "<one of: decision | reasoning | conclusion | caution | mechanism | elimination>",
      "reason": "<one sentence explaining why changing this span would make the answer wrong>"
    }
  ],
  "summary": "<2-3 sentence summary of what is semantically pivotal in this answer>"
}
\end{Verbatim}
Type definitions:\\
  decision    - a clinical, surgical, or scientific action to take or avoid\\
  reasoning   - a pivotal logical step that drives the conclusion\\
  conclusion  - the final answer or diagnosis, expressed as a concept not a letter\\
  caution     - a risk, contraindication, warning, or negation that prevents a wrong action\\
  mechanism   - a biological, pharmacological, or molecular mechanism the answer hinges on\\
  elimination - reasoning that rules out a wrong option (critical for NOT/incorrect/except questions)\\

Rules:\\
  * "span" must be an exact character-for-character substring of the answer text\\
  * Highlight every span that is semantically pivotal -- neither more nor less\\
  * Do not add spans to reach a minimum count\\
  * Do not drop spans to stay under a maximum count\\
  * A short answer may have 1-2 highlights; a long reasoning chain may have more\\
  * Spans may overlap; that is fine\\
  * Do NOT include any text outside the JSON object."
\end{tcolorbox}

\begin{figure}[tb]
\centering
\includegraphics[width=\linewidth]{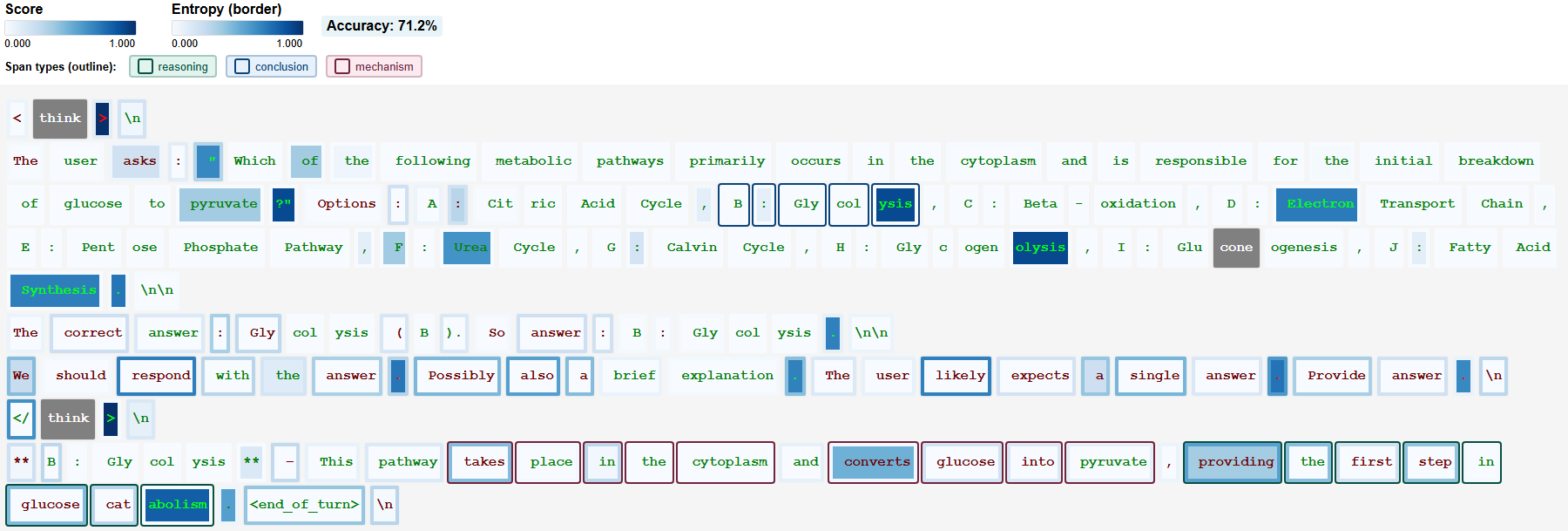}
\caption{Visualization comparing the expert's and the student's answers to the following question: which of the following metabolic pathways primarily occurs in the cytoplasm and is responsible for the initial breakdown of glucose to pyruvate? A: Citric Acid Cycle; B: Glycolysis; C: Beta-oxidation; D: Electron Transport Chain; E: Pentose Phosphate Pathway; F: Urea Cycle; G: Calvin Cycle; H: Glycogenolysis; I: Gluconeogenesis; J: Fatty Acid Synthesis.}
\label{fig:medgemma_highlight}
\end{figure}

Figure~\ref{fig:medgemma_highlight} shows an example containing three types of highlights: reasoning ("providing the first step in glucose catabolism"), conclusion ("B: Glycolysis"), and mechanism ("takes place in the cytoplasm" and "converts glucose into pyruvate").

These highlights help identify which parts of a reasoning trace deserve closer attention. While a mistake on the example shown here would be inconsequential, this is not the case for many other medical questions, where an incorrect answer could cause real harm. These highlights can also be used in the labeling process: for example, an answer may be marked incorrect if an error appears in any span belonging to one of these six categories.    

\subsection{Positional biases of calibrators}
\label{appendix:positional_biases}

To build the AI ontology calibrator, we need to label each student token as correct or not. This token-level correctness is a fundamentally ill-defined, almost philosophical question: there is no principled way to decide which individual token is correct or incorrect. In this paper, we implement a simple technique: we compare all student tokens with the expert tokens at the same positions and label a student token as correct if the student and the expert have the same token at that position, and as incorrect otherwise.

\begin{figure}[tb]
\centering
\begin{subfigure}[b]{0.49\linewidth}
    \includegraphics[width=\linewidth]{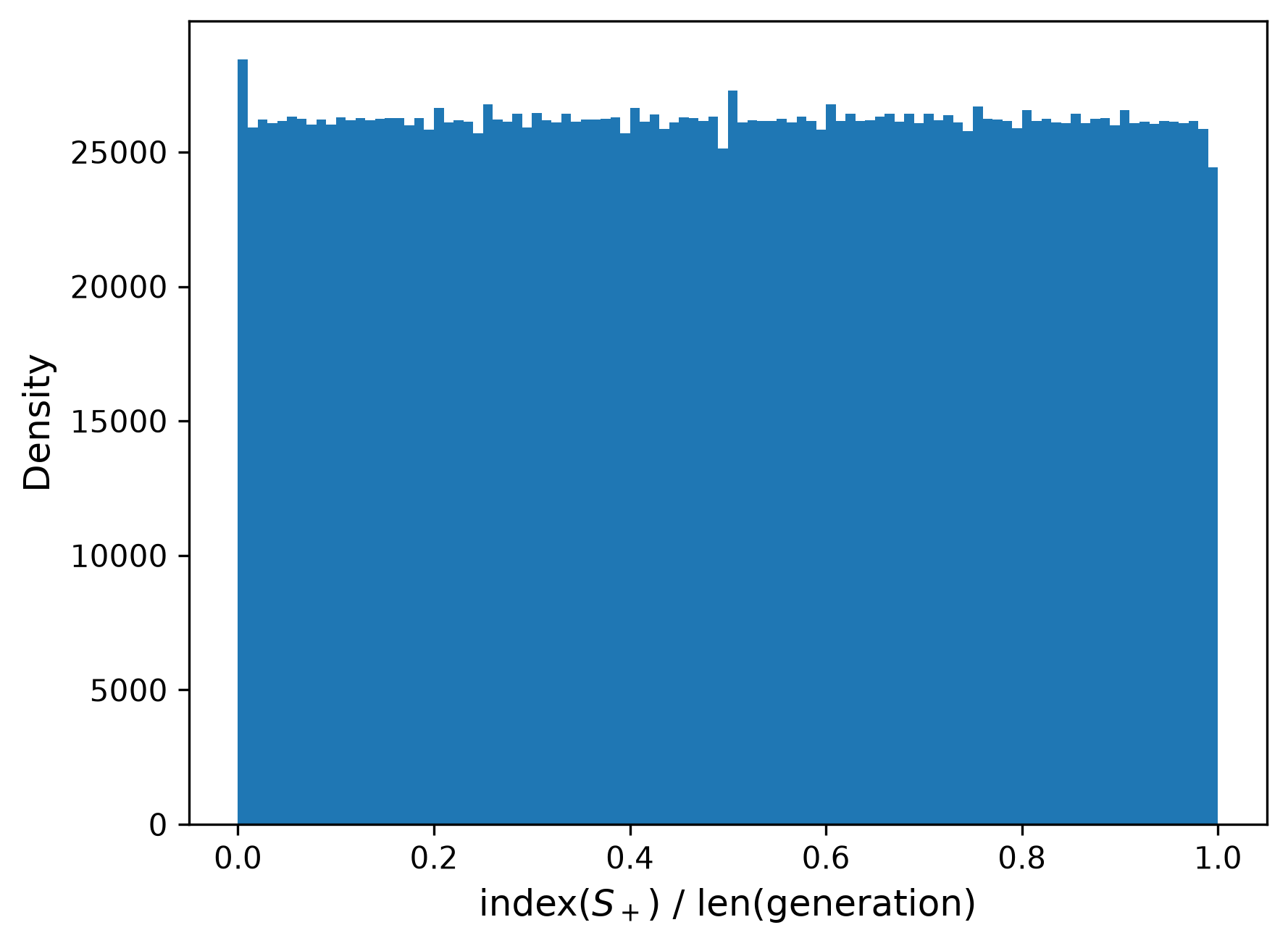}
    \caption{Qwen3-8B (Math).}
    \label{subfig:pos_bias_math}
\end{subfigure}
\hfill
\begin{subfigure}[b]{0.49\linewidth}
    \includegraphics[width=\linewidth]{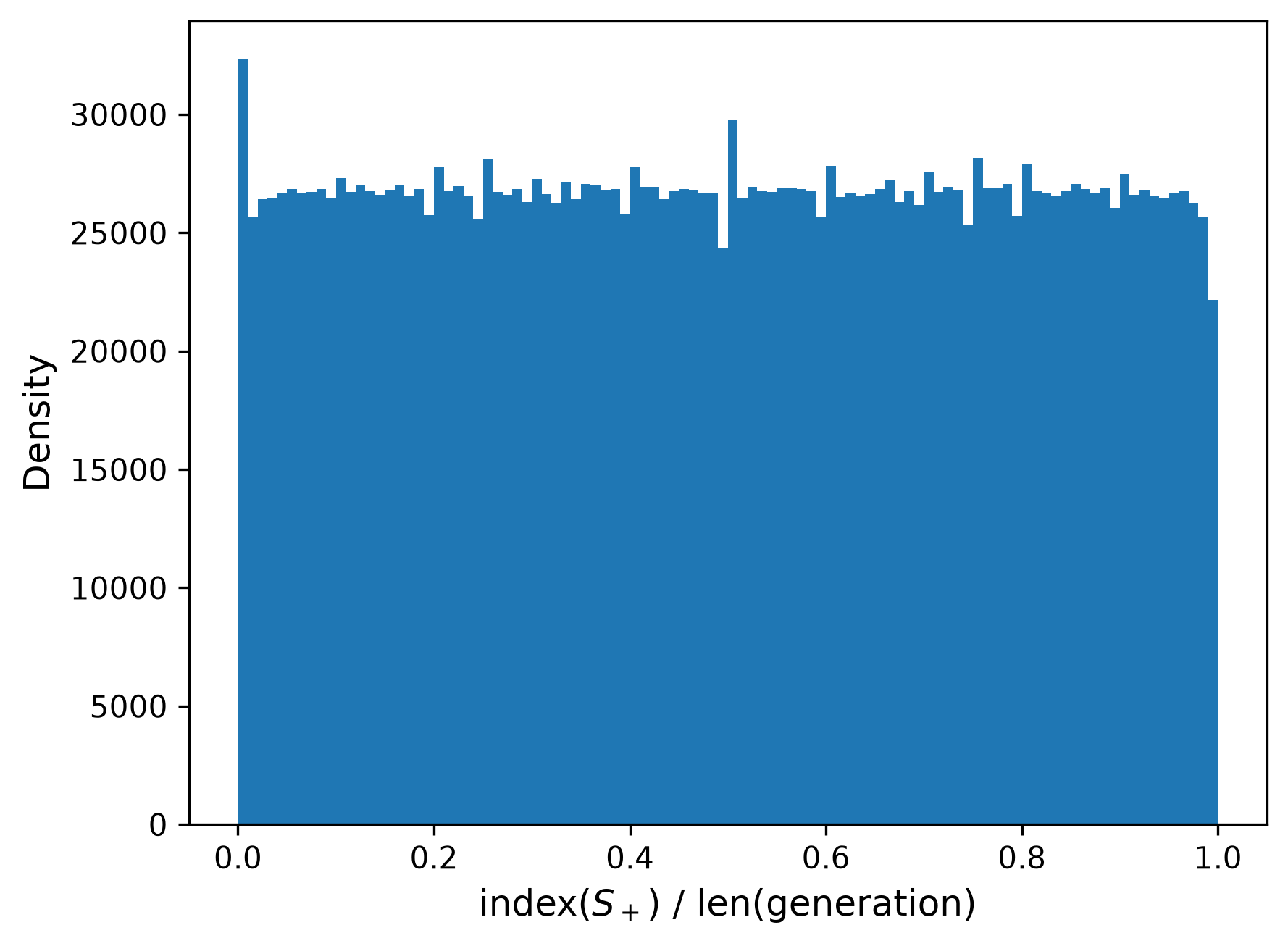}
    \caption{MedGemma (Medicine).}
    \label{subfig:pos_bias_mg}
\end{subfigure}
\caption{Distribution of the relative position of tokens in $S_{+}$ within the student generation, $\text{index}(S_{+}) / \text{len}(\text{generation})$, for (a) Math with Qwen3-8B and (b) Medicine with MedGemma as the student. The distributions are approximately uniform, showing that tokens labeled as correct are not concentrated at the beginning, middle, or end of the response.}
\label{fig:pos_bias}
\end{figure}

To ensure this approach does not introduce any biases, we additionally conduct the following experiment: we plot the distributions of normalized indices of tokens from $S_{+}$. Figure~\ref{fig:pos_bias} shows the results for Qwen3-8B on Math and MedGemma on Medicine. In both cases, the distribution is nearly uniform, so no positional bias is found. Nevertheless, even if it is present, it does not harm our theory. 

\section{Relations of derived indicators for classifiers}
\subsection{Analysis of indicators and baselines on all ontologies}
\label{appendix:clf_features_correlations}

\begin{figure}[tb]
\centering
\includegraphics[width=0.95\linewidth]{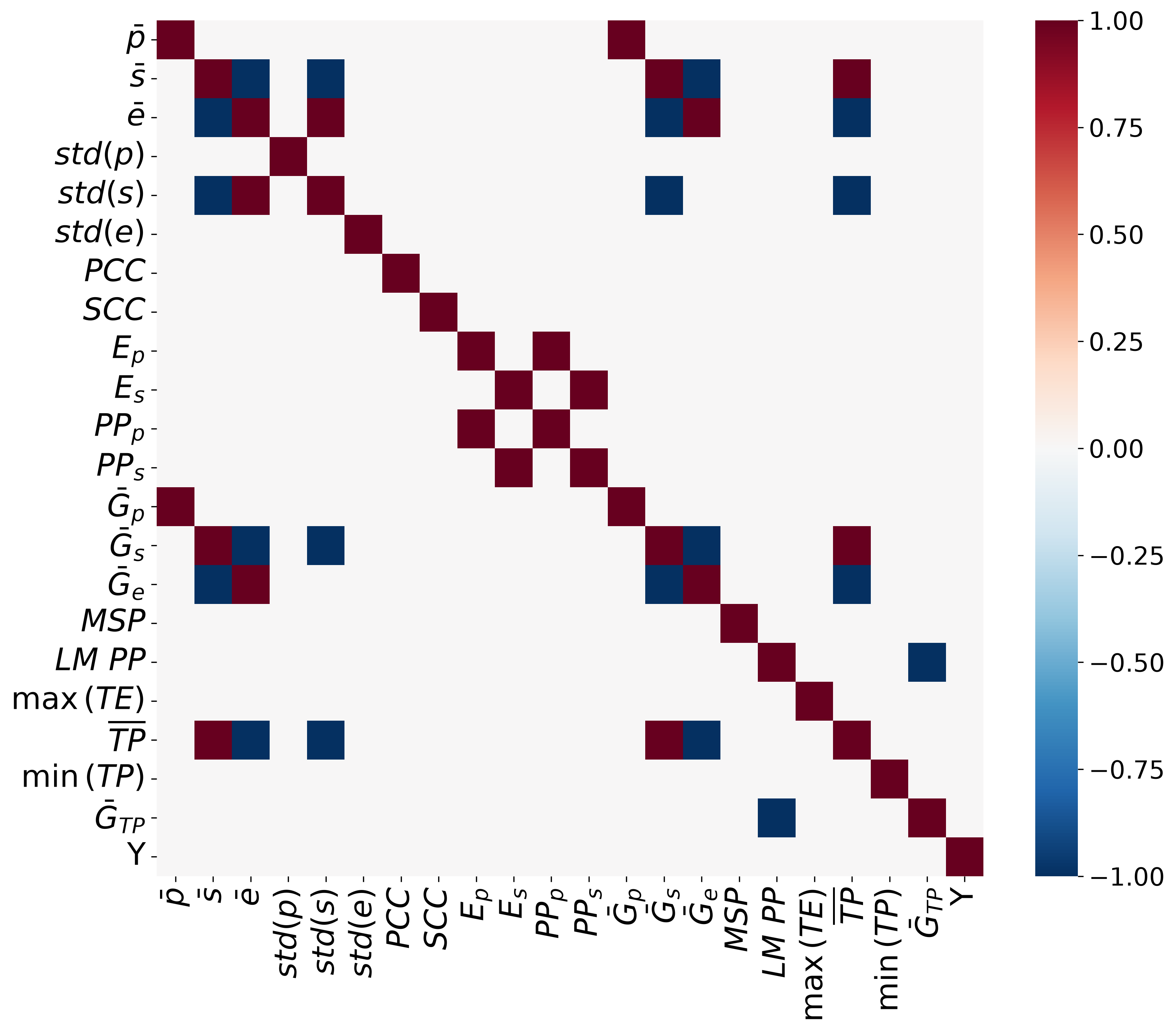}
\caption{Statistically significant strong Spearman correlations (absolute coefficient $> 0.6$, p-value $\le 0.05$) between the $15$ classifier indicators and the target label, aggregated across all ontologies and models. Only pairs for which all correlations satisfy both criteria are shown. The target labels are presented with $Y$.}
\label{fig:corrs_15_features}
\end{figure}

To analyse the relations between the $15$ indicators and six baselines used for training classifiers, we calculate pairwise Spearman correlations, including correlations with the target label for all our domains, labelings, and students.
Figure~\ref{fig:corrs_15_features} presents feature pairs with a statistically significant (p-value $\le 0.05$) absolute correlation coefficient exceeding $0.6$ for all models and ontologies. The calibrator-based probability indicators provide signals largely independent from the entropies- and softmaxes-based indicators. In general, the following indicators provide their unique signals with no correlations with others: $std(p), std(e),\ PCC, SCC,\ MSP, \max{(TE)}, \min{(TP)}$. In this list, we did not include $\bar{p}$ and $\bar{G}_p$ as they correlate with each other, which is expected (will be additionally mentioned below). 

The obtained correlations can be divided into the following groups: 
\begin{enumerate}
    \item \textbf{Mathematically trivial}: the correlation coefficient is not an empirical finding but a mathematical identity or a consequence of a monotone transformation. The correlated pairs of $\left(E_s, PP_s \right), \ \left(E_p, PP_p \right), \ \left(LM\ PP, \frac{1}{\bar{G}_{TP}} \right)$ are the monotone transformations.
    
    \item \textbf{Information theory}: the correlation is fully predicted by information theory. Two distinct sub-patterns appear here. First, \emph{confidence-proxy convergence}: $\left(\bar{s}, \overline{TP}\right)$, $\left(\bar{G}_s, \overline{TP}\right)$, and $\left(\bar{G}_s, \bar{G}_{TP}\right)$. $\bar{s}$ aggregates the probability of the single most likely token at each position, regardless of what token was actually produced, while $\overline{TP}$ aggregates the probability the model assigns to the token it actually generated; the two are identical under greedy decoding and remain strongly correlated otherwise, since sequences where the model's best guess is confident also tend to be sequences where the realised token was highly probable. Second, the \emph{entropy--probability inverse relationship}: $\left(\bar{e}, \frac{1}{\bar{s}}\right)$, $\left(\bar{e}, \frac{1}{\bar{G}_s}\right)$, $\left(\bar{s}, \frac{1}{\bar{G}_e}\right)$, 
    $\left(\bar{G}_s, \frac{1}{\bar{G}_e}\right)$, 
    $\left(\bar{G}_e, \frac{1}{\overline{TP}}\right)$, and $\left(\bar{e}, \frac{1}{\overline{TP}}\right)$. A peaked (low-entropy) distribution necessarily places more mass on its top token, so any entropy-side statistic is inversely related to any probability-side statistic; these six pairs are the same relationship instantiated across the arithmetic/geometric aggregation choices on each side.
    
    \item \textbf{Aggregation robustness}: $\left(\bar{e}, \bar{G}_e\right)$, $\left(\bar{s}, \bar{G}_s\right)$, and $\left(\bar{p}, \bar{G}_p\right)$ confirm that switching from the arithmetic to the geometric mean of the same underlying per-token quantity (entropies, softmaxes, or calibrator probabilities) does not change the ranking of instances, i.e. the results are not artefacts of the choice of aggregator.
	
    \item \textbf{Bounded Support Artifact}: the correlation arises from the bounded range $(1/K, 1]$ that top-1 softmax values are confined to, not from a deep property of model behavior. Since each top-1 softmax value $s_t \in (1/K, 1]$ is bounded above, sequences that are consistently confident cluster near the ceiling and sequences that are consistently uncertain cluster near the floor $1/K$; in both regimes the values are compressed and $std(s)$ is low, so $\bar{s}$ and $std(s)$ move in opposite directions: $\left(\bar{s}, \frac{1}{std(s)}\right)$. Only an unstable mix of confident and uncertain positions yields a high $std(s)$. The pair $\left(std(s), \frac{1}{\bar{G}_s}\right)$ is not an independent finding: it follows by transitivity from $\left(\bar{s}, \frac{1}{std(s)}\right)$ combined with the aggregation-robustness identity $\bar{G}_s \sim \bar{s}$.				
    
    \item \textbf{Surprising / Non-Obvious}: within a single token position, a more peaked softmax distribution mechanically implies both a higher top-1 value and a lower entropy -- a negative relationship. The non-obvious finding is that, across instances, the sign flips: $\left(\bar{e}, std(s)\right)$ is \emph{positively} correlated -- instances that are uncertain on average (high $\bar{e}$) are also the instances whose top-1 confidence is most \emph{unstable} across positions (high $std(s)$), rather than uniformly low, reflecting instance-level difficulty where some positions remain confident while others do not, inflating the variance of $s_t$. This aligns with a second, distinct pair, $\left(std(s), \frac{1}{\overline{TP}}\right)$: instances with high positional variability in top-1 confidence also show low average softmaxes on the tokens actually produced, i.e. the model is simultaneously erratic in its best guess \emph{and} less committed to its actual output. Combined, these compose into $\left(\bar{e}, \frac{1}{\overline{TP}}\right)$: average per-token uncertainty positively correlated with low realised-token confidence, mediated by $std(s)$.

\end{enumerate}

In can also be noted that none of the provided indicators correlate with the target label. 

\subsection{Analysis of indicators including verbalized confidence on medical ontology}
\label{appendix:clf_features_correlations_medicine}
Some papers~\citep{kadavath2022language, lin2022teaching} report a correlation between generation correctness and the model's self-assessment labels. To verify this claim, we additionally prompt our students on the medical ontology data to label the correctness of their answers and assess their confidence on a scale from $1$ to $10$.

For self-correctness assessment, we tried to follow the technique from~\citep{kadavath2022language} and reuse their prompt; however, our models produced outputs whose verdicts were almost impossible to parse. Therefore, for self-correctness assessment we used the following prompt:
\begin{tcolorbox}[breakable, colback=gray!5, colframe=black]
I will give you a question and a possible answer with reasoning traces. Please output whether the possible answer is correct or not. Output only a single word: "correct" or "wrong".\\  
 
Question:\\
\texttt{\{text\_of\_prompt\}}\\
         
Possible Answer:\\
\texttt{\{text\_of\_student\_response\}}\\

The possible answer is:
\end{tcolorbox}

The prompt used for self-confidence assessment is as follows:
\begin{tcolorbox}[breakable, colback=gray!5, colframe=black]
I will give you a question and your answer with reasoning traces. Please output whether you are confident in the provided answer. Output only a single value from 1 to 10, where 1 means you are not confident and 10 means you are completely confident. \\   

Question:\\
\texttt{\{text\_of\_prompt\}}\\
            
Your Answer:\\
\texttt{\{text\_of\_student\_response\}}\\

Confidence score is:
\end{tcolorbox}

\begin{figure}[tb]
\centering
\begin{subfigure}[b]{0.49\linewidth}
    \includegraphics[width=\linewidth]{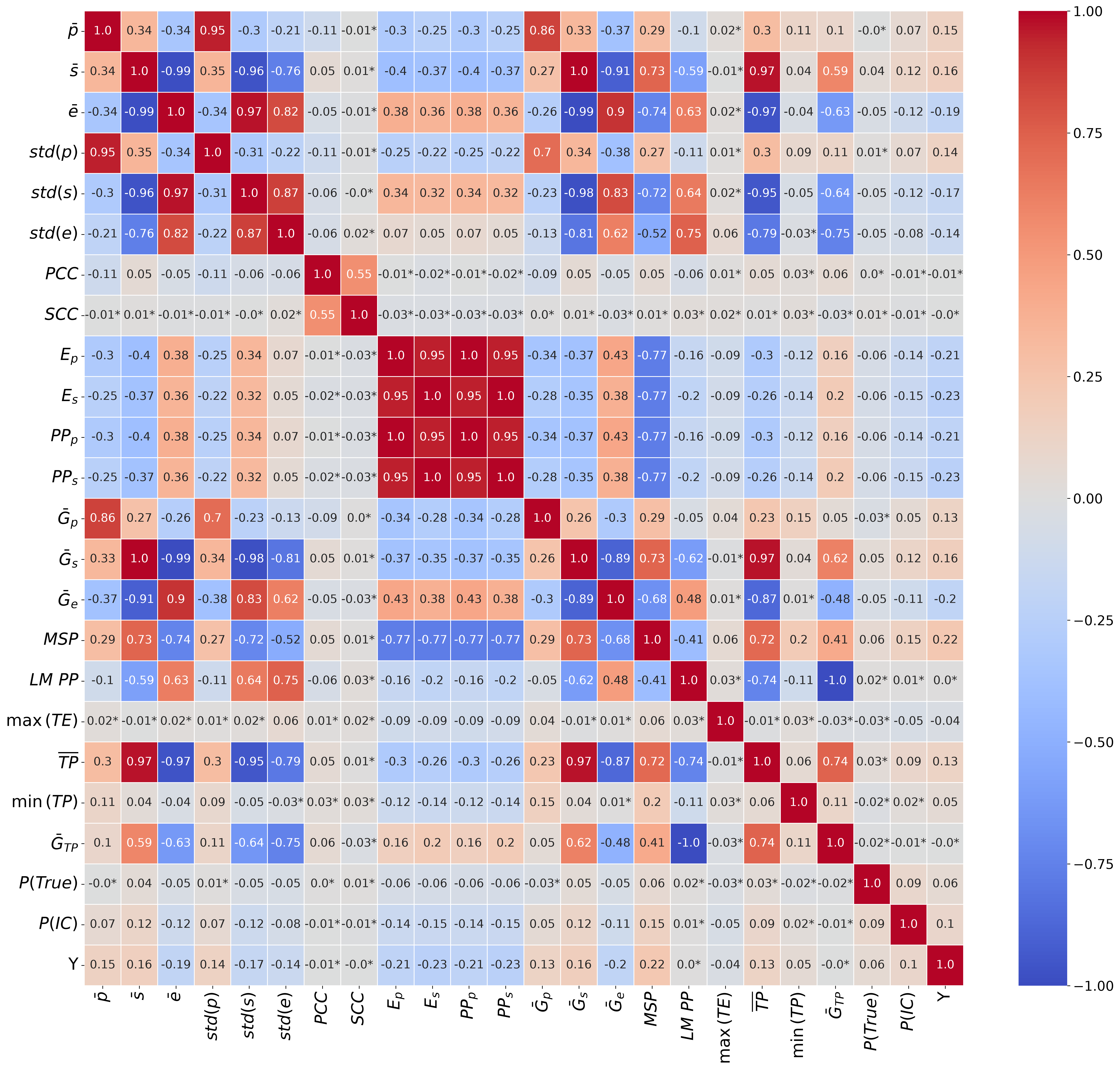}
    \caption{MedGemma.}
    \label{subfig:corrs_medgemma}
\end{subfigure}
\hfill
\begin{subfigure}[b]{0.49\linewidth}
    \includegraphics[width=\linewidth]{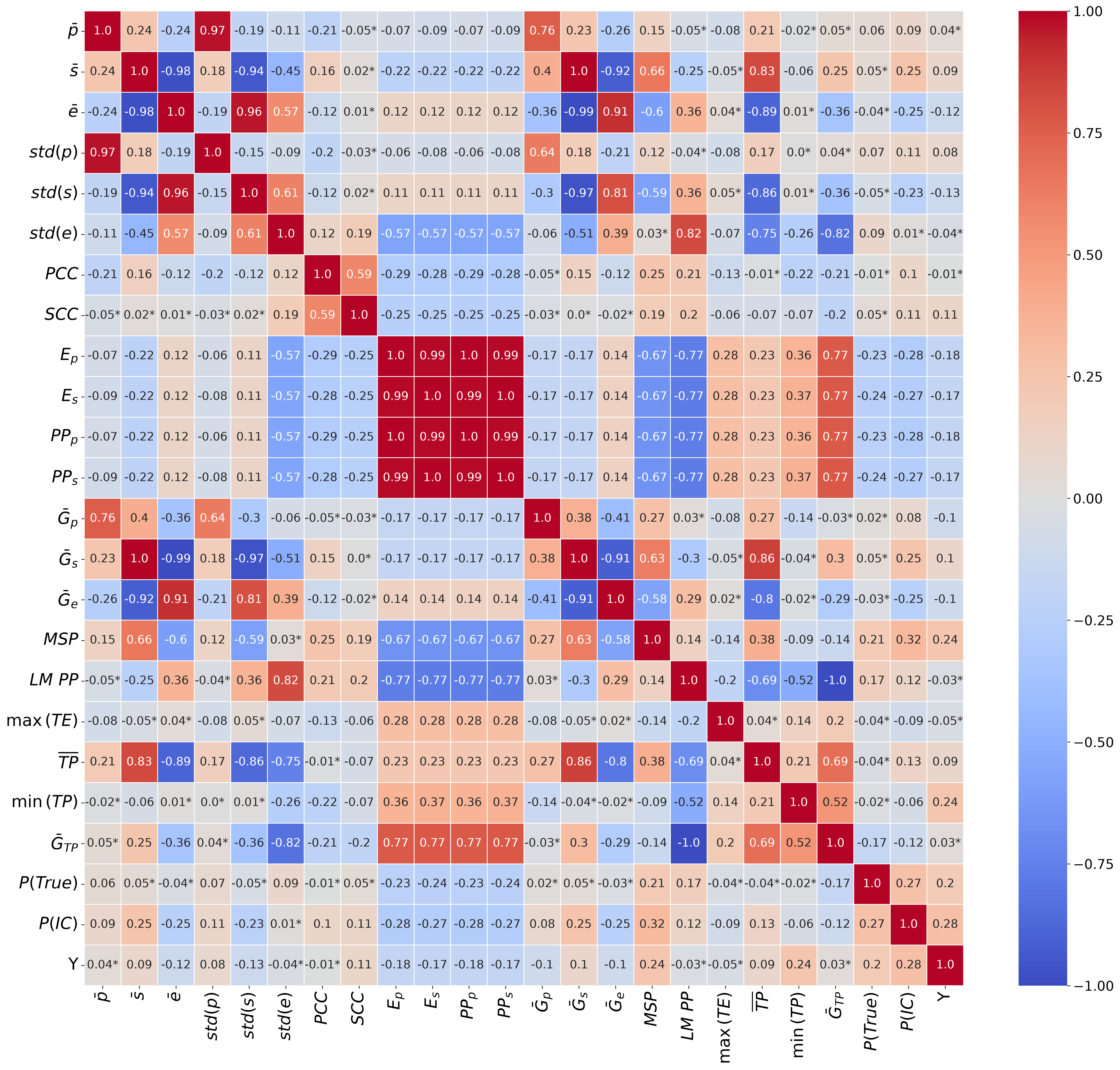}
    \caption{Qwen3-8B.}
    \label{subfig:corrs_q3_8b}
\end{subfigure}
\vspace{0.5em}
\begin{subfigure}[b]{0.49\linewidth}
    \includegraphics[width=\linewidth]{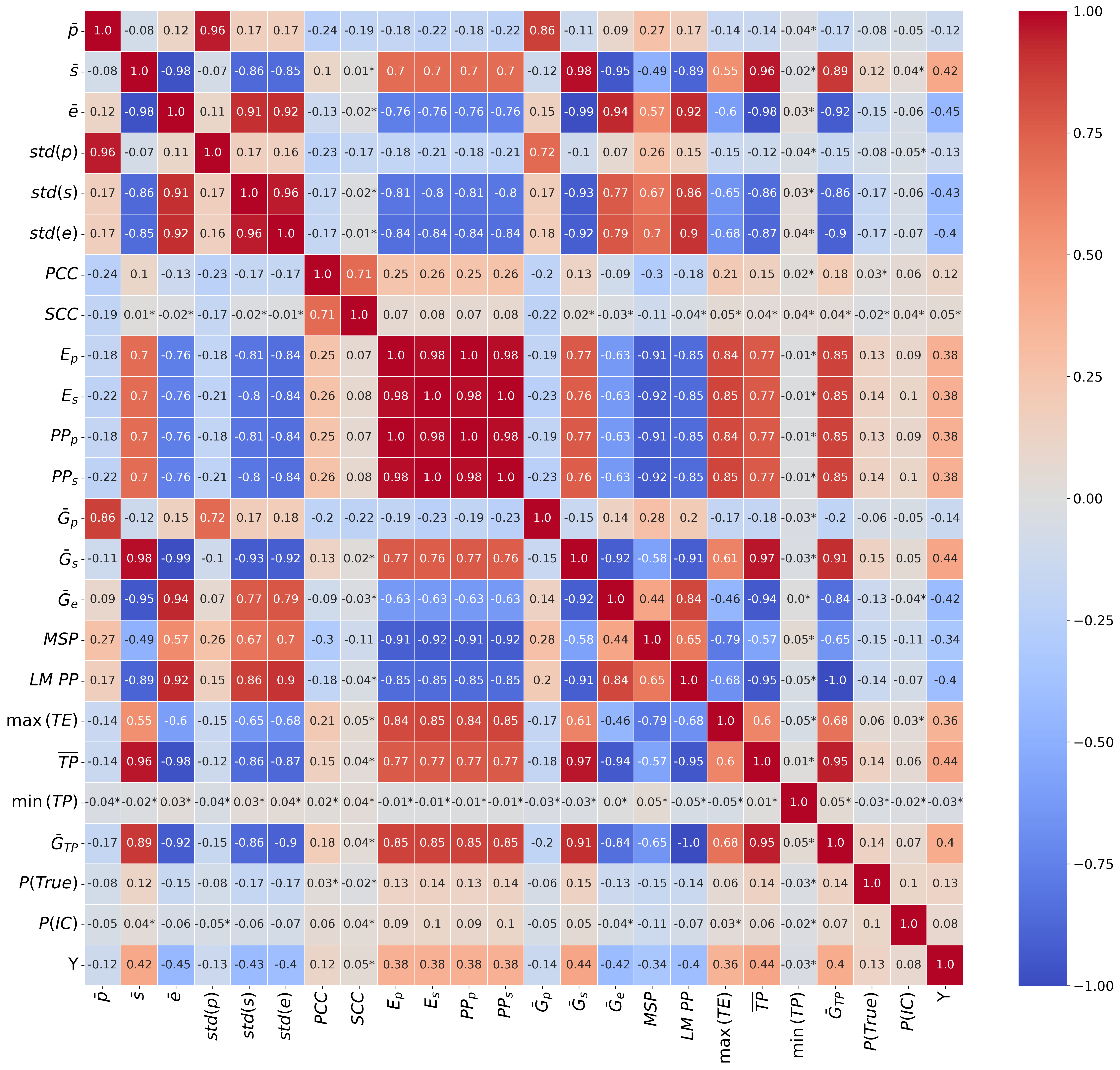}
    \caption{Qwen3-0.6B.}
    \label{subfig:corrs_q3_06b}
\end{subfigure}
\hfill
\begin{subfigure}[b]{0.49\linewidth}
    \includegraphics[width=\linewidth]{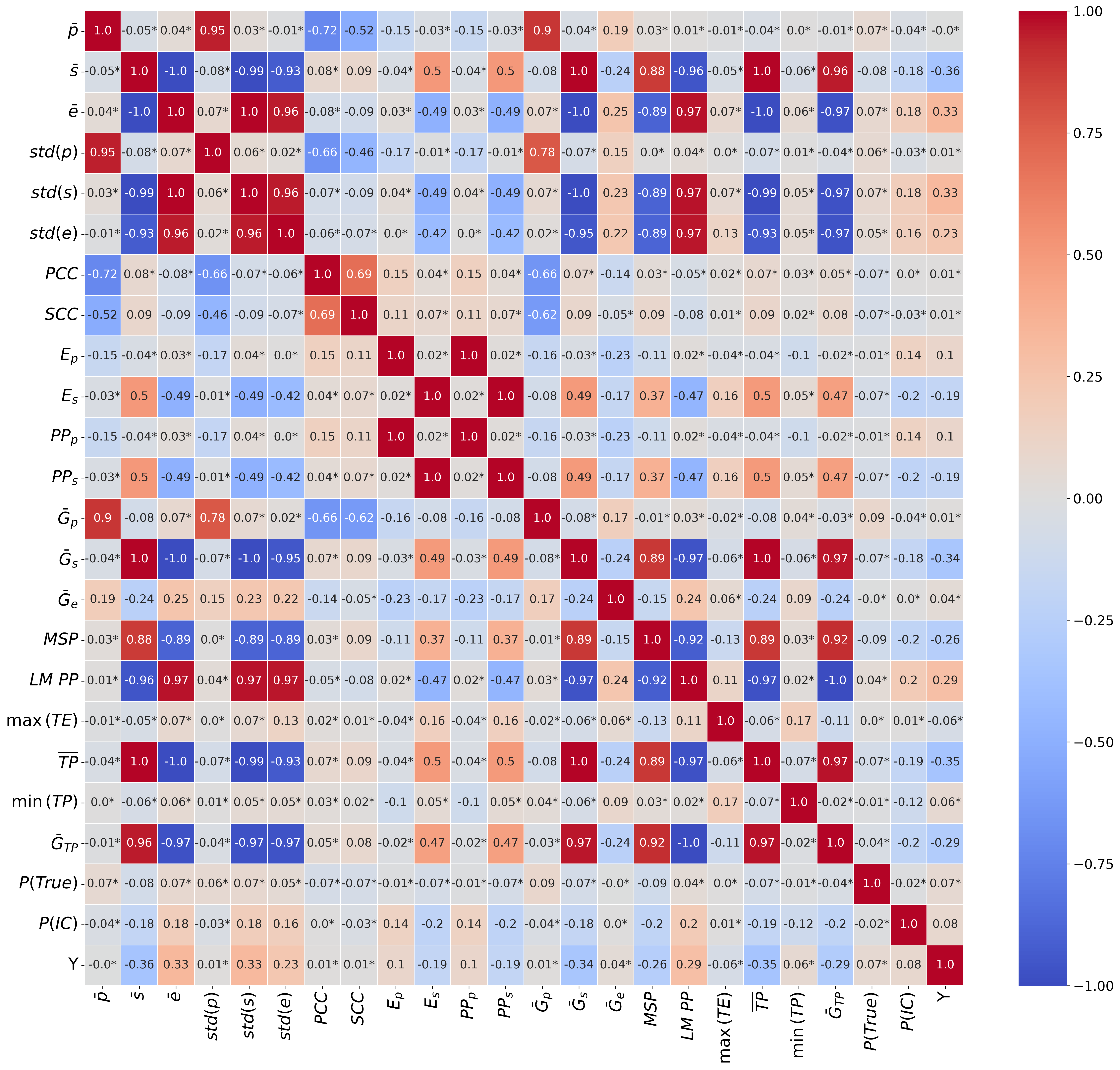}
    \caption{Qwen3.5-0.8B.}
    \label{subfig:corrs_q35_08b}
\end{subfigure}
\caption{Spearman correlation matrices for four students on the medical ontology data. ProcessBench labels are denotes as $Y$.}
\label{fig:corrs_with_student_is_correct_confident}
\end{figure}

Figure~\ref{fig:corrs_with_student_is_correct_confident} presents the correlation of indicators with baselines and self-assessment labels: assessment of response correctness ($P(True)$) and assessment of confidence in a response ($P(IC)$). For self-correctness and self-confidence assessment, we considered the responses labeled with ProcessBench. As shown in  Figure\ref{fig:corrs_with_student_is_correct_confident}, only about half of such data have self-assessment labels for the considered students: MedGemma -- $58.7\%$ ($3,509$ responses with both self-assessment labels out of $5,978$ labeled with ProcessBench), Qwen3-8B --  $40.8 \%$ ($1,308$ out of $3,207$), Qwen3-0.6B --  $38.6 \%$ ($1,415$ out of $3,663$), and Qwen3.5-0.8B --  $31.5 \%$ ($615$ out of $1,954$). Nevertheless, we only consider responses from which self-assessment labels can be easily obtained: if a response contained a label but its parsing was difficult, it was not included in our list. Overall, unlike what is reported in other papers, we do not observe strong correlations of either $P(True)$ or $P(IC)$ with $Y$.

\section{Technical details of built classifiers}
\label{appendix:clf_techincal_details}
\begin{figure}[tb]
\centering
\includegraphics[width=0.6\linewidth]{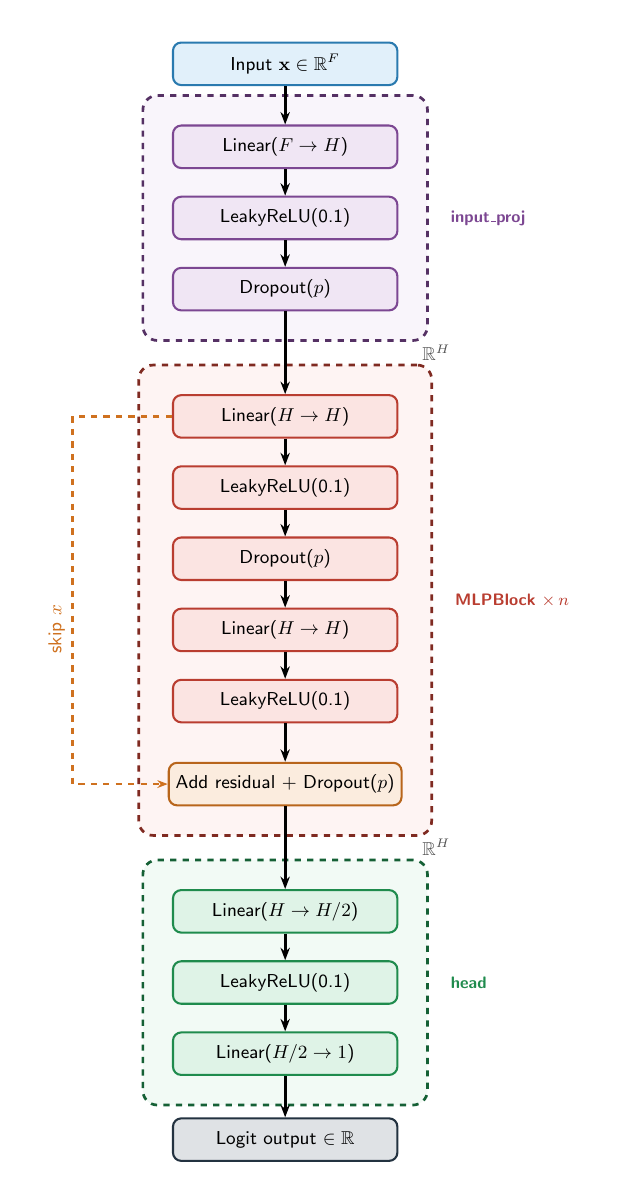}
\caption{The scheme of the MLP architecture.}
\label{fig:mlp_architecture}
\end{figure}

In Section~\ref{sec:experiments} we provide the metrics of trained Logistic Regression (with default hyperparameters (with \texttt{random\_state}$= 123$) on each of our $15$ indicators, six baselines, and two cases of several indicators usage: all $15$ indicators and the first $12$ indicators (the set of all indicators without $\bar{G}_{\cdot}$-based). We additionally conduct the training of XGBoost with default hyperparameters (with \texttt{random\_state}$= 123$) and MLP with the architecture presented in Figure~\ref{fig:mlp_architecture} on all possible combinations of our $15$ indicators. The classifiers were trained on an internal cluster, consisting of NVidia Tesla H200 141GB and NVidia RTX 4090 96GB GPUs. Total training time was around $96$ hours for both ablations and the final training run.

For the MLP, we use $n = 1$ block, a hidden dimension of $H = 128$, and a dropout rate of $p = 0.2$. 

All classifiers are trained on $20$-folds cross-validation with stratification w.r.t. target labels. For XGBoost and MLP, we used weighted \texttt{BCEWithLogitsLoss} loss function (with \texttt{pos\_weight}$=N_\text{zeros} / N_\text{ones}$) for all the ontologies. For MLP, we additionaly choose $20\%$ of training part on each fold for validation and normalise the data. We also use \texttt{AdamW} optimizer with learning rate $=10^{-3}$ and weight decay $=10^{-4}$, \texttt{ReduceLROnPlateau} scheduler maximizing F1 score with patience $=5$, factor $=0.5$, \texttt{EarlyStopping} with patience $=7$ and delta $=10^{-4}$. We train all out MLPs with $300$ epochs for each features set.

\begin{figure}[tb]
\centering
\begin{subfigure}[b]{0.49\linewidth}
    \includegraphics[width=\linewidth]{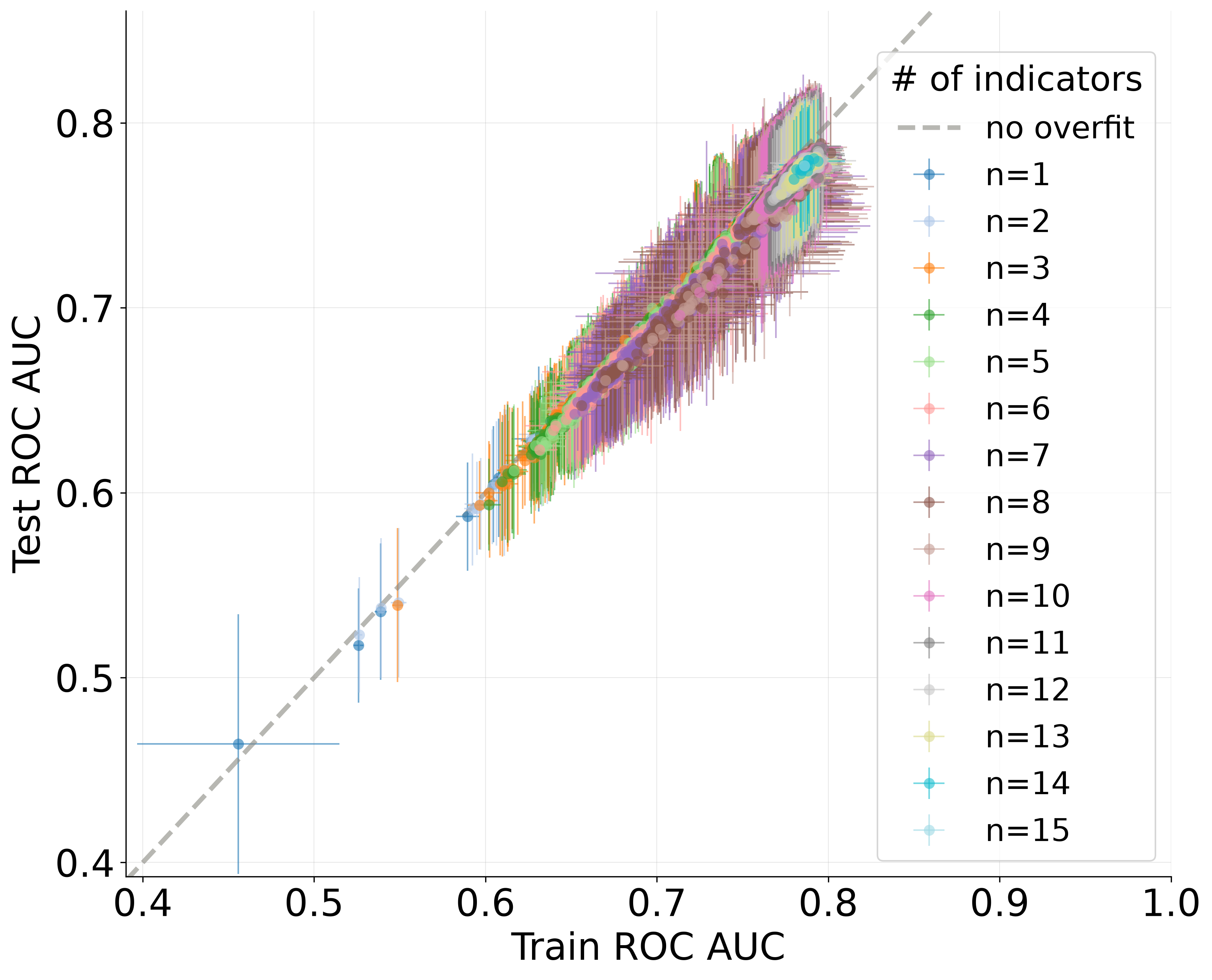}
    \caption{MLP classifier.}
    \label{fig:overfit_mg_mlp}
\end{subfigure}
\hfill
\begin{subfigure}[b]{0.49\linewidth}
    \includegraphics[width=\linewidth]{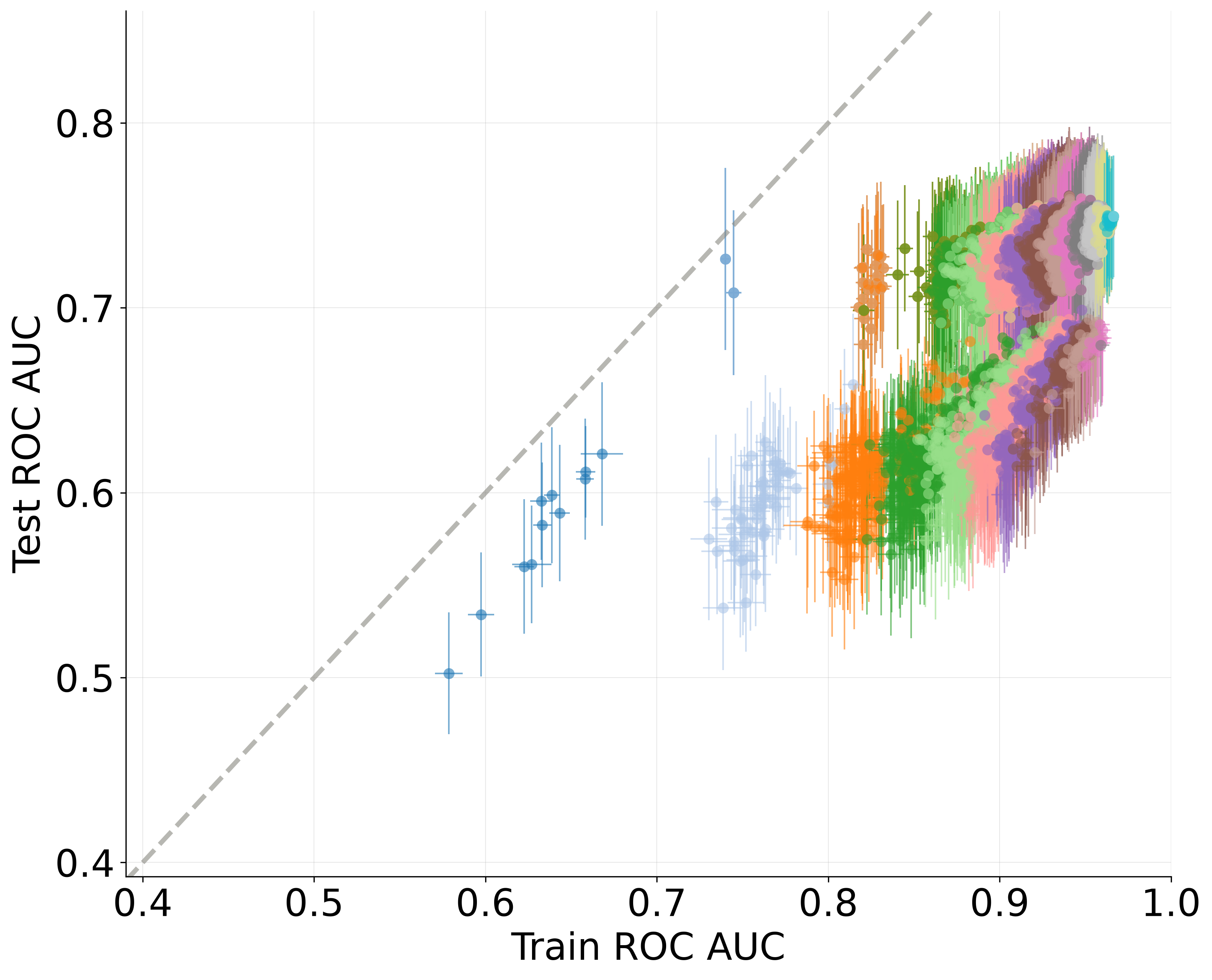}
    \caption{XGBoost classifier.}
    \label{fig:overfit_mg_xgb}
\end{subfigure}
\caption{Training and testing ROC~AUC of classifiers trained on MedGemma features.}
\label{fig:train_test_metrics_clfs}
\end{figure}

Figure~\ref{fig:train_test_metrics_clfs} shows the training process of XGBoost and MLP classifiers on the medical ontology with MedGemma. Although both classifiers achieve comparable ROC~AUC scores on the test data, XGBoost exhibits clear overfitting, as its training score is substantially higher and is close to the ideal value of $1$. We observe similar behavior across the other ontologies. Therefore, in the paper, we focus on MLP and in Section~\ref{sec:experiments} we present only the results for the best MLPs.

\section{Statistical significance of the classifiers' results}
\label{appendix:stat_signif_of_results}

We additionally conducted the statistical tests to check whether adding the calibrator-based indicators significantly improves correctness classification. For this, we ran a one-tailed paired t-test with H1: $\Delta = \text{mean}(d_j) > 0$, where $j \in [1, 20]$ is the fold and $d_j = BA_j(\text{traditional + calibrator indicators}) - BA_j(\text{traditional})$, where $BA$ denotes balanced accuracy. The test parts of the folds do not overlap, but the training sets do. We therefore also applied the Nadeau-Bengio corrected t-test, and Table~\ref{tab:stat_signif_bf} reports these results for BF ontology. We adjusted p-values with the Holm method, separately for the two comparison groups (traditional indicators with and without geometric means). \textbf{A result is significant if $\Delta > 0$ and $p(\text{Holm}) < 0.05$}.

\begin{table*}[t]
    \caption{Statistical significance of the balanced accuracy gain from adding calibrator-based indicators to traditional indicators, on the BF ontology with the Custom Transformer student. Balanced accuracy is reported as $\mathbf{mean} \pm \mathbf{std}$ over $20$ folds. $\Delta$ is the mean per-fold balanced accuracy difference relative to the traditional baseline, with a one-sided $95\%$ confidence interval. Calibrator indicators are computed from the probability of rejection ($p_{rej}$), the probability of acceptance (denoted as $p$ in Section~\ref{sec:methodology}), or the combined probability of two previous ones ($p_{comb} = \frac{p}{p + (1 - p_{rej})}$). All presented entries are significant ($\Delta > 0 \text{ and } p(\text{Holm}) < 0.05$).}
    \label{tab:stat_signif_bf}
    \centering
\begin{NiceTabular}{c c c l l}[hlines={1-3,6,7,10-12}, vlines={2,3}]
  \Block{1-2}{Indicators} & & 
  \makecell{Test mean $\pm$ std \\ Balanced \\ Accuracy} & 
  \makecell{$\Delta$ vs traditional \\ {[95\% CI]}} & 
  \makecell{$p$(Holm)} \\


  \Block{1-2}{Traditional \\ 
  $\{\bar{s}, \bar{e},\ std(s),$ $std(e),\ E_{s}, PP_{s}\}$} && 
  $0.7846 \pm 0.0188$ & 
  reference & 
  --- \\

  \Block{3-1}{The first 12 \\ indicators} & 
  with $p_{rej}$-based & 
  $0.8079 \pm 0.0210$ & 
  \makecell[l]{$+0.0233$ \\ {[+0.0121, $+\infty$)}} & 
  $0.0001$ \\

  & 
  with $p$-based & 
  $0.7975 \pm 0.0232$ & 
  \makecell[l]{$+0.0129$ \\ {[+0.0032, $+\infty$)}} & 
  $0.0038$ \\

  & 
  with $p_{comb}$-based & 
  $0.7983 \pm 0.0177$ & 
  \makecell[l]{$+0.0137$ \\ {[+0.0064, $+\infty$)}} & 
  $0.0002$ \\

  \Block{1-2}{Traditional \\ 
  $\{\bar{s}, \bar{e},\ std(s),$ $std(e),\ E_{s}, PP_{s},$ \\ $\bar{G}_s, \bar{G}_e\}$} && 
  $0.7948 \pm 0.0133$ & 
  reference & 
  --- \\


  \Block{3-1}{The first 15 \\ indicators} & 
  with $p_{rej}$-based & 
  $0.8125 \pm 0.0184$ & 
  \makecell[l]{$+0.0177$ \\ {[+0.0030, $+\infty$)}} & 
  $0.0043$ \\

  & 
  with $p$-based & 
  $0.8090 \pm 0.0215$ & 
  \makecell[l]{$+0.0142$ \\ {[+0.0007, $+\infty$)}} & 
  $0.0052$ \\

  & 
  with $p_{comb}$-based & 
  $0.8122 \pm 0.0189$ & 
  \makecell[l]{$+0.0174$ \\ {[+0.0043, $+\infty$)}} & 
  $0.0038$ \\

  \Block{1-2}{22 indicators\\(traditional + all probabilities)} & & 
  $0.8174 \pm 0.0174$ & 
  \makecell[l]{$+0.0226$ \\ {[+0.0099, $+\infty$)}} & 
  $0.0003$ \\

  \Block{1-2}{Traditional + $PP_{p_{comb}}$} & & 
  $0.8061 \pm 0.0157$ & 
  \makecell[l]{$+0.0113$ \\ {[+0.0026, $+\infty$)}} & 
  $0.0056$ \\
\end{NiceTabular}%
\end{table*}

\begin{figure}[tb]
\centering
\includegraphics[width=0.95\linewidth]{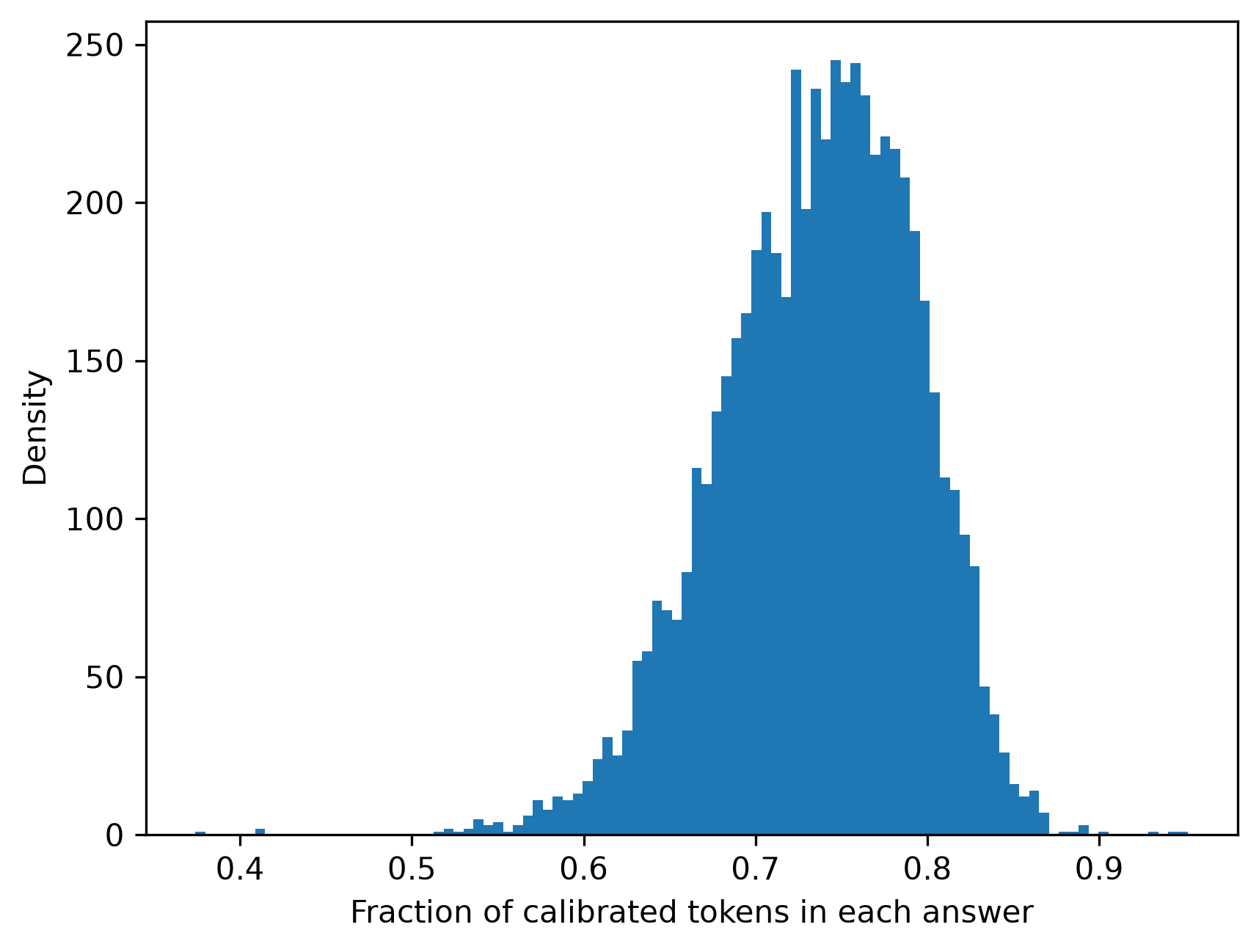}
\caption{Distribution of the fraction of calibrated tokens per answer for MedGemma on the medical dataset. Most answers have $70-80\%$ of their tokens calibrated.}
\label{fig:frac_calib_tokens}
\end{figure}

For Medicine, the effect is more dependent on how many tokens in an answer can be calibrated. Some questions and answers contain rare, domain-specific terms, whereas we build calibrators only for tokens observed at least $1,000$ times. Figure~\ref{fig:frac_calib_tokens} shows the distribution of the fraction of calibrated tokens per answer for MedGemma. Although the largest group of answers has $70–80\%$ calibrated tokens, a non-trivial fraction has lower coverage. We therefore performed an additional analysis in which, for each training fold, we retained only training examples whose calibrated-token fraction exceeded a threshold $TH$. We considered $TH \in [0.40, 0.85]$ in increments of $0.05$, as well as no filtering ($TH = 0$). This analysis is intended to characterize how coverage affects the usefulness of the indicators. 

\begin{figure}[tb]
\centering
\includegraphics[width=0.95\linewidth]{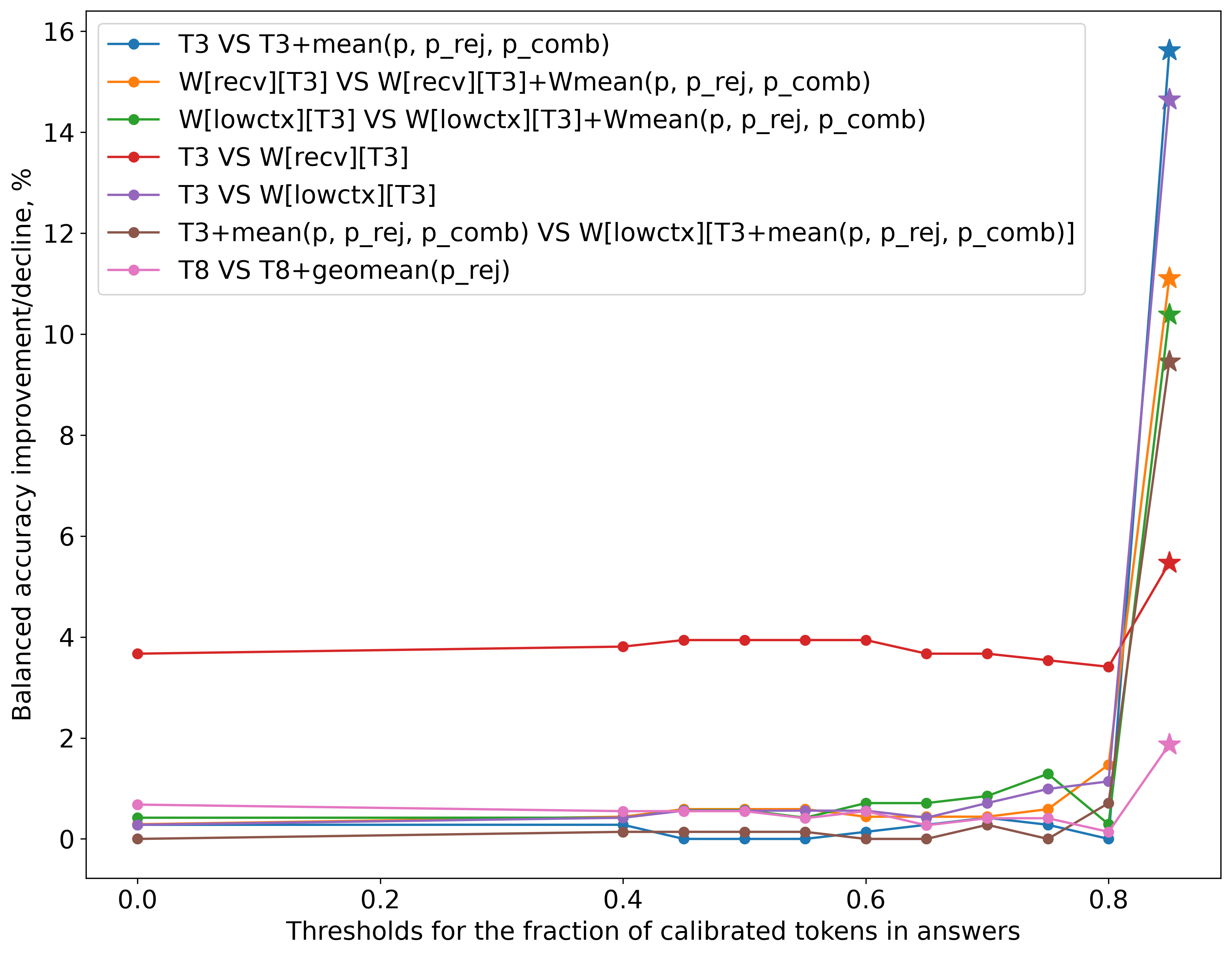}
\caption{Gain in balanced accuracy (\%) as a function of the minimum calibrated-token fraction (threshold $TH$) kept in the training folds, for MedGemma. $TH = 0$ means no filtering. ``T3" is three traditional indicators ($\{\bar{s}, \bar{e},\ std(e)\}$), ``T8" is all eight traditional indicators ($\{\bar{s}, \bar{e},\ std(s), std(e),\ E_{s}, PP_{s},\ \bar{G}_s, \bar{G}_e\}$), ``$W[\text{recv}]$" and ``$W[\text{lowctx}]$" denote attention-weighted versions, and $p$, $p_{rej}$ and $p_{comb}$ denote the probabilities of acceptance, rejection and the combined probability. Stars mark statistically significant improvements. }
\label{fig:mg_stat_signif}
\end{figure}

In Figure~\ref{fig:mg_stat_signif}, stars indicate statistically significant improvements for the comparisons shown. For MedGemma, the gains become statistically significant at the higher calibrated-token coverage threshold ($TH = 0.85$) for the indicated comparisons. This result suggests that the contribution of calibrator-based indicators depends on coverage and should be interpreted conditionally, rather than as a uniform improvement across all medical examples.

\section{Transferability of classifiers}
\label{subset:indicators_transfer}
\begin{figure}[tb]
\centering
\begin{subfigure}[t]{0.49\linewidth}
    \includegraphics[height=4.5cm]{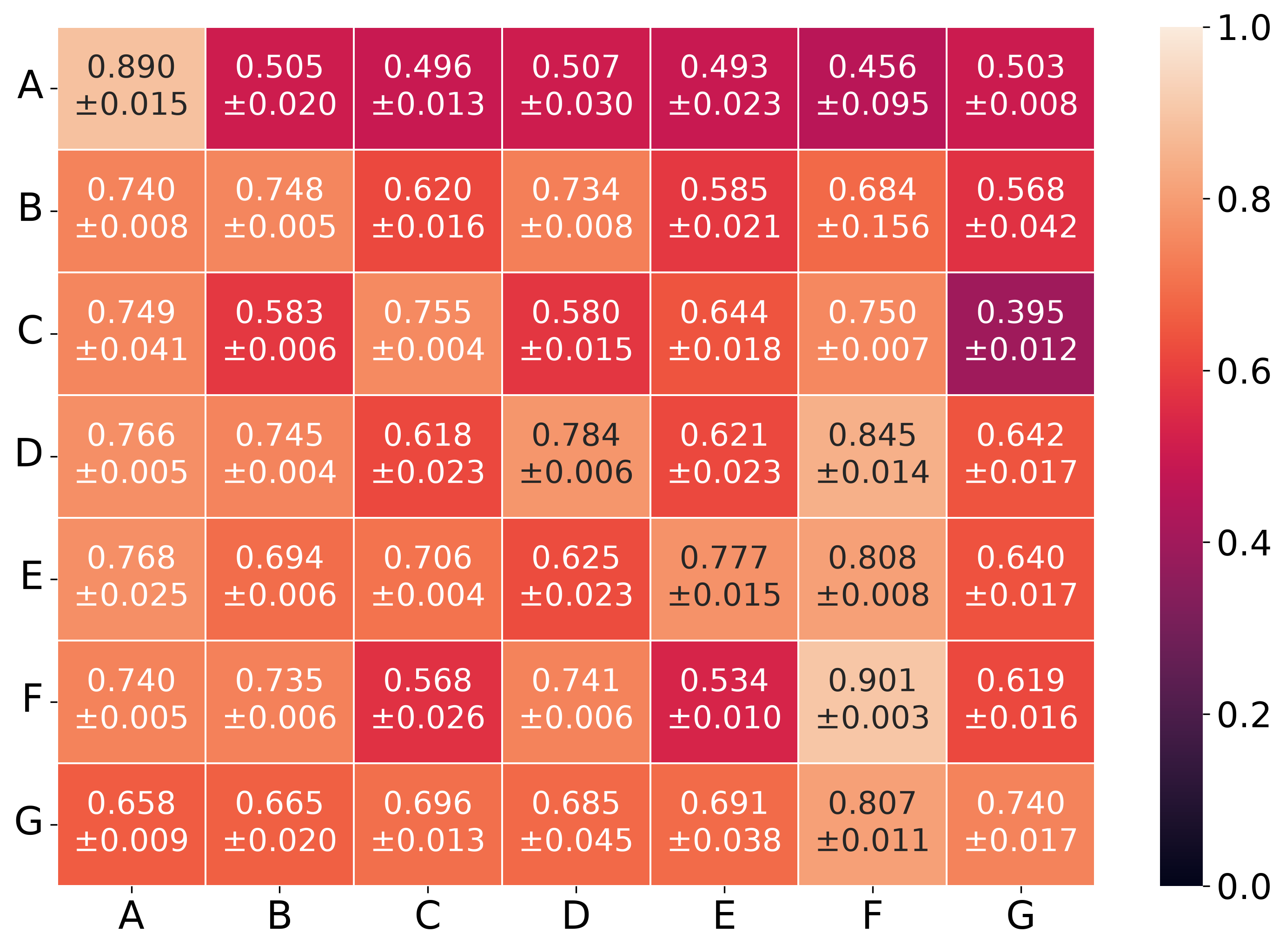}
    \caption{$ \{\mathbf{\bar{p}}, \bar{s}, \bar{e},\ std(s), std(e),\ E_{s},\ PP_{s},\ \bar{G}_e\}$.}
    \label{subfig:transfer_with_probs_roc_auc}
\end{subfigure}
\hfill
\begin{subfigure}[t]{0.49\linewidth}
    \includegraphics[height=4.5cm]{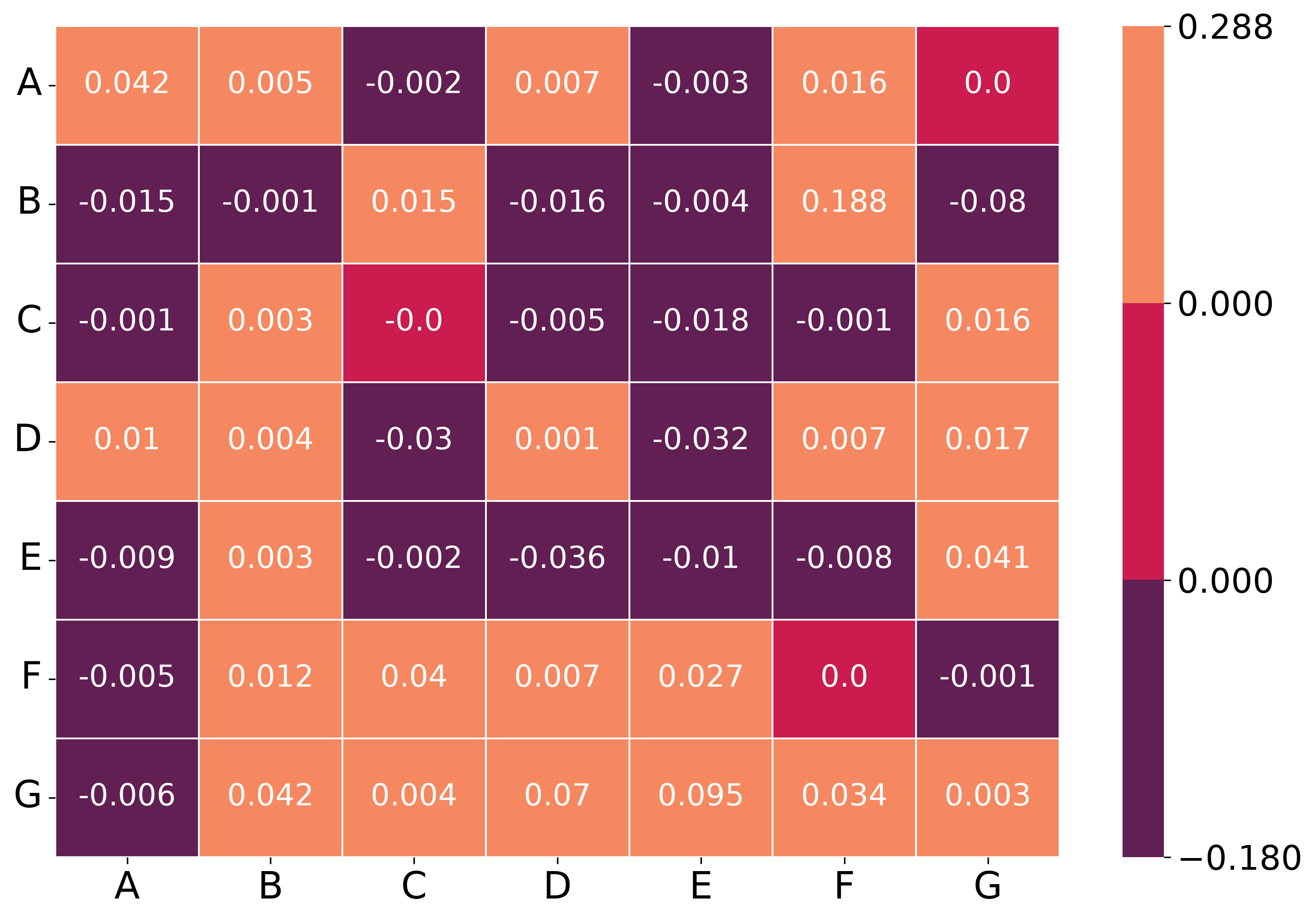}
    \caption{Difference between mean ROC~AUC scores of $ \{\mathbf{\bar{p}}, \bar{s}, \bar{e},\ std(s), std(e),\ E_{s},\ PP_{s},\ \bar{G}_e\}$ and mean F1 scores of $ \{\mathbf{\bar{G}_s}, \bar{s}, \bar{e},\ std(s), std(e),\ E_{s},\ PP_{s},\ \bar{G}_e\}$.}
    \label{subfig:transfer_traditional_roc_auc}
\end{subfigure}
\caption{Aggregated transfer results (method IDs as in Table~\ref{tab:exp_setup}). Each of the $20$ classifiers from $20$-fold cross-validation is transferred to the whole data of each ontology; we report the mean and standard deviation of the \textbf{ROC~AUC score}. In both heatmaps, rows denote the training ontology and columns the test ontology. \textbf{Left:} classifiers using the indicators given in the subcaption. \textbf{Right:} difference in mean ROC~AUC between classifiers on $\{\mathbf{\bar{p}}\} \cup I$ and on $\{\mathbf{\bar{G}_s}\} \cup I$ (traditional indicators only), where $I = \{\bar{s}, \bar{e}, std(s), std(e), E_{s}, PP_{s}, \bar{G}_e\}$; $25$ of $49$ ($51~\%$) transfers favor the indicators with our $\bar{p}$.}
\label{fig:clfs_transfer_roc_auc}
\end{figure}

\begin{figure}[tb]
\centering
\begin{subfigure}[t]{0.49\linewidth}
    \includegraphics[height=4.5cm]{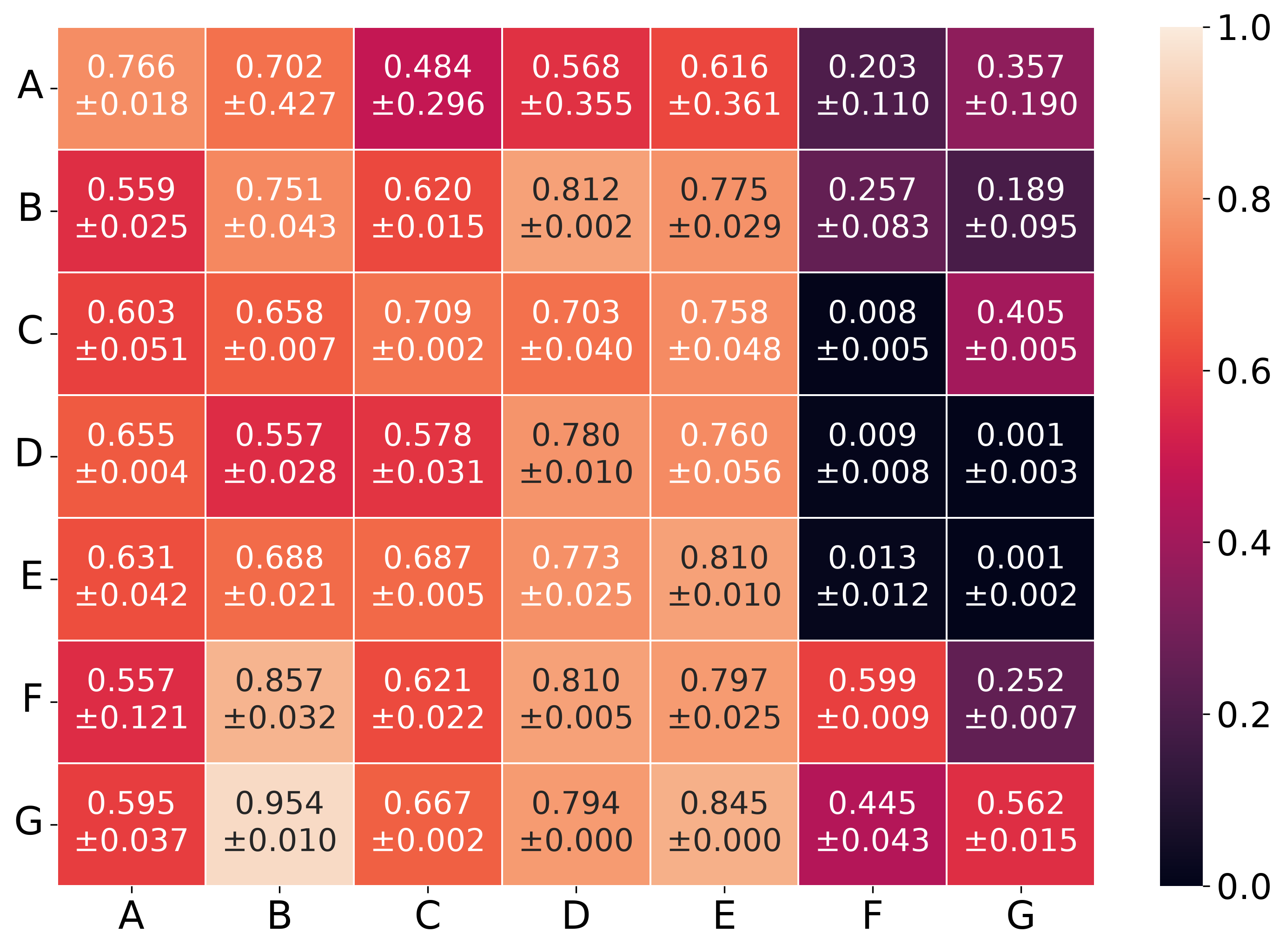}
    \caption{$ \{\mathbf{\bar{p}}, \bar{s}, \bar{e},\ std(s), std(e),\ E_{s},\ PP_{s},\ \bar{G}_e\}$.}
    \label{subfig:transfer_with_probs_f1}
\end{subfigure}
\hfill
\begin{subfigure}[t]{0.49\linewidth}
    \includegraphics[height=4.5cm]{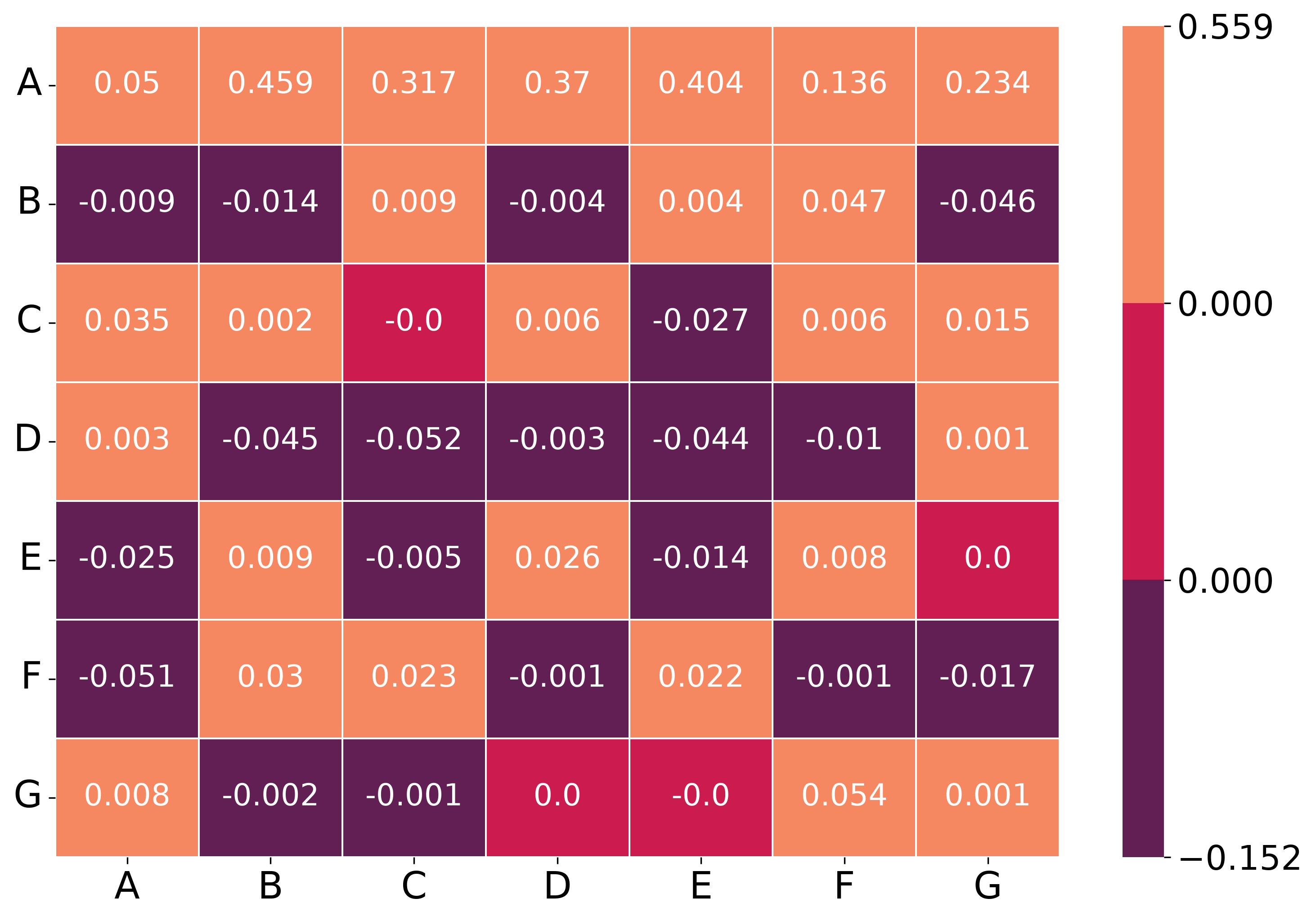}
    \caption{Difference between mean F1 scores of $ \{\mathbf{\bar{p}}, \bar{s}, \bar{e},\ std(s), std(e),\ E_{s},\ PP_{s},\ \bar{G}_e\}$ and mean F1 scores of $ \{\mathbf{\bar{G}_s}, \bar{s}, \bar{e},\ std(s), std(e),\ E_{s},\ PP_{s},\ \bar{G}_e\}$.}
    \label{subfig:transfer_traditional_f1}
\end{subfigure}
\caption{Aggregated transfer results (method IDs as in Table~\ref{tab:exp_setup}). Each of the $20$ classifiers from $20$-fold cross-validation is transferred to the whole data of each ontology; we report the mean and standard deviation of the \textbf{F1 score}. In both heatmaps, rows denote the training ontology and columns the test ontology. \textbf{Left:} classifiers using the indicators given in the subcaption. \textbf{Right:} difference in mean F1 between classifiers on $\{\mathbf{\bar{p}}\} \cup I$ and on $\{\mathbf{\bar{G}_s}\} \cup I$ (traditional indicators only), where $I = \{\bar{s}, \bar{e}, std(s), std(e), E_{s}, PP_{s}, \bar{G}_e\}$; $26$ of $49$ ($53~\%$) transfers favor the indicators with our $\bar{p}$.}
\label{fig:clfs_transfer_f1}
\end{figure}

For a subset of indicators, we evaluate the transferability of all trained MLP-based classifiers across all ontology--model combinations. Figures~\ref{fig:clfs_transfer_roc_auc} and~\ref{fig:clfs_transfer_f1} compares the classifiers' transfer across ontologies in two settings: using the set of traditional indicators with one replaced by our $\bar{p}$ and using the traditional indicators only; the same number of indicators is therefore used in both settings. 
As the figures show, replacing $\bar{G}_s$ with $\bar{p}$ improves the transfer quality in most cases, while in the cases where the traditional indicators perform better, the maximum difference in their favor is only $0.036$ ROC~AUC ($0.052$ F1).

\begin{figure}[tb]
\centering
\begin{subfigure}[t]{0.3\linewidth}
    \centering
    \vspace{0pt}
    \includegraphics[width=\linewidth]{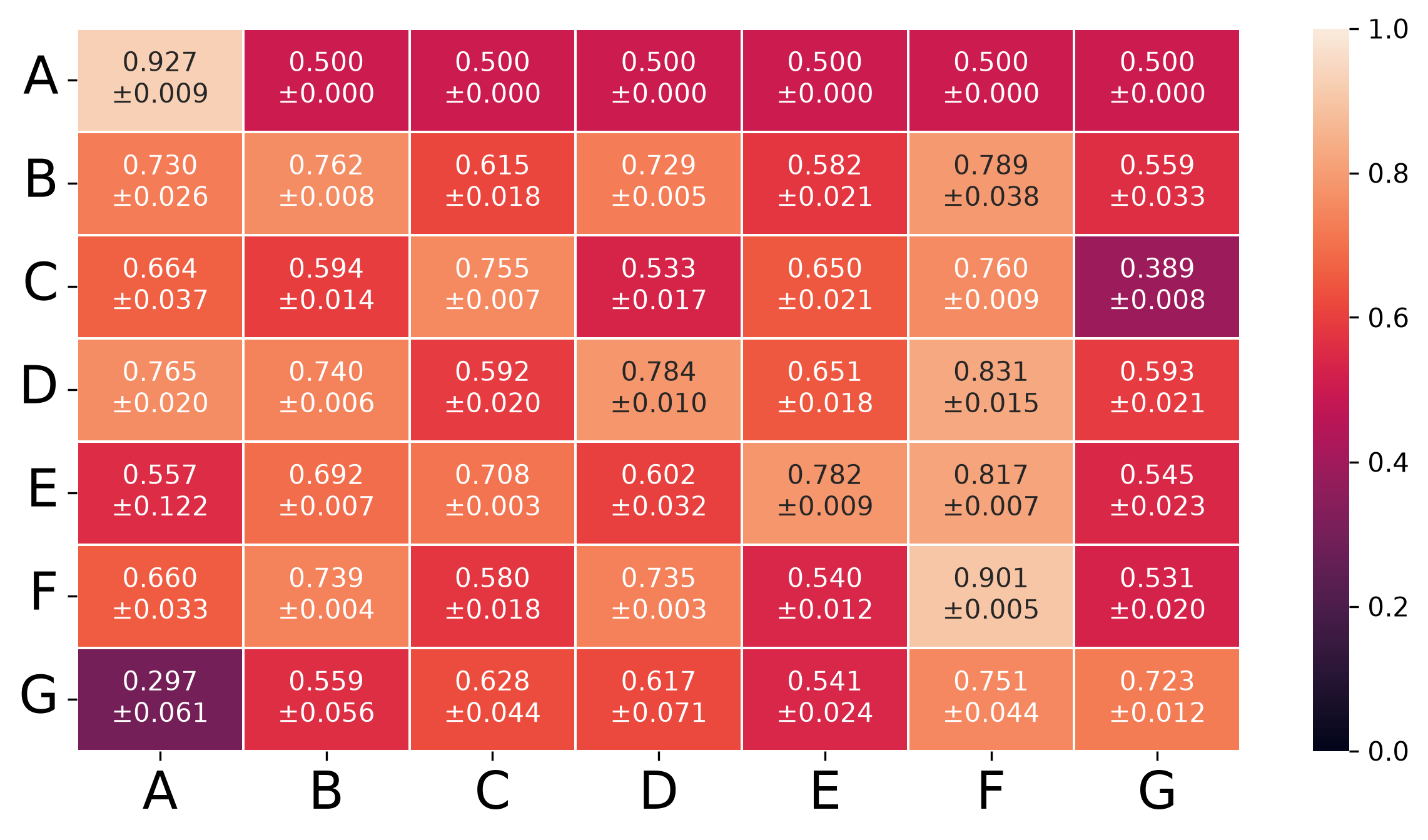}
    \caption{The first $12$ indicators.}
    \label{fig:12_indicators}
\end{subfigure}
\hfill
\begin{subfigure}[t]{0.3\linewidth}
    \centering
    \vspace{0pt}
    \includegraphics[width=\linewidth]{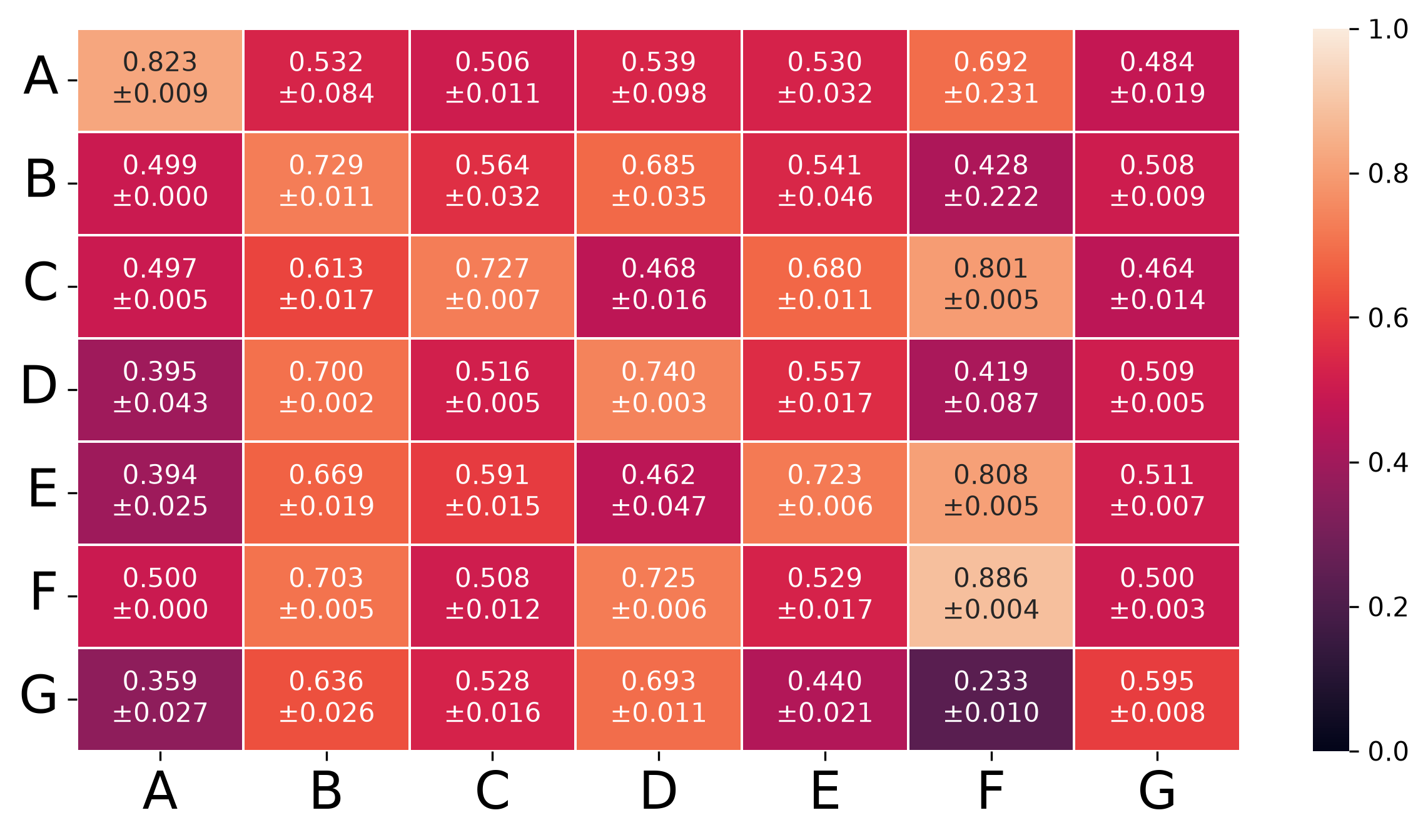}
    \caption{Calibrator-only indicators.}
    \label{fig:calibrator_only_indicators}
\end{subfigure}
\hfill
\begin{subfigure}[t]{0.3\linewidth}
    \centering
    \vspace{0pt}
    \includegraphics[width=\linewidth]{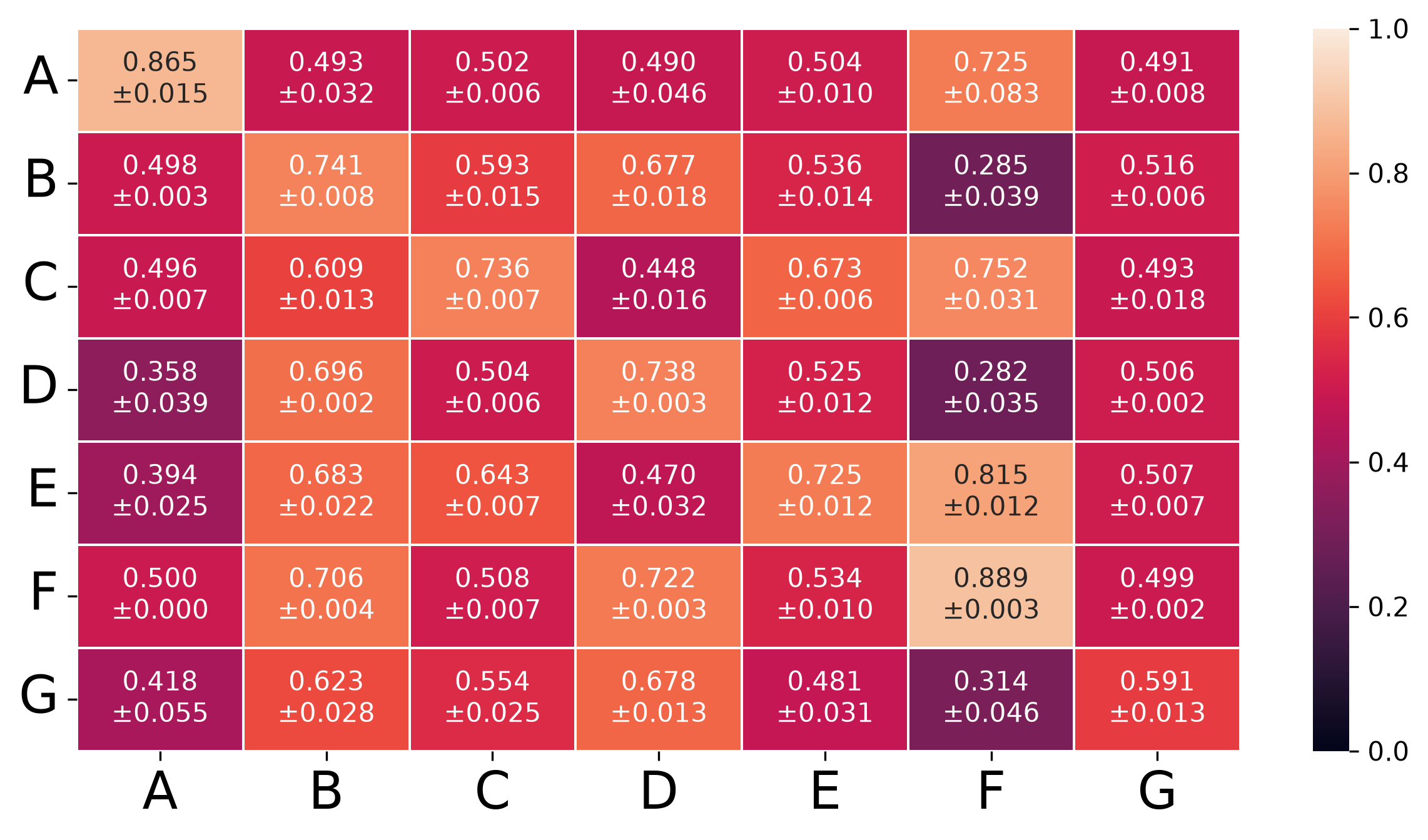}
    \caption{Calibrator-only and mixed indicators.}
    \label{fig:calibrator_mixed_indicators}
\end{subfigure}

\caption{Aggregated transfer results (method IDs as in Table~\ref{tab:exp_setup}). Each of the $20$ classifiers from $20$-fold cross-validation is transferred to the whole data of each ontology; we report the mean and standard deviation of the \textbf{ROC~AUC score}. In all heatmaps, rows denote the training ontology and columns the test ontology. 
}
\label{fig:transfer_additional}
\end{figure}

However, using more calibrator-based indicators does not necessarily improve transfer quality. Figure~\ref{fig:transfer_additional} provides the results for classifier transfer using the first $12$ features ($\{\bar{p}, \bar{s}, \bar{e},\ std(p), std(s), std(e),\ PCC, SCC,\ E_{p}, E_{s},\ PP_{p}, PP_{s}\}$), calibrator-only indicators ($\{\bar{p}, std(p), E_{p}, PP_{p}, \bar{G}_p\}$), and calibrator-only and mixed indicators ($\{\bar{p}, std(p), PCC, SCC, E_{p}, PP_{p}, \bar{G}_p\}$).  The results show that the indicator sets provided in Figure~\ref{fig:transfer_additional} have lower transferability than those provided in Figure~\ref{fig:clfs_transfer_roc_auc}, despite the fact that Figure~\ref{subfig:transfer_with_probs_roc_auc} contains only one calibrator probability.

\section{Additional classifiers' results}
\label{appendix:additional_clf_res}
Our objective is broader than response-correctness prediction alone: we aim to construct ontology-specific, token-level typicality indicators with finite-sample guarantees, and then test whether their aggregation helps predict response correctness. A direct probe addresses the downstream prediction task, but it does not provide the same token-level statistics or acceptance/rejection bounds. Nevertheless, it is an important complementary baseline, and we added a SAPLMA-style comparison.

A SAPLMA-style probe~\citep{azaria-mitchell-2023-internal} is a supervised classifier that predicts response correctness from the model’s hidden activations at the last answer token. Unlike the original, which uses a small feed-forward network and compares several layers, our probe is a Logistic Regression classifier (standardized inputs and class weighting, i.e., the same classifier family as ours) trained on the last-layer hidden state, with fixed hyperparameters and no layer, pooling or model selection. The last-layer state is the input of the LM head, the same representation that our method reads, but our method reads it at every answer token, whereas the probe reads it only at the last one. The probe is trained on the same MedGemma student answers, with the same correctness labels and the same $20$-fold stratified cross-validation as our classifier.

We have also added a comparison of indicators computed from the lower bounds on the probabilities of correct acceptance ($p$) and correct rejection ($p_{rej}$), and from their combination ($p_{comb}$). Results are in Table~\ref{tab:add_clfs_mg}. The SAPLMA-style probe ranks among the top seven methods judged by balanced accuracy; however, unlike our method, it provides no guarantees on the probabilities of correct acceptance and rejection. Moreover, our method yields token-level statistics that reflect how typical each generated token is, and their aggregation reflects the correctness of the whole generation. The SAPLMA-style probe, as proposed, reads the activation of a single token and returns one response-level score, so it provides no per-token characteristics. 


\begin{table}[t]
\centering
\caption{Mean $\pm$ standard deviation of balanced accuracy of Logistic Regression across $20$ folds for MedGemma. Top-1 scores are in \textbf{bold}, top-2 in \underline{\textbf{bold underlined}}, and top-3 in \textbf{\textit{bold italic}}.}
\label{tab:add_clfs_mg}
\resizebox{\textwidth}{!}{%
\begin{NiceTabular}{cclc}
\hline
\makecell{Probability\\type} & \makecell{Indicator\\group} & Indicators & \makecell{Balanced\\Accuracy} \\ \hline

\Block{11-1}{\makecell{Probability\\of rejection}} & \Block{7-1}{Individual} & $\bar{p}_{rej}$ & $0.5189 \pm 0.0210$ \\
& & $\text{std}(p_{rej})$ & $0.5412 \pm 0.0263$ \\
& & $E_{p_{rej}}$ & $0.6668 \pm 0.0367$ \\
& & $PP_{p_{rej}}$ & $0.6674 \pm 0.0413$ \\
& & $\bar{G}_{p_{rej}}$ & $0.5479 \pm 0.0239$ \\
& & $\{ \bar{p}_{rej}, \text{std}(p_{rej}), E_{p_{rej}}, PP_{p_{rej}}, \bar{G}_{p_{rej}} \}$ & $0.6764 \pm 0.0390$ \\
& & $\{ \bar{p}_{rej}, \text{std}(p_{rej}), E_{p_{rej}}, PP_{p_{rej}}, \bar{G}_{p_{rej}}, PCC, SCC \}$ & $0.6787 \pm 0.0331$ \\ \cline{2-4}
& \Block{4-1}{Mixed} & $PCC$ & $0.5866 \pm 0.0255$ \\
& & $SCC$ & $0.5546 \pm 0.0305$ \\
& & 12 indicators & $0.7218 \pm 0.0301$ \\
& & 15 indicators & \underline{\textbf{0.7292 $\pm$ 0.0260}} \\ \hline

\Block{11-1}{\makecell{Probability\\of acceptance}} & \Block{7-1}{Individual} & $\bar{p}$ & $0.5248 \pm 0.0318$ \\
& & $\text{std}(p)$ & $0.5484 \pm 0.0275$ \\
& & $E_{p}$ & $0.6665 \pm 0.0356$ \\
& & $PP_{p}$ & $0.6665 \pm 0.0413$ \\
& & $\bar{G}_{p}$ & $0.5488 \pm 0.0296$ \\
& & $\{ \bar{p}, \text{std}(p), E_{p}, PP_{p}, \bar{G}_{p} \}$ & $0.6690 \pm 0.0377$ \\
& & $\{ \bar{p}, \text{std}(p), E_{p}, PP_{p}, \bar{G}_{p}, PCC, SCC \}$ & $0.6695 \pm 0.0378$ \\ \cline{2-4}
& \Block{4-1}{Mixed} & $PCC$ & $0.5280 \pm 0.0327$ \\
& & $SCC$ & $0.5272 \pm 0.0336$ \\
& & 12 indicators & $0.7211 \pm 0.0321$ \\
& & 15 indicators & $0.7271 \pm 0.0217$ \\ \hline

\Block{11-1}{\makecell{Combined\\probability\\$p_{comb} = \frac{p}{p + (1 - p_{rej})}$}} & \Block{7-1}{Individual} & $\bar{p}_{comb}$ & $0.5313 \pm 0.0187$ \\
& & $\text{std}(p_{comb})$ & $0.5537 \pm 0.0262$ \\
& & $E_{p_{comb}}$ & $0.6662 \pm 0.0364$ \\
& & $PP_{p_{comb}}$ & $0.6669 \pm 0.0413$ \\
& & $\bar{G}_{p_{comb}}$ & $0.5298 \pm 0.0177$ \\
& & $\{ \bar{p}_{comb}, \text{std}(p_{comb}), E_{p_{comb}}, PP_{p_{comb}}, \bar{G}_{p_{comb}} \}$ & $0.6779 \pm 0.0378$ \\
& & $\{ \bar{p}_{comb}, \text{std}(p_{comb}), E_{p_{comb}}, PP_{p_{comb}}, \bar{G}_{p_{comb}}, PCC, SCC \}$ & $0.6758 \pm 0.0350$ \\ \cline{2-4}
& \Block{4-1}{Mixed} & $PCC$ & $0.5853 \pm 0.0299$ \\
& & $SCC$ & $0.5395 \pm 0.0329$ \\
& & 12 indicators & $0.7233 \pm 0.0333$ \\
& & 15 indicators & $0.7284 \pm 0.0251$ \\ \hline

\Block{3-1}{\makecell{Several\\probabilities}} & \Block{3-1}{Mixed} & \makecell[l]{22 indicators (traditional + all probabilities)} & $0.7286 \pm 0.0244$ \\
& & \makecell{$\{ \bar{p}_{rej}, \text{std}(p_{rej}), E_{p_{rej}}, PP_{p_{rej}}, \bar{G}_{p_{rej}}$ \\ $\bar{p}, \text{std}(p), E_{p}, PP_{p}, \bar{G}_{p} \}$} & $0.6858 \pm 0.0326$ \\
& & \makecell{21 indicators\\(all with $p_{rej}$, $p$, and $p_{comb}$)} & $0.6924 \pm 0.0333$ \\ \hline

\Block{10-1}{Traditional} & \Block{8-1}{Individual} & $\bar{e}$ & $0.6123 \pm 0.0243$ \\
& & $\text{std}(e)$ & $0.6579 \pm 0.0235$ \\
& & $\bar{G}_e$ & $0.6079 \pm 0.0340$ \\
& & $\bar{s}$ & $0.5907 \pm 0.0265$ \\
& & $\text{std}(s)$ & $0.6113 \pm 0.0222$ \\
& & $E_s$ & $0.6767 \pm 0.0367$ \\
& & $PP_s$ & $0.6747 \pm 0.0418$ \\
& & \color{baselinered}$\bar{G}_s$ & \color{baselinered}$0.5952 \pm 0.0241$ \\ \cline{2-4}
& \Block{2-1}{Mixed} & $\{\bar{s}, \bar{e},\ \text{std}(s), \text{std}(e),\ E_{s},PP_{s} \}$ & $0.7211 \pm 0.0325$ \\
& & $\{\bar{s}, \bar{e},\ \text{std}(s), \text{std}(e),\ E_{s},PP_{s},\ \bar{G}_s, \bar{G}_e\}$ & \textbf{\textit{0.7291 $\pm$ 0.0245}} \\ \hline

\makecell{Traditional\\+ Ours} & Mixed & \makecell[l]{Best combination: \\$\{\bar{s}, \bar{e},\ \text{std}(s), \text{std}(e),\ E_{s},PP_{s},\ \bar{G}_s, \bar{G}_e, PCC(p_{rej})\}$} & \textbf{0.7319 $\pm$ 0.0260} \\ \hline

\Block{11-2}{\color{baselinered}Baselines}
& & \color{baselinered}SAPLMA & \color{baselinered}$0.7247 \pm 0.0202$ \\
& & \color{baselinered}MSP & \color{baselinered}$0.6637 \pm 0.0339$ \\
& & \color{baselinered}LM Perplexity & \color{baselinered}$0.5977 \pm 0.0245$ \\
& & \color{baselinered}Max Token Entropy & \color{baselinered}$0.6648 \pm 0.0259$ \\
& & \color{baselinered}Mean Token Probability & \color{baselinered}$0.5911 \pm 0.0264$ \\
& & \color{baselinered}Min Token Probability & \color{baselinered}$0.6376 \pm 0.0353$ \\
& & \color{baselinered}Geomean Token Probability & \color{baselinered}$0.5959 \pm 0.0221$ \\
\cline{3-4}
 & & \color{baselinedarkred}Const = 1 & \color{baselinedarkred}$0.5000 \pm 0.0000$ \\
& & \color{baselinedarkred}Const = 0 & \color{baselinedarkred}$0.5000 \pm 0.0000$ \\
& & \color{baselinedarkred}Random -- fair coin & \color{baselinedarkred}$0.5018 \pm 0.0304$ \\
& & \color{baselinedarkred}Random -- $p$ of major class & \color{baselinedarkred}$0.4935 \pm 0.0320$ \\
\hline
\end{NiceTabular}%
}
\end{table}

Moreover, it is necessary to isolate the contribution of calibrator-based indicators from that of traditional confidence indicators.
To address this concern, we conducted an additional ablation study comparing three feature sets: (1) the eight traditional indicators, (2) the five calibrator-based indicators together with the two mixed indicators, and (3) the complete set of $15$ indicators used in the submitted paper. For each setting, we report the best-performing feature combination under the experimental selection procedure. This comparison separates the performance of traditional indicators alone from that of calibrator-derived and combined indicators.
Table~\ref{tab:best_clfs_categories} presents the mean and standard deviation of balanced accuracy obtained by an MLP across $20$ folds for all seven student/domain settings.

\begin{table}[t]
\centering
\caption{Mean $\pm$ standard deviation of balanced accuracy of MLP across $20$ folds for all students. Top-1 scores are in \textbf{bold}, top-2 in \underline{\textbf{bold underlined}}.}
\label{tab:best_clfs_categories}
\resizebox{\textwidth}{!}{%
\begin{NiceTabular}{c|cccc|c|cc}
\hline
\Block{2-1}{}
 & \Block{1-4}{Medicine} & & & & BF & \Block{1-2}{Math, V1} & \\
\cline{2-8}
 & \makecell{MedGemma}
 & \makecell{Qwen3-8B}
 & \makecell{Qwen3-0.6B}
 & \makecell{Qwen3.5-0.8B}
 & \makecell{Custom\\Transformer}
 & \makecell{LLM-Judge}
 & \makecell{Process\\Bench} \\
\hline

\Block{2-1}{\makecell[l]{Best combination\\among eight traditional\\indicators}}
 & \underline{\textbf{0.7068$\pm$0.0106}}
 & \underline{\textbf{0.7259$\pm$0.0159}}
 & \underline{\textbf{0.8556$\pm$0.0101}}
 & \underline{\textbf{0.6971$\pm$0.0360}}
 & 0.6238$\pm$0.1339
 & \underline{\textbf{0.6357$\pm$0.2082}}
 & \underline{\textbf{0.7128$\pm$0.2361}} \\
\cline{2-8}
 & \makecell{$\bar{e}$, $\bar{G}_e$,\\$\bar{s}$, $E_s$}
 & \makecell{$std(e)$, $\bar{G}_e$,\\$\bar{s}$, $std(s)$, $E_s$}
 & \makecell{$std(e)$, $\bar{G}_e$,\\$\bar{s}$, $PP_s$, $\bar{G}_s$}
 & \makecell{$std(e)$, $\bar{G}_e$,\\$E_s$, $\bar{G}_s$}
 & \makecell{$\bar{e}$, $std(e)$,\\$\bar{G}_e$, $\bar{s}$,\\$std(s)$, $PP_s$,\\$\bar{G}_s$}
 & \makecell{$\bar{e}$, $\bar{G}_e$,\\$std(s)$}
 & \makecell{$\bar{e}$, $std(e)$,\\$\bar{G}_e$, $\bar{s}$} \\
\hline

\Block{2-1}{\makecell[l]{Best combination\\among five calibrator\\indicators + two mixed}}
 & 0.6724$\pm$0.0135
 & 0.6744$\pm$0.034
 & 0.8514$\pm$0.019
 & 0.5827$\pm$0.029
 & \underline{\textbf{0.6323$\pm$0.0164}}
 & 0.6236$\pm$0.1954
 & 0.6891$\pm$0.2167 \\
\cline{2-8}
 & \makecell{$std(p)$,\\$E_p$, $PP_p$}
 & \makecell{$\bar{p}$, $E_p$,\\$PCC$, $SCC$}
 & \makecell{$std(p)$,\\$E_p$, $PP_p$}
 & \makecell{$PP_p$, $\bar{G}_p$}
 & \makecell{$\bar{p}$, $std(p)$, $\bar{G}_p$,\\$PCC$, $SCC$}
 & \makecell{$std(p)$, $PP_p$,\\$\bar{G}_p$, $PCC$}
 & \makecell{$\bar{p}$, $E_p$, $PP_p$,\\$\bar{G}_p$, $PCC$, $SCC$} \\
\hline

\Block{2-1}{\makecell[l]{Best combination\\among all 15\\indicators (from our\\submitted paper)}}
 & \textbf{0.7141$\pm$0.0286}
 & \textbf{0.7284$\pm$0.0417}
 & \textbf{0.8604$\pm$0.0334}
 & \textbf{0.7035$\pm$0.0557}
 & \textbf{0.8299$\pm$0.1130}
 & \textbf{0.6389$\pm$0.2116}
 & \textbf{0.7153$\pm$0.2354} \\
\cline{2-8}
 & \makecell{$\bar{s}$, $\bar{e}$, $std(s)$,\\$E_p$, $PP_p$,\\$PP_s$, $\bar{G}_p$,\\$\bar{G}_s$, $\bar{G}_e$}
 & \makecell{$\bar{s}$, $\bar{e}$, $std(s)$,\\$E_p$, $E_s$,\\$PP_s$, $\bar{G}_s$,\\$\bar{G}_e$}
 & \makecell{$\bar{p}$, $std(p)$,\\$\bar{e}$, $std(e)$,\\$\bar{s}$, $E_s$,\\$\bar{G}_p$, $\bar{G}_s$}
 & \makecell{$\bar{s}$, $std(e)$,\\$E_p$, $PP_s$,\\$\bar{G}_s$, $\bar{G}_p$}
 & \makecell{$\bar{p}$, $\bar{s}$, $std(p)$,\\$std(s)$, $std(e)$,\\$PCC$, $E_p$, $E_s$,\\$PP_s$, $\bar{G}_p$,\\$\bar{G}_s$, $\bar{G}_e$}
 & \makecell{$\bar{s}$, $\bar{e}$,\\$PCC$, $SCC$,\\$E_p$, $PP_p$,\\$\bar{G}_e$}
 & \makecell{$\bar{p}$, $std(p)$,\\$std(s)$, $E_p$,\\$E_s$, $PP_p$,\\$PP_s$, $\bar{G}_p$,\\$\bar{G}_s$, $\bar{G}_e$} \\
\hline
\end{NiceTabular}%
}
\end{table}

The results show that traditional indicators alone provide strong baselines, while calibrator-based indicators alone generally achieve lower performance. During training, the softmaxes are configured so that the overall generation is correct. Obviously, ours do not have to be correct, because we focus on typicality, not correctness. When combined, there is a cumulative effect that traditional indicators do not show. Where there is ground truth (BF), correctness is linked to typicality, and our metrics are better than the traditional ones.
These results suggest that the calibrator-based indicators provide complementary information to traditional confidence indicators, rather than simply replacing them. The combined feature set consistently achieves higher mean balanced accuracy than the best combination of traditional indicators alone.

\section{Licenses on used data, models and software}
\label{appendix:licences}

In the paper the following models, software and licences are used:

\begin{enumerate}
    \item Qwen-2.5 -- Apache 2.0 license;
    \item DeepSeek V3.2 -- MIT license;
    \item MedGemma-2.5-4B -- health-ai-developer-foundations license;
    \item SGLang inference server -- Apache 2.0 license;
    \item XGBoost algorithm -- Apache 2.0 license;
\end{enumerate}

\section{Societal impacts}
\label{appendix:societal_impacts}

The proposed framework aims to improve the reliability of generative AI systems by providing lightweight indicators of atypical and potentially incorrect outputs, which could be valuable in high-stakes ontologies such as medicine, software engineering and decision support. If used appropriately, such methods may highlight fluent but unreliable generations and help route unreliable cases to stronger verification pipelines. At the same time, the method should not be interpreted as a guarantee of factual correctness: false negatives may induce false confidence, while false positives may suppress useful outputs. These risks are particularly important in medical ontology, where detector scores should serve only as additional evidence rather than as expert judgment. Careful ontology-specific calibration, transparent reporting of detector non-confidence, and validation under distribution shift are necessary before deployment.

\end{document}